%% file: main.tex
\documentclass[nonacm, acmlarge, screen, natbib=false]{acmart}

\RequirePackage{textcase}

\usepackage{packages/standard}
\usepackage{packages/commands}

\RequirePackage[
  datamodel=acmdatamodel,
  style=acmauthoryear,
  backend=biber,
  giveninits=true,
  uniquename=init,
  natbib=true
]{biblatex}

\makeatletter
\let\oldabstract\abstract
\let\oldendabstract\endabstract
\renewenvironment{abstract}
  {\oldabstract\defcounter{biburlnumpenalty}{0}\defcounter{biburlucpenalty}{0}}
  {\oldendabstract}
\makeatother

\begin{document}

\title[Runtime Efficient Network to Tree Transformation (RENTT)]{Efficiently Transforming Neural Networks into Decision Trees: A Path to Ground Truth Explanations with RENTT}


\author{Helena Monke}
\authornote{These authors contributed equally to this work.}
\orcid{0009-0002-9844-4277}
\email{helena.monke@ipa.fraunhofer.de}
\affiliation{%
  \institution{Fraunhofer Institute for Manufacturing Engineering and Automation IPA}
  \city{Stuttgart}
  \country{Germany,}
}
\affiliation{%
  \institution{Institute of Industrial Manufacturing and Management IFF, University of Stuttgart}
  \city{Stuttgart}
  \country{Germany}
}

\author{Benjamin Fresz}
\authornotemark[1]
\orcid{0009-0002-7463-8907}
\email{benjamin.fresz@ipa.fraunhofer.de}
\affiliation{%
  \institution{Fraunhofer Institute for Manufacturing Engineering and Automation IPA}
  \city{Stuttgart}
  \country{Germany,}
}
\affiliation{%
  \institution{Institute of Industrial Manufacturing and Management IFF, University of Stuttgart}
  \city{Stuttgart}
  \country{Germany}
}

\author{Marco Bernreuther}
\email{marco.bernreuther@ipa.fraunhofer.de}
\affiliation{%
  \institution{Fraunhofer Institute for Manufacturing Engineering and Automation IPA}
  \city{Stuttgart}
  \country{Germany}
}

\author{Yilin Chen}
\email{yilin.chen@tum.de}
\affiliation{%
  \institution{Technical University Munich}
  \city{Munich}
  \country{Germany}
}
\affiliation{%
  \institution{Institute of Industrial Manufacturing and Management IFF, University of Stuttgart}
  \city{Stuttgart}
  \country{Germany}
}

\author{Marco F. Huber}
\orcid{0000-0002-8250-2092}
\email{marco.huber@ieee.org}
\affiliation{%
  \institution{Fraunhofer Institute for Manufacturing Engineering and Automation IPA}
  \city{Stuttgart}
  \country{Germany}
}
\affiliation{%
  \institution{Institute of Industrial Manufacturing and Management IFF, University of Stuttgart}
  \city{Stuttgart}
  \country{Germany}
}

\renewcommand{\shortauthors}{Monke, Fresz, Bernreuther, Chen \& Huber}

\begin{abstract}
\textbf{Background:} 
Although neural networks are a powerful tool, their widespread use is hindered by their black-box nature. Methods in the field of explainable AI (XAI) try to solve this problem by explaining entire models or single decisions. But they are plagued by missing faithfulness/correctness, meaning that their explanations do not always conform to the model in question. Recently, methods to transform neural networks into decision trees have been proposed to overcome such problems. Unfortunately, they typically lack exactness, scalability, or interpretability as the size of the neural network grows.
 
\noindent\textbf{Objectives:} We provide a way to efficiently compute ground truth explanations for neural networks via a theoretically founded and practically implemented transformation from neural networks into multivariate decision trees, called Runtime Efficient Network to Tree Transformation (RENTT). These trees can - when simplified via feature importance explanations - serve as ground truth explanations which perfectly conform to the neural network in question.
 
\noindent\textbf{Methods:} We generalize previous results, especially by considering convolutional neural networks, recurrent neural networks, non-rectified linear unit activation functions, and bias terms. Our findings are accompanied by rigorous proofs and we present a novel algorithm RENTT designed to compute an exact equivalent decision tree representation of neural networks in a runtime and memory efficient manner. The resulting decision trees are multivariate and thus possibly too complex to understand. To alleviate this problem, we also provide a method to calculate the ground truth feature importance for neural networks via the equivalent decision trees---for entire models (global), specific input regions (regional), or single decisions (local). We compare the runtime and memory consumption of our transformation to an existing implementation from \textcite{Ngu20} for rectified linear unit (ReLU)-based classification networks and the results of our method for regression and classification tasks to common feature importance methods (local: LIME, SHAP, BreakDown; global: Shapley Effects, SAGE).
All code is publicly available via \href{https://github.com/HelenaM23/RENTT}{GitHub}.
 
\noindent\textbf{Results:} The experimental results emphasize the computational efficiency and scalability of our algorithm, which directly stem from our theoretical insights. For more complex networks and data, e.g., image tasks, scalability issues may still arise. Common feature importance methods produce different explanations than RENTT, while only RENTT conforms to expectation for cases with known ground truth explanations. These differences increase in more complex cases.
 
\noindent\textbf{Conclusions:} Further developing the implementation shows significant potential for practically achievable ground truth explainability for complex neural networks. Post-hoc explanation methods should be treated with care, as they fail to reproduce known ground truth explanations.

\end{abstract}


\maketitle


\input{Chapter/Acronym}

\input{Chapter/1_Introduction}

\input{Chapter/2_RelatedWorks}

\input{Chapter/3_Theory}

\input{Chapter/4_Algorithm}

\input{Chapter/5_FIExperiments}

\input{Chapter/7_Conclusion}

\begin{acks}
This paper is funded in parts by the German Research Foundation (DFG) Priority Programme (SPP 2422) “Data-Driven Process Modelling in Forming Technology” (Subproject “Transparent AI-supported process modeling in drop forging” 520195047)
\end{acks}







\printbibliography

\clearpage

\appendix
\input{Chapter/Appendix_cleaned}



\end{document}

%% file: Chapter/Acronym.tex
\begin{acronym}
\acro{cnn}[CNN]{convolutional neural network}
\acro{nn}[NN]{neural network}
\acro{rnn}[RNN]{recurrent neural network}
\acro{fcnn}[FcNN]{fully connected neural network}
\acro{ml}[ML]{machine learning}
\acro{xai}[XAI]{eXplainable Artificial Intelligence}
\acro{dt}[DT]{decision tree}
\acro{fi}[FI]{feature importance}
\acro{rmse}[RMSE]{Root Mean Square Error}
\acro{mae}[MAE]{ Mean Abolute Error}
\acrodef{re}[RE]{Relative Error}
\acro{ram}[RAM]{Random-access memory}
\acro{rentt}[RENTT]{Runtime Efficient Network to Tree Transformation}
\acro{rentt-fi}[RENTT-FI]{RENTT-Feature Importance}
\acro{relu}[ReLU]{Rectified Linear Unit}
\acro{lime}[LIME]{Local Interpretable Model-agnostic Explanations}
\acro{shap}[SHAP]{SHapley Additive Explanations}
\acro{bd}[BD]{BreakDown}
\acro{sage}[SAGE]{Shapley Additive Global importancE}
\acro{se}[SE]{Shapley Effects}
\end{acronym}

%% file: Chapter/1_Introduction.tex
\section{Introduction}
\Acp{nn} have gained significant importance in the field of \ac{ml}, offering high-performing solutions across various domains like image recognition, natural language processing, and more~\cite{pinecone_imagenet_2023}. Despite their success, the inherent opacity of NNs poses a significant challenge, limiting their broader adoption in sensible or safety-related application. The lack of interpretability and the ``black-box'' nature of \acp{nn} often lead to skepticism regarding their trustworthiness and internal decision-making processes, as users cannot understand and thus, trust, the rationale behind the outputs. This problem is particularly acute in fields such as healthcare, finance, and legal systems, where the consequences of decisions can be profound~\cite{dzindolet2003,Rudin2019}.

To address these concerns, the field of \ac{xai} aims developing techniques that explain and clarify the decision-making processes of \ac{ml}. The goal of \ac{xai} is to make the outputs of \ac{ml} models more understandable to humans, which can help in validating the fairness and reliability of the \ac{ml} model's decisions.

In general, \ac{xai} methods try to provide insights into the reasoning behind a model's behavior, which could be used to show biases or other flaws in models or to calibrate user trust~\cite{Zhang2020,ma2023itrustaimyself}.
Despite ongoing efforts, many current \ac{xai} methods struggle with providing explanations that accurately reflect the true reasoning of the \acp{nn}~\cite{kindermans2017unreliabilitysaliencymethods,adebayo2018sanityChecks}. Common methods to provide explanations for given ``black-box'' models are either local ones like LIME (Local interpretable model-agnostic explanations)~\cite{Ribeiro2016LIME}, SHAP (SHapley Additive exPlanations)~\cite{Lundberg2017SHAP}, and gradient based methods or  global methods for example SE (Shapley Effects) and SAGE (Shapley Additive Global importancE). 

For enhancing interpretability---the general intelligibility of an \ac{ml} model---a common approach is training an interpretable surrogate  model like a \ac{dt}. \Acp{dt}, if small enough, are more interpretable because their structure allows for the tracing of decision paths, providing clear and logical reasoning for each decision. Instead of training a surrogate model, new approaches aim to transform \acp{nn} into equivalent \acp{dt}~\cite{Ayt22,Ngu20}. Nonetheless, the transformation from \acp{nn} to \acp{dt} is fraught with challenges, such as providing a correct derivation of the transformation also for advanced network architectures such as \acp{cnn} and \acp{rnn} with general activation functions and not just \acp{fcnn} with \ac{relu} activation. Additionally, an efficient implementation of the transformation, which also works for large networks, is missing. As the size of the decision tree grows, it becomes increasingly difficult to maintain its interpretability due to the complex branching and depth of the tree.
To accurately represent the \acp{nn}, the corresponding \acp{dt} need to be multivariate, i.e., use multiple features per decision node.
As such, direct interpretation of these \acp{dt} is challenging.
Nevertheless, they provide a foundation for deriving more intuitive explanations, in our case so-called feature importance.

To alleviate the various limitations of existing \ac{nn} to \ac{dt} transformation methods, we make the following contributions to the field of XAI methods:
\begin{itemize}
    \item A clear and comprehensive derivation of the theory of transforming \acp{nn} to multivariate \acp{dt}, covering traditional networks and more complex architectures such as \acp{cnn} and \acp{rnn}, with general (including non-\ac{relu}-like) activation functions and bias terms, as well as networks for regression and classification tasks.
    \item A novel algorithm for efficiently computing an exact equivalent \ac{dt} representation of an \ac{nn}, called \ac{rentt}, enhancing computational efficiency in terms of memory consumption and runtime.
    \item A theoretical framework and implementation for extracting the ground truth feature importance from \acp{nn}, called \ac{rentt-fi}, at global (entire dataset), regional (specific input regions), and local (individual decisions) levels, providing comprehensive insights into model decision-making and identifying key output-driving features.
\end{itemize}

We begin by contextualizing our work within the existing literature in Section~\ref{sec:related_works}. Section~\ref{sec:Theory} introduces the theoretical foundation for transforming \acp{nn}---including \acp{fcnn}, \acp{cnn}, and \acp{rnn}—into \acp{dt}, while Section~\ref{sec:Algorithms} presents our \ac{rentt} transformation algorithm. We validate our approach through comprehensive numerical experiments that demonstrate the effectiveness of the proposed transformation. In Section~\ref{sec:fi-definition}, we introduce \ac{rentt-fi}, a \ac{fi} calculation method  based on the transformed \ac{dt}. Section~\ref{sec:FIexperiments} presents experimental validation of this approach, assessing both the correctness and utility of our \ac{fi} calculations, as well as evaluating the computational efficiency of our implementation. By providing a more accurate and scalable approach to \ac{xai}, we aim to bridge the gap between the capabilities of \acp{nn} and the need for their decisions to be interpretable and trustworthy. Section~\ref{sec:discussion-limitations} discusses our results and their limitations, while outlining future research directions aimed at overcoming these identified challenges. Finally, Section~\ref{sec:conclusion} provides our concluding remarks.

%% file: Chapter/2_RelatedWorks.tex
\section{Related Works}
\label{sec:related_works}
The need for better understanding the inner workings of \acp{nn} has led to a plethora of innovative approaches. In this section, we will explore how these methodologies aim to provide a clearer understanding of \acp{nn}' internal mechanisms and decision processes. 
Traditional methods for interpreting \acp{nn} often lack guarantees of exactness and consistency, leading to potential misinterpretations. Some approaches address this by providing an equivalent description of an \ac{nn}. It has been shown that  \acp{nn} can be exactly written as compositions of max-affine spline operators (MASOs)~\cite{Balestriero.2021}, which also enable the approximation of non-piecewise linear activation functions with nonlinearities.  This representation simplifies the understanding of their internal mechanics and decision-making processes. By establishing connections to approximation theory, the authors highlight how the structure of deep \ac{nn} influences their own behavior and performance. In~\textcite{Wang2019} this is extended to a closer focus on \acp{rnn} and their representation as simple piecewise affine spline operators. This representation reveals how RNNs partition the input space over time and suggests that the affine slope parameters function as input-specific templates, enhancing interpretability through a matched filtering perspective. In~\textcite{Sudjianto.2020} \ac{relu} networks are formulated as local linear models, where each activation pattern of the \ac{nn} results in a local linear model. The activation regions can then be visualized for better interpretability or used to calculate the joint importance of the feature.~\textcite{Chu.2018} proposes a closed-form solution, specifically designed for piecewise linear \acp{nn}, where the network is transformed into a mathematically equivalent set of local linear classifiers. This transformation allows exact interpretations of the model's decision boundary by identifying the features that dominate predictions for each local linear classifier.

MASOs or local linear models can also be used to represent an \ac{nn} as a \ac{dt}, as shown in~\textcite{Ngu20,Ayt22}. This claim aims to tackle the interpretability issues associated with \acp{nn}. In~\textcite{Ngu20}, the authors provide two algorithms, one for an exact and one for an approximate but less complex transformation of fully connected \ac{relu} networks used for binary or multiclass classification tasks into multivariate \acp{dt}. Their method aims to extract interpretable decision rules and preserves the decision boundaries learned by the network. Similarly,~\textcite{Ayt22} emphasizes the transformation of \acp{nn} into \acp{dt}, with the goal that any \ac{nn} can be represented as a \ac{dt}. The research covers various types of \acp{nn}, including feedforward \acp{nn}, single-layer \acp{rnn}, and \acp{cnn}, focusing on regression and classification tasks. 
Together, these studies highlight a concerted effort in the research community to enhance the interpretability of \acp{nn}, providing insights and methodologies that facilitate a deeper understanding of these complex models. The integration of spline representations, \ac{dt} transformations, and closed-form solutions paves the way for future advancements in the interpretability of \acp{nn}, fostering trust and reliability in \ac{ml} models.

However, these findings are either lacking practical usage and implementation for a transformation or have potential gaps in mathematical rigor, for example by excluding considerations for activation function offsets or bias terms in the transformation process. This especially leaves open questions regarding the implementation. All in all, there exists no efficient and scalable algorithm for transforming an \ac{fcnn}, \ac{rnn} or \ac{cnn} with any picewise linear activation function into an interpretable model, while also providing a thorough mathematical proof of its validity. This makes the usage of existing transformations questionable when it comes to efficient implementation in real-world scenarios, and further exploration is needed to validate these claims, especially in the context of more complex architectures and practical implementations.


Another approach to give post-hoc explanations are \ac{fi} methods such as LIME~\cite{Ribeiro2016LIME} and SHAP~\cite{Lundberg2017SHAP}. Despite their popularity, these methods are not without criticism.~\cite{Molnar2022} discuss general pitfalls in applying these methods, such as incorrect context usage, poor generalization, and ignoring feature dependencies. They propose solutions for correct interpretation and highlight areas that need further research.
\textcite{Covert2021ExplainingByRemoving} propose a unified framework for model explanation using removal-based methods, including SHAP and LIME. They criticize the complexity and variability in existing methods, suggesting a theoretical foundation for improving research in explainability. 
Similarly,~\textcite{Schroeder2023a} criticize saliency methods like SHAP and LIME for failing to capture latent \ac{fi} in time series data, especially in medical domains. They recommend redesigning certain methods to improve latent saliency scoring.
Further extending this critique,~\textcite{Schroder2023b} emphasize the need for improved latent feature detection in time series classification, identifying the limitations of current saliency methods. 
In the domain of network intrusion detection,~\textcite{Tritscher2023} criticize SHAP for its challenges related to anomaly importance and the impact of replacement data, underscoring the need for rigorous evaluations.
\textcite{Sundarajan2020ManyShapleyValues} and~\textcite{Kumar2020SHAPProblems} both criticize Shapley values used in model explanations for their non-intuitive attributions and mathematical complexity, respectively. They argue that these methods require causal reasoning for accurate interpretations. 
\cite{Merrick2020ExplanationGame} point out substantial attributions to non-influential features and propose a framework for generating explanations that include confidence intervals and contrastive explanations.
In the field of natural language processing (NLP),~\textcite{Mosca2022} review SHAP-based methods, criticizing their adaptations and core assumptions, while~\textcite{vanDenBroeck2022} address the computational complexity challenges of SHAP explanations, highlighting intractability in common settings.~\textcite{Mangalathu2020} employ SHAP for seismic damage assessment but criticize existing strategies for not adequately identifying failure modes, suggesting the need for improved methods.
\textcite{slack2020fooling} demonstrate vulnerabilities in LIME and SHAP, showing how they can be manipulated to produce misleading explanations, and~\textcite{Kim2022} propose a benchmarking framework to evaluate saliency methods, calling for improvements based on ground-truth model reasoning.
These studies collectively provide a critical view of SHAP and LIME, emphasizing their limitations and suggesting areas for improvement and future research in model explainability for \ac{fi}.

One fundamental issue of the \ac{fi} methods above is the difficulty in establishing commonly agreed properties and related evaluation metrics for \ac{xai}.
Here,~\textcite{Nauta2022} provide a list of 12 so-called "Co-Properties" and corresponding evaluation metrics, often more than one metric per property.
They note that some of the properties are in conflict with each other, e.g., Completeness and Compactness of explanations.
Two properties are especially relevant in the context of this work:
Correctness, which denotes whether an explanation agrees with the model it is created for, and Consistency, which denotes whether an \ac{xai} method is deterministic and implementation-invariant, resulting in the same results for the same data sample and model. 
For Correctness, it has been observed that various evaluation metrics do not consistently agree, leading to conflicting results regarding which XAI method is accurate. Consequently, this inconsistency makes it difficult to obtain easily interpretable or reliable results~\cite{tomsett2019sanityformetrics}.
For Consistency and especially implementation-invariance, different tests were proposed~\cite{adebayo2018sanityChecks} but also criticized~\cite{yona2021sanityChecksRevisit}, resulting in the conclusion ``that saliency map evaluation beyond ad-hoc visual examination remains a fundamental challenge''.
Our \acf{rentt} and its \ac{fi} remedy this in two ways: \Ac{rentt} produces two functionally equivalent models, the \ac{nn} and the \ac{dt}, which can be used to test whether \ac{fi} methods provide the same \ac{fi} for both, i.e., whether the \ac{fi} methods are implementation-invariant.
Additionally, \ac{rentt-fi} provides explanations that are correct by design---as well as deterministic for a given \ac{nn}---as they are directly based on the \ac{nn} weights, allowing to compare the other \ac{fi} methods to ``ground-truth'' explanations.
Note that ``importance'' can be interpreted differently, thus such a comparison may also be skewed towards one's favored interpretation of this concept~\cite{Sundarajan2020ManyShapleyValues,Merrick2020ExplanationGame}.
For further details see Chapters~\ref{sec:fi-definition} and~\ref{sec:FIexperiments}.


%% file: Chapter/3_Theory.tex
\section{Transformation Theory}
\label{sec:Theory}
In this section we provide a rigorous, step-by-step derivation of the underlying transformation theory. It describes how to transform an \ac{nn} into an equivalent \ac{dt}.

\subsection{Linearizing Fully Connected Neural Networks}
\label{subsec:linearization}
Let us start by introducing the notation for formulating an \ac{fcnn}. 
\begin{definition}[Fully Connected Neural Networks]
\label{def:fcnn}
We assume that the network has an input layer, indexed with 0,  $\NumLayers-1\in \N$ hidden layers, indexed with $i = 1,\ldots, \NumLayers-1$, and an output layer, indexed with $\NumLayers$. The dimension of each layer is $\DimLayer_i \in \N$ for $\layer=0,\ldots, \NumLayers$ with neurons indexed with $\neuron=0,\ldots,\DimLayer_i-1$. The input to the network is $\Feature_0$, and the output is $\Feature_\NumLayers$. For each layer $i$, the input is $\Feature_{i-1}$ and the output is $\Feature_i$. The set of all input feature vectors is denoted as $\mathcal{X}_0 \subset \R^{\DimLayer_0}$. Then, we define a \ac{fcnn} as 
\begin{align*}
 \FcNN \colon  \mathcal{X}_0 \to \R^{\DimLayer_\NumLayers}~, \quad \Feature_0 \mapsto \FcNN(\Feature_0)\coloneqq\FcNN_\NumLayers \circ \ldots \circ \FcNN_1(\Feature_0)~,
\end{align*}
where each mapping $\FcNN_i$ is defined via 
\begin{align}
     \FcNN_{i} \colon \R^{\DimLayer_{i-1}} \to \R^{\DimLayer_{i}}~, \quad \Feature_{i-1} \mapsto \Feature_i \coloneqq \Activation_{i}(\WeightMatrix_{i-1} \Feature_{i-1})~, \quad i = 1, \ldots, \NumLayers~.
    \label{eq:FcNNmap}
\end{align}
Here, $\WeightMatrix_{i-1} \in \mathbb{R}^{\DimLayer_i \times \DimLayer_{i-1}}$ is the weight matrix, and $\Activation_i \colon \mathbb{R}^{\DimLayer_i} \to \mathbb{R}^{\DimLayer_i}$ is the activation function for the $i$-th layer. 
\end{definition}
\begin{remark}[Bias]
    To simplify the notation, we assume that the biases are already incorporated into the weight matrices, i.e., there exists an original input feature vector  $\Tilde{\Feature}_0 \in \R^{n_0-1}$, original weight matrices $\Tilde{\WeightMatrix}_i \in \R^{(\DimLayer_{i+1}-1) \times (\DimLayer_{i}-1)}$, bias vectors $\Bias_i \in \R^{\DimLayer_i - 1}$ and original activation functions $\tilde{\Activation}_i \colon \R^{\DimLayer_i - 1} \to \R^{\DimLayer_i - 1}$ such that
\begin{align*}
     \WeightMatrix_{i} \coloneqq \begin{pmatrix} 1 & 0 \\ \Bias_i & \Tilde{\WeightMatrix}_{i} \end{pmatrix} \in \R^{\DimLayer_{i+1} \times \DimLayer_{i}}~, \quad \Feature_0 \coloneqq (1, \Tilde{\Feature}_0)^\top \in \R^{\DimLayer_i}~,\\
     \Activation_i \colon \R \times \R^{\DimLayer_i - 1}  \to \R \times \R^{\DimLayer_i - 1}~,\quad \Activation_i = ( \mathrm{id}_\R, \tilde{\Activation}_i)^\top~.
\end{align*}
The weight matrices $\WeightMatrix_i$ thus include the bias and additional constant dummy entries of $0$ and $1$. Similarly, the input feature $\Feature_0$ has one dummy entry of $1$. This setup ensures that the dummy input feature propagates through the network and is used to incorporate the bias. This bias entry is represented in the \ac{nn} for example as shown in Figure~\ref{fig:net}.
\end{remark}
In the following, we show how to linearize the \ac{fcnn}, which will then be used in the next section to build a \ac{dt}. For that, some assumptions are necessary.
\begin{assumption}[Activation Function]
\label{ass:activationfunction}
We assume, that 
\begin{enumerate}
    \item the activation functions ${\Activation}_i$ are Lipschitz continuous,
    \item the activation functions ${\Activation}_i$ act component-wise, where for a given layer, all component-wise activation functions are the same, and
    \item per layer and component, the activation functions $\tilde{\Activation}_i$ are piecewise linearly defined within $r_i \in \mathbb{N}$ activation regions that partition $\mathbb{R}$. In a monotonically increasing order, the boundary points of all linear regions, i.e., all intervals, are $\SplitRegions_i \in \R^{\LinearRegions_i - 1}$.
\end{enumerate}
From Assumption~\ref{ass:activationfunction}.1, it follows by Rademacher's theorem, see for instance~\cite{Eva2018}, that $\Activation_i$ is differentiable almost everywhere. Therefore, we will assume, without loss of generality, that the activation functions are differentiable at the points of interest.
\end{assumption}
\noindent Common activation functions such as \ac{relu}, Leaky\ac{relu}, or parametric \ac{relu} satisfy Assumption~\ref{ass:activationfunction}.
\begin{remark}[Non-piecewise-linear Activation Function]
\label{remark:non_piecewise}
If the activation function is not inherently piecewise linear, it can be approximated as such, allowing the following derivation to serve as an approximation rather than in an exact manner.
\end{remark}
However, this approximating approach may result in a significant increase in the number of linear activation regions. Consequently, this would not be feasible for the transformation into a \ac{dt}.
Therefore, the linearization of the \ac{fcnn} excludes the last layer's activation function, which typically lacks Assumption~\ref{ass:activationfunction}.3 (e.g., $\tanh$ or sigmoid function). Instead, this activation function is applied to the resulting linear representation.

For layer $i<\NumLayers$, we formulate the mapping $\FcNN_i$ acting between layers $i - 1$ and $i$ of the \ac{fcnn} as 
\begin{align}
\label{eq:linearization}
    \FcNN_i(\Feature_{i - 1}) & = \Activation_i(\WeightMatrix_{i-1} \Feature_{i-1}) \nonumber \\
    & = \underbrace{\DActivation_i(\WeightMatrix_{i-1} \Feature_{i-1})}_{
        \eqqcolon\, \Slope_i\, \in\, \R^{\DimLayer_i  \times \DimLayer_i}
    } \WeightMatrix_{i-1} \Feature_{i-1}  + \underbrace{\Activation_i(\WeightMatrix_{i-1} \Feature_{i-1}) - \DActivation_i(\WeightMatrix_{i-1} \Feature_{i-1} ) \WeightMatrix_{i-1} \Feature_{i-1}}_{\eqqcolon\, \Intercept_i  \in\, \R^{\DimLayer_i}} \nonumber \\
    & = \SlopeIntercept_i \WeightMatrix_{i-1} \Feature_{i-1}~,
\end{align}
where $\DActivation_i$ is the derivative of $\Activation_i$. This reformulation is a linearization of the \ac{fcnn} mapping. The linear slope of activation $\Slope_i$ and the offset $\Intercept_i$ for each activation region in which $\WeightMatrix_{i-1} x_{i-1}$ lays, are represented jointly by the activation matrix $\SlopeIntercept_i  \in \R^{\DimLayer_i \times \DimLayer_i }$, which is defined by
\begin{align}
    (\SlopeIntercept_i)_{kl} = \begin{cases} (\Slope_i)_{kl} + (\Intercept_i)_k,  & l = 1~, \\
    (\Slope_i)_{kl}, & l > 1~.  \end{cases}
    \label{eq:example_linearization}
\end{align}
It exploits the block-matrix form of $\WeightMatrix_{i - 1}$ and the fact that the first entry of $\Feature_i$ is always $1$, so the leftmost column of the activation matrix includes the offset $\Intercept_i$. 
Due to Assumption~\ref{ass:activationfunction}.2 this linearization is exact. 
\begin{example}[Activation Function]
\label{ex:activationfunction}
For clarity, let us state two special cases:
\begin{enumerate}
\item In the one-dimensional case, $\Slope_i \in \mathbb{R}$ represents the slope, and $\Intercept_i \in \mathbb{R}$ denotes the $y$-intercept.
\item In case of a \ac{relu} activation functions $\Activation(x)=\max(0,x)$, we obtain a diagonal matrix $\Slope_i$ with $(\Slope_i)_{jj} \in \{0, 1\}$ and a vector $\Intercept_i = 0$.
\label{en:relu}
\end{enumerate}
\end{example}
\noindent Intuitively, the activation matrix $\SlopeIntercept_i$ contains the slope and intercept of the piecewise linear activation functions to each linear piece for each neuron at layer $i$.
In a ReLU network, as in Example~\ref{ex:activationfunction}.2, it contains either slope~$1$ or slope~$0$ for each neuron.

Overall, the \ac{fcnn} can be reformulated as
\begin{align*}
    \FcNN(\Feature_0) & = \Activation_\NumLayers\left(\WeightMatrix_{\NumLayers-1} (\SlopeIntercept_{\NumLayers-1} \ldots \WeightMatrix_1(\SlopeIntercept_1 (\WeightMatrix_0 \Feature_0)))\right) \nonumber \\ 
    &= \Activation_\NumLayers\left(\prod_{i=1}^{\NumLayers-1}(\WeightMatrix_{i} \SlopeIntercept_{i}) \WeightMatrix_0 \Feature_0\right)~. \nonumber \\
\end{align*}
\begin{remark}[Product Operator]
Throughout this paper the product operator is defined to add higher indexed elements to the left, contrary to the conventional right-sided application.
\end{remark}
To further simplify the notation in the following, let us introduce the iteratively defined effective weight matrices similar to~\cite{Ayt22}
\begin{align}
\label{eq:effectiveMatrix}
\EffectiveMatrix_0 &\coloneqq \WeightMatrix_0 \quad \in \R^{\DimLayer_1 \times \DimLayer_0}~,\\\nonumber
\EffectiveMatrix_{i} &\coloneqq (\WeightMatrix_i \SlopeIntercept_i) \EffectiveMatrix_{i-1} \quad\in \R^{\DimLayer_{i+1}\times \DimLayer_{0}}, \quad i = 1, \ldots, \NumLayers - 1~,
\end{align}
which allow to formulate the application of the \ac{fcnn} as
\begin{align}
\FcNN(\Feature_0) & =\sigma_\NumLayers(\EffectiveMatrix_{\NumLayers-1}\Feature_0)~.
\label{eq:effectiveMatrixNetwork}
\end{align}

The activation matrix $\SlopeIntercept_i$ depends on $\Feature_{i-1}$ and $\Activation_i(\WeightMatrix_{i-1} \Feature_{i-1})$  and can only be calculated once $\Feature_{i-1}$ is known. But due to the linearization, each activation region possesses an explicit formula to calculate the corresponding output, as we will explain in detail in the following. As a start, we now introduce the activation pattern of sample~$\Feature_0$.

\begin{definition}[Activation Pattern]\label{def:activation_pattern}
For an \ac{fcnn} as defined in Definition~\ref{def:fcnn} the $i$-th partial activation pattern is defined as the following mapping
\begin{align*}
    \ActivationPattern_i \colon \R^{\DimLayer_0} \to 
    \bigtimes\limits_{\tilde{i} = 1}^i \left\{0, \ldots, \LinearRegions_{\tilde{i}}\right\}^{\DimLayer_{\tilde{i}}}~,
\end{align*}
where $\bigtimes$ denotes the Cartesian product of sets and $\LinearRegions_{\tilde{i}}$ is the number of the activation regions, which are concatenated for each neuron in the layer. $\ActivationPattern_i(\Feature_0)$ represents the activation states of each neuron in the network up to layer $i$ for input $\Feature_0 \in \R^{\DimLayer_0}$. Furthermore, the notation $\ActivationPattern(\Feature_0) = \ActivationPattern_{\NumLayers-1}(\Feature_0)$ is used and simply referred to as activation pattern. The number of different activation patterns $|\ActivationPattern(\mathcal{X}_0)|$ is determined by all possible combinations of the activation region for the neurons.
\end{definition}
\noindent This definition of activation patterns is adopted from~\cite{Rag17}, to which we refer for further details.
For a better intuition of the activation pattern, an example is given in Section~\ref{sec:example_fcnn}.
Intuitively, the activation pattern captures which piece of the piecewise linear activation function is active for each neuron at layer $i$ for a given input. In a \ac{relu} network, as in Example~\ref{ex:activationfunction}.2, it encodes which neurons are ``firing'' (slope~$1$) versus ``not firing'' (slope~$0$). Inputs with identical activation patterns traverse the same sequence of linear regions and are therefore processed identically by the network, they follow the same computational path and have identical activation matrices $\SlopeIntercept_i$. This grouping allows us to partition the input space into regions where the network behaves as the same affine function.
Each input vector $\Feature_0$ determines a family of activation matrices $(\SlopeIntercept_i)_{i = 1}^{\NumLayers-1}$ and a family of partial activation patterns $(\ActivationPattern_i)_{i = 1}^{\NumLayers-1}$. 
Therefore, the input features $\Feature_0$ can be grouped into $|\ActivationPattern_i(\mathcal{X}_0)|$ regions, each having a unique activation pattern $\leftindex_m\ActivationPattern_i$ with $m=1,\ldots,|\ActivationPattern_i(\mathcal{X}_0)|$.
Consequently, we introduce the notation $\leftindex_{m}\EffectiveMatrix_{i}$ and $\leftindex_{m}\SlopeIntercept_i$ to denote that these matrices are uniquely associated to the partial activation pattern $\leftindex_{m}\ActivationPattern_i$.
The number of activation patterns $|\ActivationPattern_i(\mathcal{X}_0)|$ scales exponentially with the number of neurons and layers respectively.
For $k < i$, the activation matrices $\leftindex_{m}\SlopeIntercept_k$ can be explicitly determined based on a given partial activation pattern $\leftindex_{m}\ActivationPattern_k$ by knowing the activated region of the activation function based on the activation patter as shown in Equation~\eqref{eq:example_linearization}. 
In \ac{relu} Example~\ref{ex:activationfunction}.2, $\leftindex_{m}\SlopeIntercept_k$ consists of either $1$ or  $0$ at the diagonal depending on whether the activation pattern is $1$ or $0$. 
This explicit calculation allows specifying Equation~\eqref{eq:effectiveMatrixNetwork} via using Equation~\eqref{eq:effectiveMatrix} with the now known matrices $\leftindex_{m}\SlopeIntercept_k$ as follows
\begin{align}
    \label{eq:effectiveMatrixNetworkPattern}
    \leftindex_{m}\FcNN(\Feature_0)=  \Activation_\NumLayers\left(\leftindex_{m}\EffectiveMatrix_{\NumLayers-1}\Feature_0\right)~,
\end{align}
with $m$ indicating the input region corresponding to activation pattern $\leftindex_m\ActivationPattern$. This corresponds to a linear description of the \ac{fcnn} for each activation pattern $\leftindex_m\ActivationPattern$, so that
\begin{align*}
     \FcNN(\Feature_0)=\leftindex_m\FcNN(\Feature_0) \quad \Leftrightarrow \quad \ActivationPattern(\Feature_0) = \leftindex_m\ActivationPattern~.
\end{align*}

\subsection{Decision Tree Representation}
\label{subsec:transformation}
This \ac{fcnn} can be represented as a fixed \acf{dt}, where the activation pattern will provide a unique index for each path in the \ac{dt}, as we will argue in detail in the following.

The $\Level_{ij}$-th level of the \ac{dt} is characterised by the unique neuron $\neuron=1,\ldots,\DimLayer_i$ in layer $\layer=1,\ldots,\NumLayers-1$, such that
\begin{align}
        \Level_{ij} = j + \sum_{\Tilde{i}=1}^{i-1} (n_{\Tilde{i}}-1)~.
        \label{eq:level}
\end{align}
Each level in the \ac{dt} corresponds to one neuron in the \ac{fcnn}, with the \ac{dt} having $\sum_{\layer=1}^\NumLayers (\DimLayer_i - 1)$ levels in total. Here, $\DimLayer_i - 1$ accounts for the number of neurons excluding the dummy bias neuron. The number of nodes $e_{ij}$ at level $\Level_{ij}$ is given by
\begin{align}
    \Node_{ij} =\LinearRegions_i^j\prod^{i-1}_{\tilde{i}=1} \LinearRegions_{\tilde{i}}^{\DimLayer_{\tilde{i}}}~,
    \label{eq:node}
\end{align}
where $\LinearRegions_i$ denotes the number of linear regions of the activation function in layer $i$ (as specified in Assumption~\ref{ass:activationfunction}.3). This formula indicates that the number of nodes grows exponentially with the number of neurons and layers respectively.
For layer $i$ and neuron $j$, the decision rule at level $\Level_{ij}$ is given by
\begin{align}
     \mathbf{1} \cdot\sum_{k=1}^{\DimLayer_0} (\leftindex_{m}\EffectiveMatrix_{i-1})_{jk} (\Feature_0)_k > \SplitRegions_{i} \quad \in \R^{\LinearRegions_i - 1}~,
    \label{eq:fcc_decisionrule}
\end{align}
where $\mathbf{1}\in \R^{r_i - 1}$ denotes a vector of all ones. The decision rule involves multiplying the input features $\Feature_0$ by the $j$-th column of the effective matrix $(\leftindex_{m}\EffectiveMatrix_i)_j$, which incorporates the weights and activations of neurons from previous layers, and then sums up for all features. These values are compared in a piecewise manner with the current activation function and thus, are separated according to the activation regions. The output is interpreted as a boolean vector, with each element representing a partial activation pattern. It  corresponds to a split at a particular level of the \ac{dt}, thereby making a decision of which activation region is used. Recall that index $m$ corresponds to the activation region associated to the different activation pattern $\leftindex_{m}\ActivationPattern$ and thus, also to the partial activation pattern $\ActivationPattern_{i-1}$. Therefore, each index $m$ provides a unique path in the \ac{dt}. And so, each unique branch determines one unique activation pattern, denoted as $\leftindex_m\ActivationPattern_i$.
Note that this decision rule is slightly more complex than for univariate \ac{dt}: Instead of using only one feature per decision node, multiple features can be used. 
As such, the resulting \ac{dt} is multivariate.

\begin{remark}[Multi-Way Tree]
For $\LinearRegions_i = 2$, the tree is binary and for $\LinearRegions_i > 2$, it becomes a multi-way tree, where nodes split into more than two branches. 
\end{remark}
Each leaf node in the \ac{dt} gives the linear equation for the \ac{nn} mapping defined in Equation~\eqref{eq:effectiveMatrixNetworkPattern} for the corresponding unique activation pattern $\leftindex_m\ActivationPattern$ without the last activation function. This often non-piecewise activation function can be applied afterwards because it is not feasible for the transformation into a \ac{dt} as stated in Remark~\ref{remark:non_piecewise}.



\subsection{Example: Transformation of Fully Connected Neural Network into Decision Tree}
\label{sec:example_fcnn}
\begin{figure}[b]
    \centering
    \includegraphics[scale=0.7]{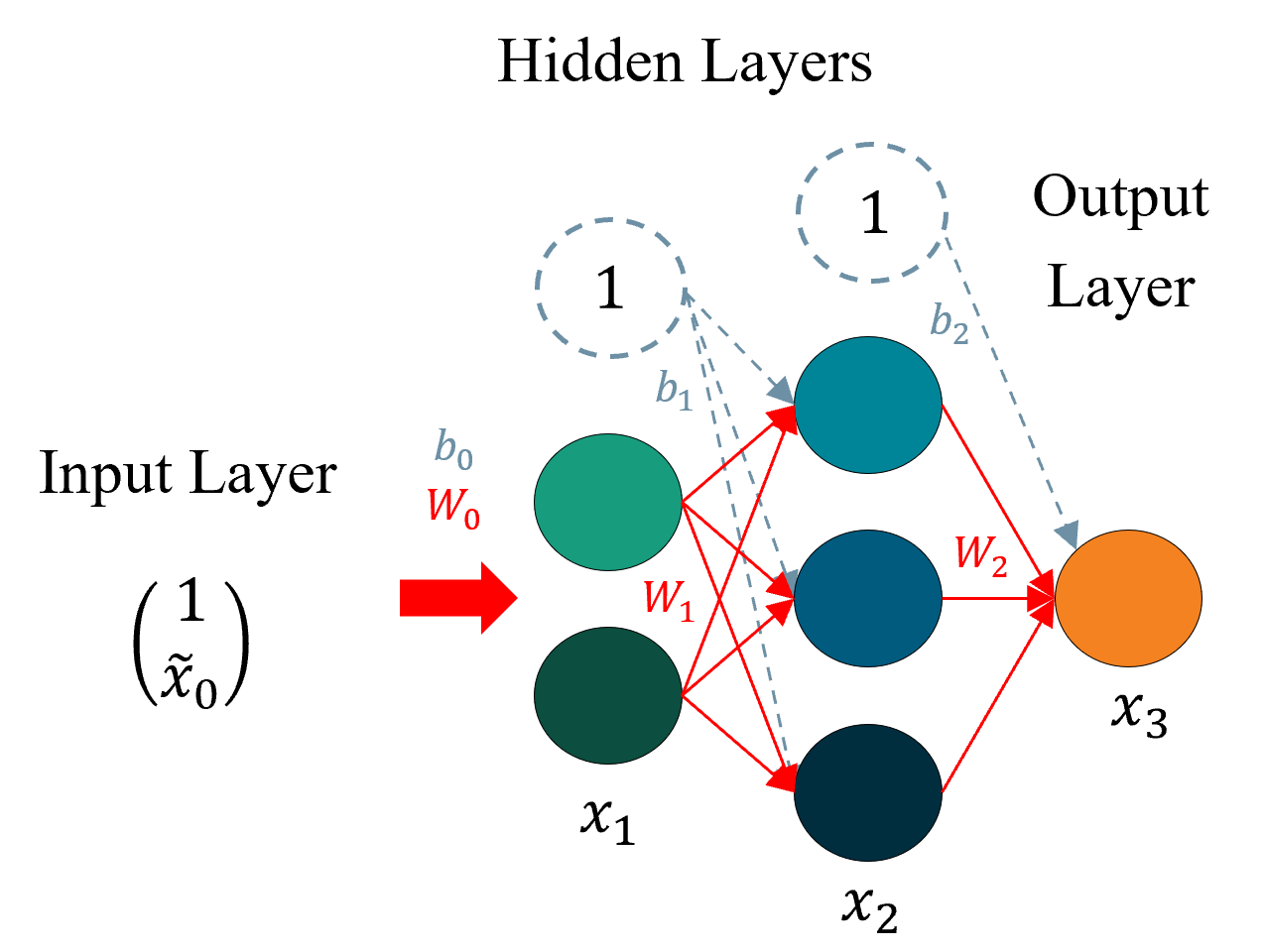}
    \caption{\Acl{fcnn} with two hidden layers of size $[2,3]$ and \ac{relu} activation. The bias is depicted as neurons with dashed lines and appended as a 1 to the input.}
    \label{fig:net}
\end{figure}
To illustrate the transformation of an \ac{fcnn} into a \ac{dt}, consider the following example. The network consists of:
\begin{itemize}
    \item Input Layer: Size $\tilde{\DimLayer}_0$ with \ac{relu} activation.
    \item Hidden Layers: first layer $\tilde{\DimLayer}_1 = 2$, second layer $\tilde{\DimLayer}_2 = 3$, with \ac{relu} activation each.
   \item Output Layer: Size $\tilde{\DimLayer}_3 = 1$, with tanh activation function.
\end{itemize}
The dimension of each layer $\DimLayer_i$ is incremented by $1$ to account for the bias term. This network is visualized in Figure~\ref{fig:net}.
The network's mapping is given by Equation~\eqref{eq:FcNNmap}, the activation matrix for \ac{relu} activation is given by Equation~\eqref{eq:example_linearization} and detailed in Example~\ref{ex:activationfunction}.2 in Section~\ref{subsec:linearization}. Using the effective matrices 
\begin{align*}
\EffectiveMatrix_0 &\coloneqq \WeightMatrix_0 \in \R^{3\times \DimLayer_{0}}~,\\
\EffectiveMatrix_{1} &\coloneqq (\WeightMatrix_1 \SlopeIntercept_1) \EffectiveMatrix_{0} \in \R^{4\times \DimLayer_{0}}~,\\
\EffectiveMatrix_{2} &\coloneqq (\WeightMatrix_2 \SlopeIntercept_2) \EffectiveMatrix_{1} \in \R^{1\times \DimLayer_{0}}~,
\end{align*}
and Equation~\eqref{eq:effectiveMatrixNetworkPattern}, the \ac{fcnn} can be reformulated as 
\begin{align*}
    \leftindex_m\FcNN(\Feature_0) = \text{tanh}(\leftindex_m\EffectiveMatrix_{2}\Feature_0). 
\end{align*}
 For each neuron $j$ in layer $i$, the level $\Level_{ij}$ in the DT is given by
\begin{align*}
        \Level_{ij} = j + \sum_{\Tilde{i} = 1}^{i-1} (\DimLayer_{\Tilde{i}}-1)~, \quad \neuron= \left\{ \begin{matrix}
    1,2\ &\text{if}\ \layer=1~,\\
    1,2,3\ &\text{if}\ \layer=2~,\\
    1\ &\text{if}\ \layer=3~.
\end{matrix}\right.\\
\end{align*}
Thus, the \ac{dt} has $\sum_{\layer=1}^3 (\DimLayer_i-1)=6$ levels. The number of nodes $\Node_{ij}$ at each level $\Level_{ij}$ are given by
\begin{align*}
    \Node_{ij} = 2^j\prod^{i-1}_{\tilde{i}=1} 2^{\DimLayer_{\tilde{i}}} = 2^{j + \sum_{\Tilde{i} = 1}^{i-1} \DimLayer_{\Tilde{i}}}~, 
\end{align*}
where the number of activation regions is $\LinearRegions_\layer=2$ for all layers.
The decision rule at level $\Level_{ij}$ representing layer $\layer$ and neuron $\neuron$ is given by Equation~\eqref{eq:fcc_decisionrule} with boundary point $\SplitRegions_i=0$.
The decision rules are calculated by
\begin{align}
     \sum_{k=1}^{\DimLayer_0} (\leftindex_{m}\EffectiveMatrix_{i-1})_{jk} (\Feature_0)_k > 0 \quad\text{with } i=1,2,3 
\end{align}
with activation function $\tanh$ being applied afterwards for the final output.
\begin{figure}[b]
    \centering
    \includegraphics[width=\textwidth]{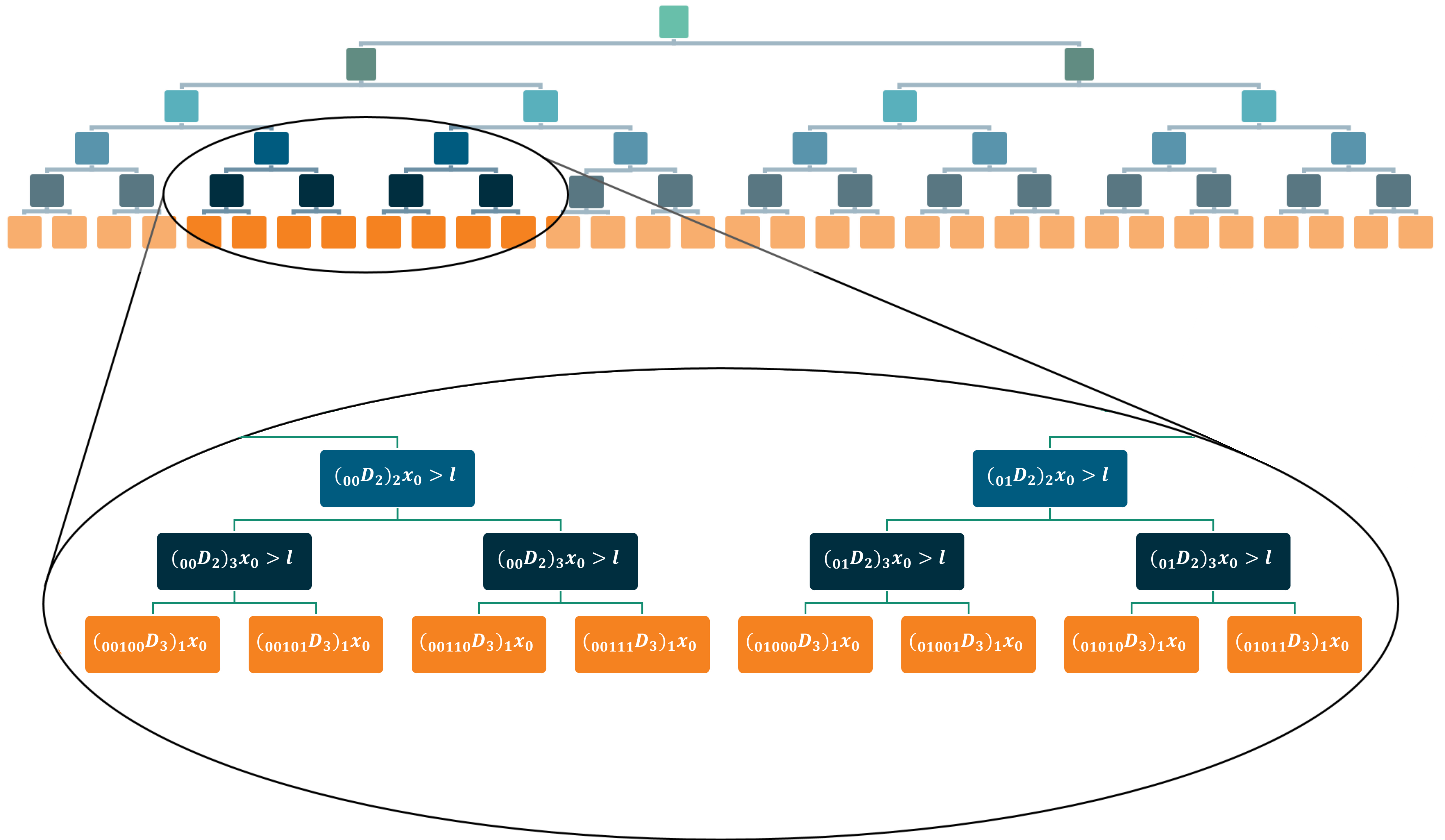}
    \caption{\ac{dt} for \ac{fcnn} with hidden layers of size $[2,3]$  and \ac{relu} activation. The colors are referring to Figure~\ref{fig:net}, i.e., green depicts the first, blue the second and orange the output layer. The different shades of the colors denote the individual neurons in each layer. Index $m$ is written in binary system for better visualization of the activation state.}
    \label{fig:dt}
\end{figure}

\noindent The second neuron in the first layer corresponds to level $\Level_{12}=2$ in the \ac{dt} with $\Node_{12}=4$ nodes. We have  the partial activation pattern
\begin{equation*}
    \ActivationPattern_1(\mathcal{X}_0)=\{(0,0),(0,1),(1,0),(1,1)\}.
\end{equation*} 
The next level $\Level_{21}=3$ with $\Node_{21}=8$ nodes has the activation pattern
\begin{equation*}
    \ActivationPattern_2(\mathcal{X}_0)=\{(0,0,0),(0,0,1),(0,1,0),(0,1,1),(1,0,0),(1,0,1),(1,1,0),(1,1,1)\}
\end{equation*}
attaching the information from the next layer's first neuron's activation. Each tuple in this pattern encodes the activation states across multiple neurons, For instance, the element $(0,1,1)$ indicates that neuron 1 of layer 1 is in region 0 (\ac{relu} slope$=0$, hence $\left[\SlopeIntercept_1\right]_1=0$), neuron 2 of layer 1 is in region 1 (\ac{relu} slope$=1$, hence $\left[\SlopeIntercept_1\right]_2=1$), and neuron 1 of layer 2 is in region 1 (\ac{relu} slope$=1$, hence $\left[\SlopeIntercept_2\right]_1=1$). Figure~\ref{fig:dt} visualizes the \ac{dt} construction, showing how each path through the \ac{dt} corresponds to a unique activation pattern and represents specific decision boundaries within the \ac{nn}.

\subsection{Extension to Convolutional Neural Networks}
The transformation described in Section~\ref{subsec:linearization} and Section~\ref{subsec:transformation} holds not only for \acp{fcnn} but can also be extended for various network architectures, for example \acp{cnn}.
In the following, we will demonstrate how to transform a \ac{cnn} into a \ac{dt} including convolutional and pooling layers, focusing primarily on max pooling layers, though other pooling types can also be used. These layers can then be combined into different \ac{cnn} architectures.

\subsubsection{Convolutional Layers}
Our linearization of a convolutional layer builds on the foundation of~\cite{Balestriero.2021}, but adopts a more implementation-oriented notation that aligns with our algorithm \ac{rentt}.
\begin{definition}[Convolutional Mapping]
Convolutional mappings $\CNN_i$ for layer $i$ are defined for input $\tilde{\Feature}_{i-1}$ with channel $\Channel_{i-1} \in \N$, height $\Height_{i-1} \in \N$ and width $\Width_{i-1} \in \N$. For each filter ${[\Psi_{i-1}]}_c \in \N$  with height $\KernelHeight_{i-1}$, width $\KernelWidth_{i-1}$ and indexed with $c=1,\ldots,\KernelChannel_i$, where $\KernelChannel_i \in \N$ is the number of filters, the mapping is given by 
\begin{align*}
     \CNN_{i} \colon & \R^{\Channel_{i-1} \times \Height_{i-1}  \times \Width_{i-1}} \to \R^{\KernelChannel_i\times\Height_{i}  \times \Width_{i}},\\
     \left[\tilde{\Feature}_{i-1}\right] &\mapsto \left[\tilde{\Feature}_i\right]_{c}\coloneqq \Activation_{i}\left(\left(\sum_{k=1}^{\Channel_{i-1}}\left[\Psi_{i-1}\right]_{c,k} * \left[\tilde{\Feature}_{i-1}\right]_{k}\right)+\left[\Bias_{i-1}\right]_{c} \right)\\
     \left[\tilde{\Feature}_i\right]_{c,w,h}  & = \Activation_{i}\left(\left(\sum_{k=1}^{\Channel_{i-1}}
     \sum_{m=1}^{\KernelHeight_{i-1}}\sum_{n=1}^{\KernelWidth_{i-1}}\left[\Psi_{i-1}\right]_{c,k,m,n}\left[\tilde{\Feature}_{i-1}\right]_{k,m+w-1,n+h-1}\right) + \left[\Bias_{i-1}\right]_{c,w,h} \right)
\end{align*}
with bias term $\Bias$ and $*$ denoting the convolution operation. The square brackets are for clearer notation. 
\end{definition}

\begin{figure}[t]
        \centering
        \includegraphics[width=0.5\textwidth]{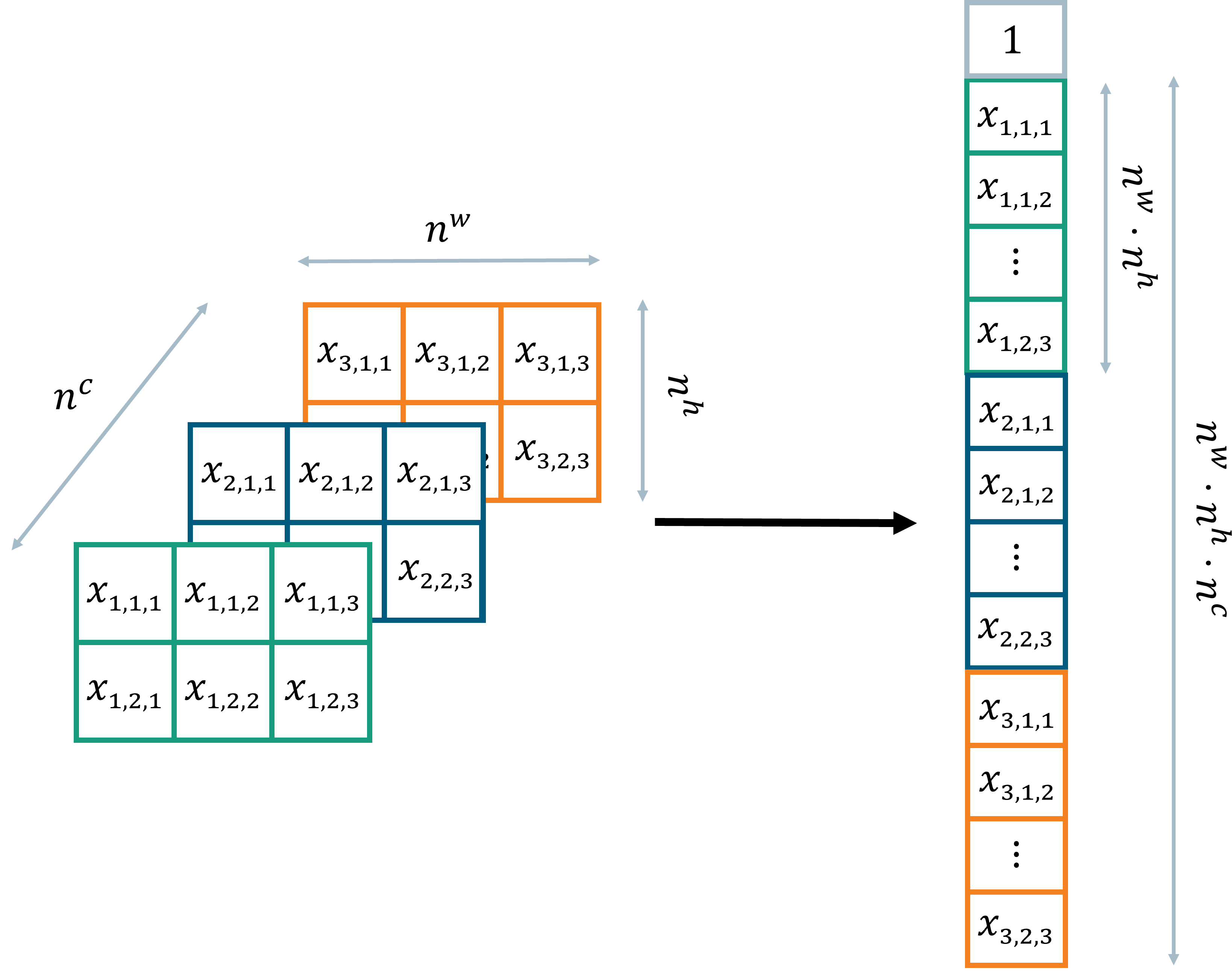}
        \caption{Flattening process of $\tilde{\Feature}_i$ with $n^c=3$, $n^w=3$ and $n^h=2$ to obtain $\Feature_i$. In the image we neglect  $\tilde{ }$ and index $i$ for better visualization.}
        \label{fig:flatteningx}
\end{figure}
\begin{figure}[t]
        \centering
        \includegraphics[width=0.8\textwidth]{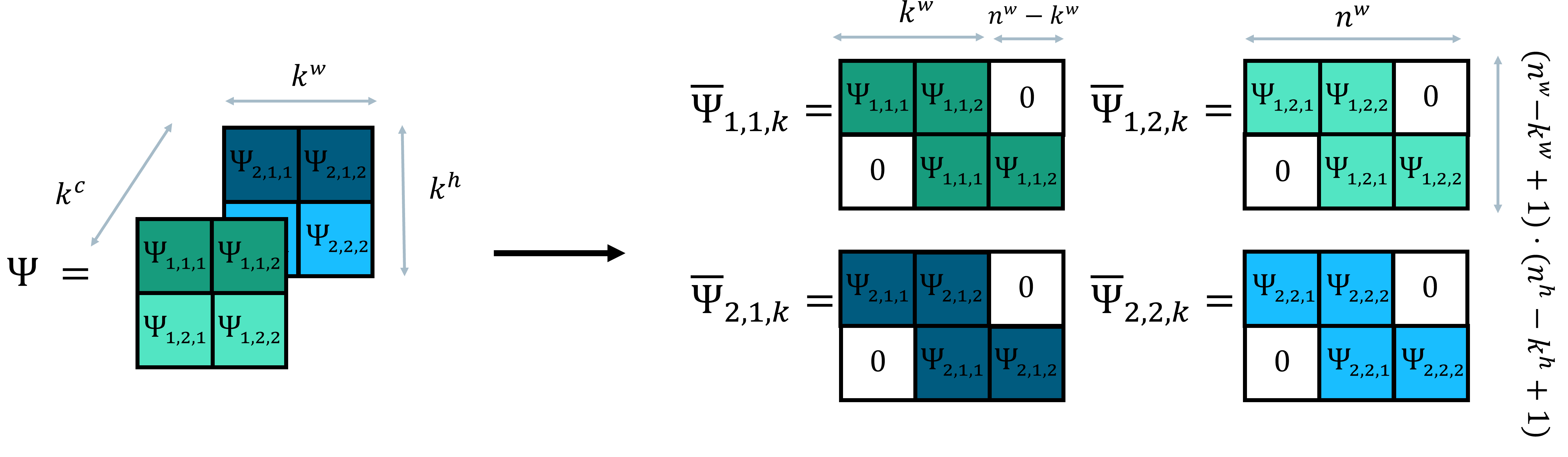}
        \caption{Flattening process of $\Psi_i$ with $k^c=2$, $k^w=2$ and $k^h=2$ when $n^w=3$ and $n^h=2$ to obtain $\left[\bar{\Psi}_{i}\right]_{c,h,k}$, where $c$ indicates the filter number, $h$ the filter height index and $k$ the channel number of $\left[\Feature_i\right]_k$. The image index $i$ is neglected for better visualization.}
        \label{fig:flatteningpsi}
\end{figure}
\begin{figure}[t]
    \centering
    \includegraphics[width=1\linewidth]{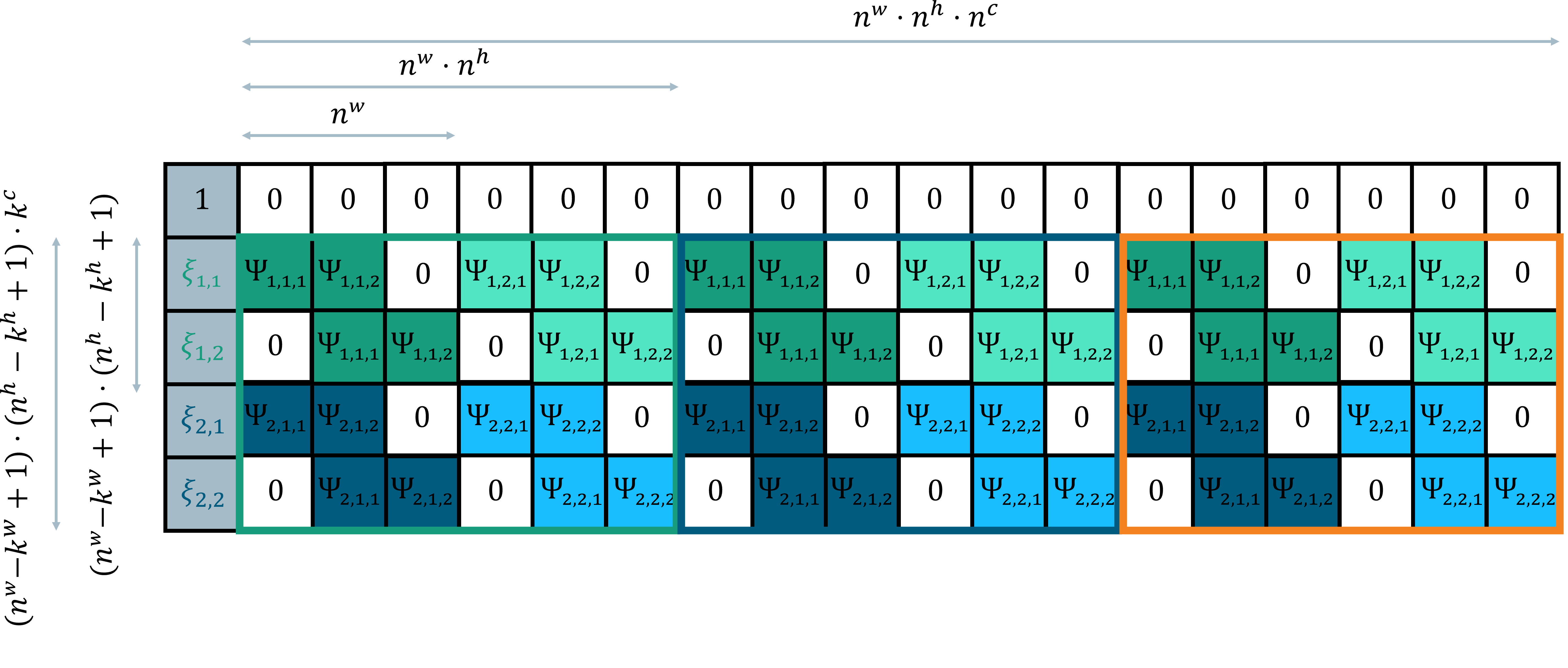}
    \caption{Super convolutional matrix $\Kernel_i$ of stacked $\left[\bar{\Psi}_{i}\right]_{c,h,k}$. Note in the image index $i$ is neglected for better visualization.}
    \label{fig:superconvolution}
\end{figure}

To express this convolution operation as matrix multiplication, the input $\tilde{\Feature}_{i-1}$ is flattened into a vector. The flattening process converts a 3D tensor into a 1D vector by sequentially processing each slice in row-major order. Starting with the first slice, elements are taken row by row, then the process repeats for subsequent slices. The elements are then concatenated into a single column vector, preserving the original order of slices and rows. Again, to include the bias, a $1$ is attached to the front to get $\Feature_{i-1}$. This whole process is shown in Figure~\ref{fig:flatteningx}. The filters $\left[{\Psi_{i-1}}\right]_c$ are flattened into matrices analogously, and stacked while filled with zeros in-between to obtain $\left[\bar{\Psi}_{i}\right]_{c,h,k}$, which represents the flattened convolution filter applied to the $k$-th channel of the input with filter height index $h$. This process is shown in Figure~\ref{fig:flatteningpsi}. These flattened filters $\left[\bar{\Psi}_{i}\right]_{c,h,k} \in \R^{(\Width_{i-1}-\KernelWidth_{i-1}+1) \cdot (\Width_{i-1}-\KernelWidth_{i-1}+1)) \times (\Width_{i-1})}$ are stacked and replicated to form the weight component of the super convolutional matrix $\Kernel$. The bias vector is obtained by replicating $\Bias_i$ on all spatial positions w.r.t. the number of filters $k^c$.
Combining both leads to the super convolutional matrix
\begin{align*}
    \Kernel_{i} = \begin{pmatrix}
        1 & 0  & \cdots & 0 & 0 & \cdots & 0 & \cdots & 0\\
        \Bias_i & \left[{\Psi}_i\right]_{1,1,1} &  \cdots & \left[{\Psi}_i\right]_{1,\Height_i,1} & \left[{\Psi}_i\right]_{1,1,2} & \cdots & \left[{\Psi}_i\right]_{1,\Height_i,2} & \cdots & \left[{\Psi}_i\right]_{1,\Height_i,\Channel_i}\\
        \vdots & \vdots  & \ddots & \vdots & \vdots  & \ddots & \vdots & \ddots & \vdots\\
        \Bias_i & \left[{\Psi}_i\right]_{\KernelChannel_i,1,1} &  \cdots & \left[{\Psi}_i\right]_{\KernelChannel_i,\Height_i,1} & \left[{\Psi}_i\right]_{\KernelChannel_i,1,2} & \cdots & \left[{\Psi}_i\right]_{\KernelChannel_i,\Height_i,2} & \cdots & \left[{\Psi}_i\right]_{\KernelChannel_i,\Height_i,\Channel_i}\\
    \end{pmatrix}~,  
\end{align*}
with $\Kernel_i \in \mathbb{R}^{n_{i+1} \times n_{i}}$, where $\DimLayer_{i} = \Channel_{i} \cdot (\Height_{i} \cdot \Width_{i})$ and $\DimLayer_{i+1} = \KernelChannel_{i+1} \cdot (\Height_{i+1} -\KernelHeight_{i+1}+1) \cdot (\Width_{i+1}-\KernelWidth_{i+1}+1))$. The super convolutional matrix reformulates the convolutional operation as a matrix multiplication, enabling us to treat convolutional layers equivalently to fully connected layers and apply the same linearization framework described in Section~\ref{subsec:linearization}. Note that $\left[{\Psi}_i\right]_{k,n,1}$, $\left[{\Psi}_i\right]_{k,n,2}, \ldots, \left[{\Psi}_i\right]_{k,n,\Channel_i}$ are identical for each $k=1,\ldots, \KernelChannel_i$ and $n=1,\ldots,\Height_i$ as indicated in Figure~\ref{fig:flatteningpsi}. This super convolutional matrix $\Kernel_i$ is also shown in Figure~\ref{fig:superconvolution}.

\noindent By using this matrix for $\Kernel_{i-1}$ and the flattened input $\Feature_{i-1}$,  the convolutional mapping can be written as
\begin{align*}
     \CNN_{i} \colon \R^{\DimLayer_{i-1}} \to \R^{\DimLayer_{i}},\quad
     \Feature_{i-1} \mapsto \Feature_i = \Activation_{i}\left(\Kernel_{i-1} \Feature_{i-1}\right)~.
\end{align*}
Analogous to the linearization of the fully connected layer in Equation~\eqref{eq:linearization}, it follows
\begin{align*}
    \CNN_{i} \colon \R^{\DimLayer_{i-1}} \to \R^{\DimLayer_{i}},\quad
     \Feature_{i-1} \mapsto \Feature_i = \SlopeIntercept_{i}\left(\Kernel_{i-1} \Feature_{i-1}\right)~.
\end{align*}
Now we can introduce the effective weight matrix analogously to Equation~\eqref{eq:effectiveMatrix}
\begin{align}
\label{eq:cnn_effectiveMatrix}
\EffectiveMatrix_0^C &\coloneqq \Kernel_0 && \in \R^{\DimLayer_1 \times \DimLayer_0}~,\\\nonumber
\EffectiveMatrix_{i}^C &\coloneqq (\Kernel_i \SlopeIntercept_i) \EffectiveMatrix_{i-1}^C && \in \R^{\DimLayer_{i+1}\times \DimLayer_{0}}~, \quad i = 1, \ldots, \NumLayers - 1~,
\end{align}
which allows to formulate the application of multiple convolutional layers as
\begin{align*}
\CNN(\Feature_0) & =\Activation_\NumLayers\left(\EffectiveMatrix_{\NumLayers-1}^C\Feature_0\right)~.
\end{align*}

\subsubsection{Max Pooling Layer}
Next, let us introduce a similar reformulation of max-pooling layers.
\begin{definition}[Max Pooling]
The max pooling mapping for layer $i$ is defined according to~\cite{Balestriero.2021} as \begin{align*}
    \maxPool_{i} \colon \R^{\DimLayer_{i-1}} \to \R^{\DimLayer_{i}}~,\quad
     \tilde{\Feature}_{i-1} \mapsto \tilde{\Feature}_i &= \left(\max_{r \in \mathcal{R}_1}[\tilde{\Feature}_{i-1}]_r, \ldots, \max_{r \in \mathcal{R}_{\DimLayer_i}}[\tilde{\Feature}_{i-1}]_r\right)^\top~,
\end{align*}
thus taking the maximum value of the input $\tilde{\Feature}_{i-1}$---without the bias dummy---in each region $\mathcal{R}_k$. 
\end{definition}
\noindent These regions $\mathcal{R}_k$ can have different cardinalities, meaning they can vary in size, and can overlap, allowing some parts of the input feature map to be included in multiple regions.

This mapping can also be formulated as a decision set, comparing each input value with the others in one filter region. Therefore, a decision matrix $\PoolingDecisionmatrix_i \in \R^{\DimRegions_{i}\cdot\DimPooling_{i}\times \DimLayer_{i}}$ is used with $\DimPooling_i=\frac{\DimFilter_{i}\cdot(\DimFilter_i-1)}{2}$, where $\DimFilter_i=\DimFilter^w_i\cdot\DimFilter^h_i$ represents the size of the pooling filter and $\DimRegions_i$ denotes the number of pooling regions. It holds $\DimLayer_{i}=\DimFilter_i\cdot\DimRegions_i+1$. Thus, $\PoolingDecisionmatrix_i$ is composed of all $\DimRegions_i$ pooling filters $\PoolingDecisionmatrixshort_{i}$ with entries in $\{-1,0,1\}$ and adding zeroes on the left for the bias handling according to
\begin{align}
    \PoolingDecisionmatrix_i &= \begin{pmatrix}
        0 & \left[\PoolingDecisionmatrixshort_1\right]_i \\
        0 &  \left[\PoolingDecisionmatrixshort_2\right]_i  \\
        \vdots & \vdots\\
        0 & \left[\PoolingDecisionmatrixshort_{\DimRegions_i}\right]_i
    \end{pmatrix} \in \R^{\DimRegions_i\cdot\DimPooling_i\times \DimLayer_{i}} \label{eq:pooling_decisionmatrix}\\[20pt]
    \text{with, e.g.,}\ \left[\PoolingDecisionmatrixshort_1\right]_i &= \begin{pmatrix}
    1 & -1 & 0 & \cdots & 0 \\
    1 & 0 & -1 & \cdots & 0\\
     & & \vdots & & \\
    1 & 0 & \cdots & 0 & -1 \\
    0 & 1 & -1 & \cdots & 0 & & 0\\
    && \vdots && \\
    0 & 1 & 0 & \cdots & -1 \\
    && \vdots && \\
    0 & \cdots & 0 & 1 & -1         
    \end{pmatrix} \in \R^{p_i\times \DimLayer_{i}-1}~,\label{eq:pooling_filter_matrix}
\end{align}
where each row contains exactly one $1$ and one $-1$, with zeros elsewhere. The $1$ selects the first input value in a pairwise comparison, while the $-1$ selects the second input value, such that each row computes the difference between two specific input values within the pooling region to determine their relative ordering.
Using this decision matrix, the decision rules are given by
\begin{align}
\label{eq:pooling_decision}
    \PoolingDecisionmatrix_{i-1}\Feature_{i-1} > \mathbf{0}~.
\end{align}
These decision rules result in a boolean vector, indicating in each entry which of the compared feature values is larger, thereby identifying the largest value in one region. The effective pooling matrix $\PoolingEffmatrix_i$ is then used to calculate the output of the pooling layer. This effective pooling matrix consists of a $1$ at the positions of the largest feature in a filter region and at the position of the bias, and zeros elsewhere, i.e.,
\begin{align}
    \PoolingEffmatrix_{i} \in \{0,1\} \in \R^{\DimRegions_i+1\times\DimLayer_{i}}~.
    \label{eq:pooling_effectiveM}
\end{align}
For the subsequent layer, the effective pooling matrix is used to incorporate the output of the pooling layer into the following decision rules and effective matrices.

\begin{figure}[H]
    \subfigure[Example for $\PoolingDecisionmatrix$  consisting of two filters $\PoolingDecisionmatrixshort_1$ and $\PoolingDecisionmatrixshort_2$, which are colored green and blue, with $d^w=2$, $d^h=2$ and $s=2$. We choose $n^w=4$, $n^h=2$, and $n^c=1$. Image index $i$ is neglected for better visualization.]{%
        \label{fig:Decision_pool}%
        \includegraphics[width=0.6\textwidth]{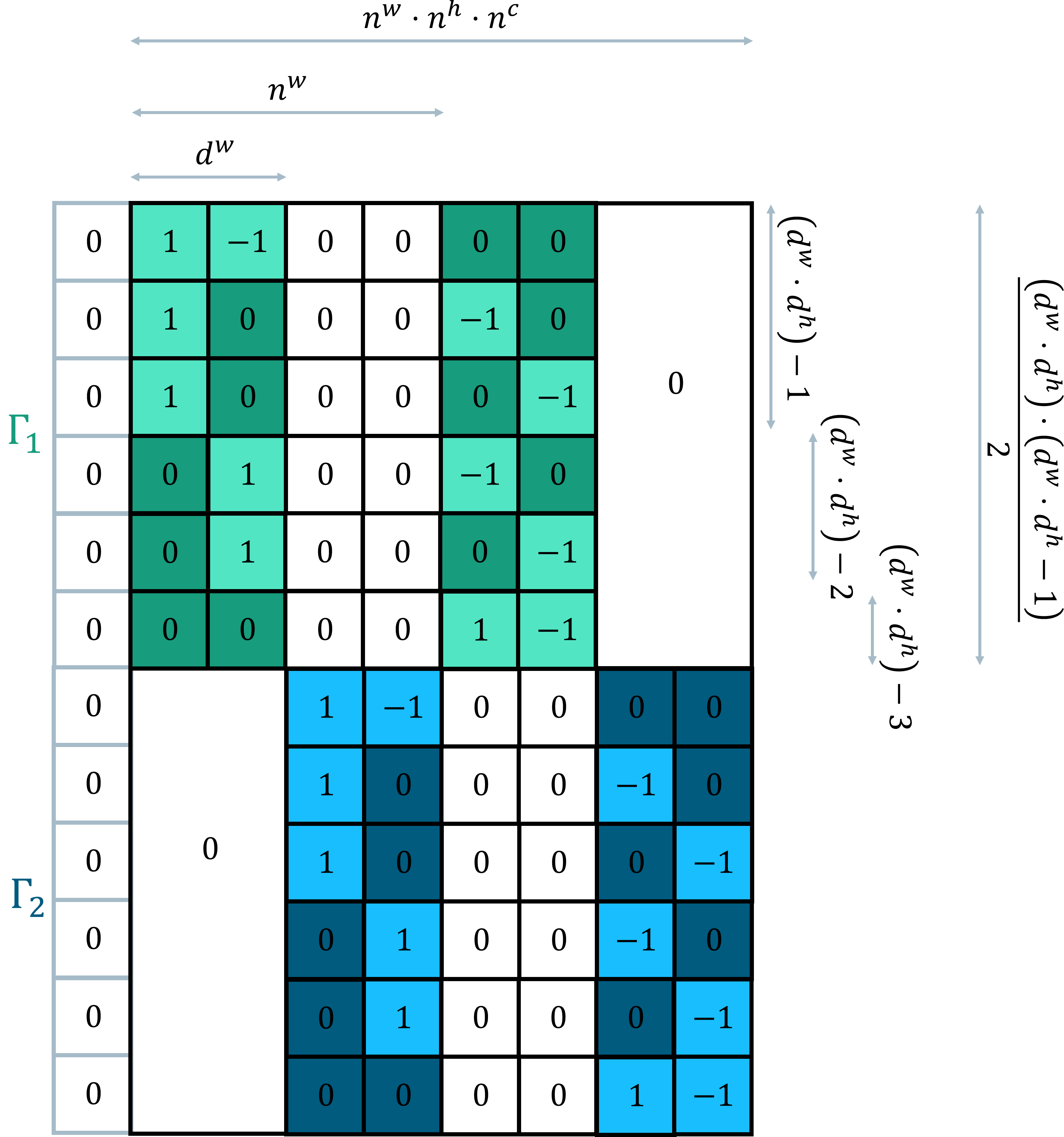}%
    }\hfill
     \subfigure[Example for resulting decision rule as in  Equation~\eqref{eq:pooling_decision}.]{%
        \label{fig:rule_pool}%
        \includegraphics[width=0.17\textwidth]{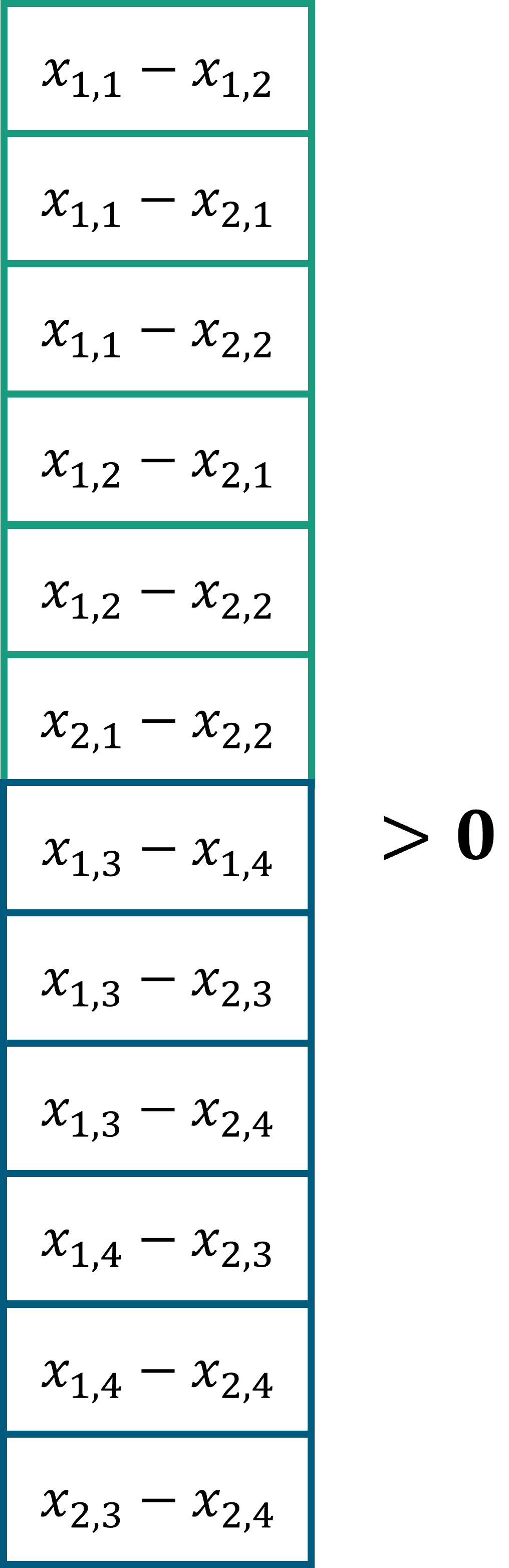}%
    }\hfill
    \subfigure[Example for flattened $\Feature$ used for max pooling with two filter regions colored green and blue.]{%
        \label{fig:flatten_pool}%
        \includegraphics[width=0.4\textwidth]{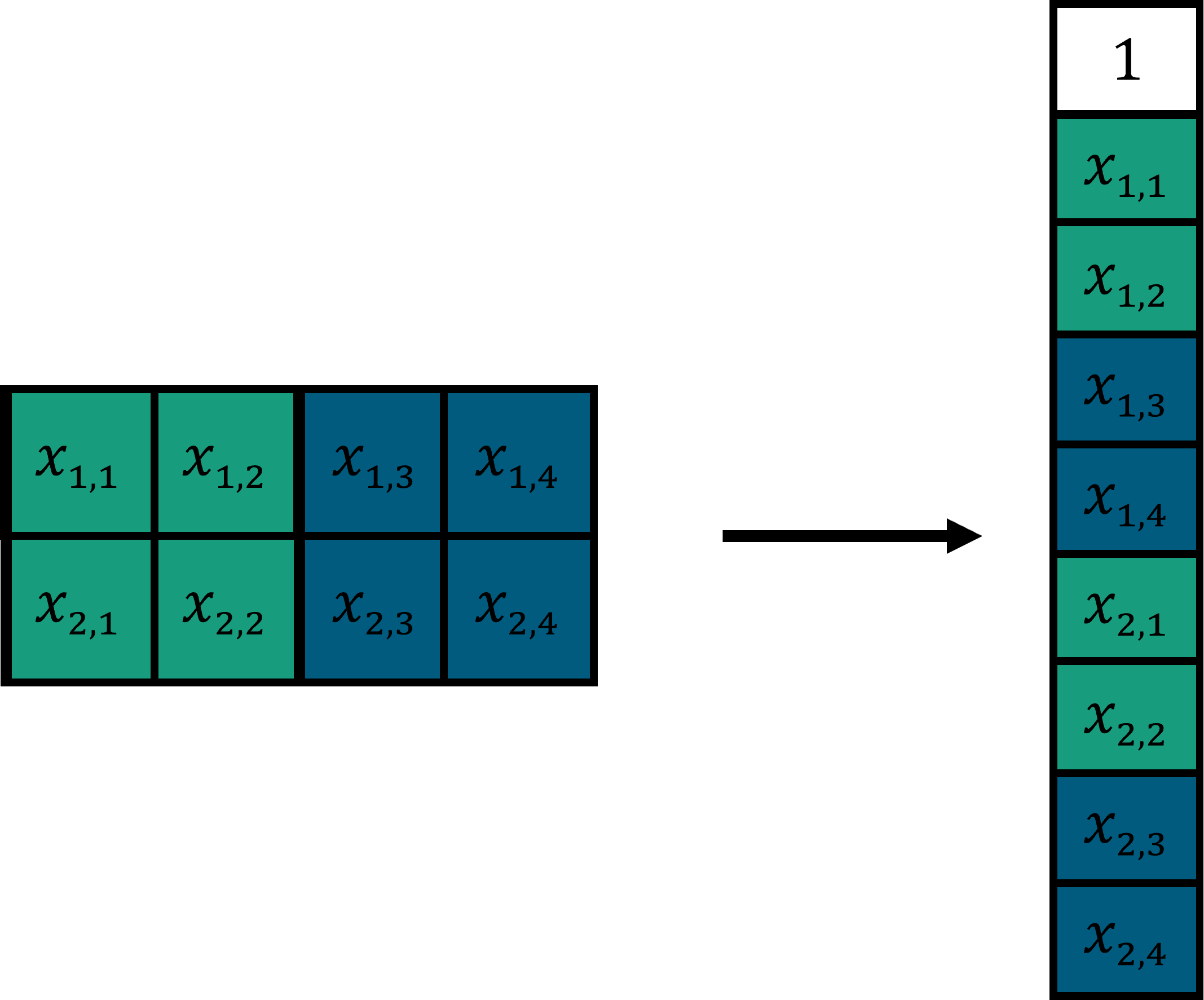}%
    }\hfill
    \subfigure[Example for resulting effective matrix $\PoolingEffmatrix$ for max pooling as in Equation~\eqref{eq:pooling_effectiveM} when $\Feature_{1,2}$ and $\Feature_{1,4}$ are largest. Note in the image index $i$ is neglected for better visualization.]{%
        \label{fig:effective_pool}%
        \includegraphics[trim=0cm 0cm 0.4cm 0cm, clip,width=0.47\textwidth]{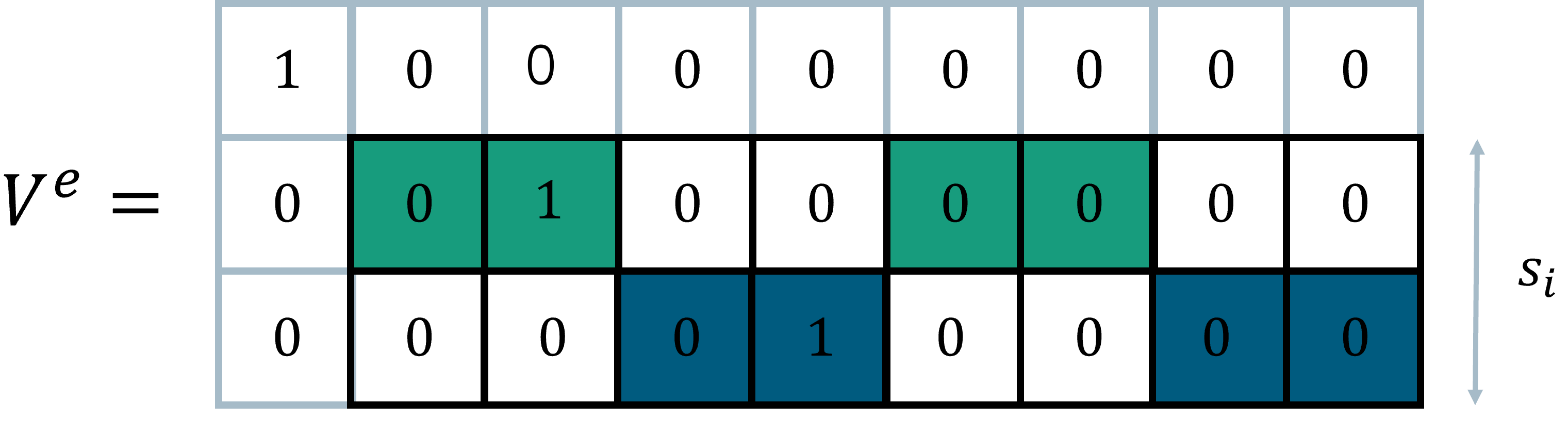}%
    }
    \caption{Example for the transformation of a max pooling layer.}
    \label{fig:max_pooling}
\end{figure}

Building on this approach, we can introduce the effective weight matrix for the pooling layers. This is analogous to Equation~\eqref{eq:effectiveMatrix} by simply omitting the activation
\begin{align}
\label{eq:pooling_effectiveweightMatrix}
\EffectiveMatrix_0^P &\coloneqq \PoolingEffmatrix_0 && \in \R^{\DimLayer_1 \times \DimLayer_0}~,\\\nonumber
\EffectiveMatrix_{i}^P &\coloneqq \PoolingEffmatrix_i  \EffectiveMatrix_{i-1}^P && \in \R^{\DimLayer_{i+1}\times \DimLayer_{0}}~, \quad i = 1, \ldots, \NumLayers - 1~.
\end{align}
An example of this transformation for $ \Feature=(1,  2,  1,  4, -1,  -4,  2,  3)^\top $ is presented in Figures~\ref{fig:max_pooling}. Figure~\ref{fig:Decision_pool} displays the decision matrix $ \PoolingDecisionmatrix $ along with the two filters $ \PoolingDecisionmatrixshort_1 $ and $ \PoolingDecisionmatrixshort_2 $. The resulting decision rules, which create the paths in the \ac{dt}, are shown in Figure~\ref{fig:rule_pool}. Figure~\ref{fig:flatten_pool} illustrates the input in both matrix form and its flattened vector version. Finally, Figure~\ref{fig:effective_pool} presents the effective matrix $ \PoolingEffmatrix $ corresponding to our example.
\subsubsection{Decision Tree Representation of Convolutional Neural Networks}
The convolutional and pooling layers can be combined and transformed into a \ac{dt} analogously to the \ac{fcnn} by choosing the effective weight matrix $\WeightMatrix_i =  \Kernel_i$ or $\WeightMatrix_i = \PoolingEffmatrix_i$ for each layer $i$ and using the decision rule of Equation~\eqref{eq:fcc_decisionrule} for a convolutional layer similar to the \ac{fcnn} or substitute it for a pooling layer via Equation~\eqref{eq:pooling_decision}. For convolutional layers the depth of the \ac{dt} is determined as for fully connected layers, defined in Equation~\eqref{eq:level}, replacing the number of nodes with 
\begin{equation*}
    \DimLayer_i\coloneqq\KernelChannel_{i} \cdot (\Height_{i} -\KernelHeight_{i}+1) \cdot (\Width_{i}-\KernelWidth_{i}+1))~.
\end{equation*}
For maxpooling, the number of nodes is replaced by 
\begin{equation*}\DimLayer_i\coloneqq\DimPooling_i=\frac{\DimFilter_i\cdot(\DimFilter_i-1)}{2}\DimRegions_i \quad.
\end{equation*}
\begin{remark}[Number of Nodes]
When containing only the information about the maximum values instead of the entire ranking of the feature values, the number of nodes can also be given by
\begin{equation*}
\DimLayer_i\coloneqq\DimRegions_i\cdot\DimFilter_i~.
\end{equation*}
\end{remark}
These types of layers can be combined for different network architectures, for example, to a typical \ac{cnn}
\begin{align*}
    \CNN\colon \R^{\DimLayer_0} \to \R^{\DimLayer_\NumLayers}, \quad \Feature_0 \mapsto \CNN(\Feature_0)\coloneqq\FcNN_\NumLayers \circ \maxPool_{\NumLayers-1}\circ \CNN_{\NumLayers-2} \circ \ldots \circ \maxPool_2\circ \CNN_1(\Feature_0)
\end{align*}
comprising an alternating sequence of convolutional and pooling layer as well as one fully connected output layer. 
For this network, the previously shown decision rules and effective matrices just have to be combined, so that a \ac{dt} similar to the one for the \ac{fcnn} is created.

\subsection{Extension to Recurrent Neural Networks}
The \acp{rnn} were introduced by~\cite{Rumelhart1986LearningRB} allowing backward connections where the output of one layer at time $t$ is fed back as input to the network at the next time step $t+1$. This introduced an internal state to the \ac{nn} and is beneficial for processing sequential data like time series or text. 
\begin{definition}[Recurrent layers]
    Following~\cite{Goodfellow2016} the mapping of a recurrent layer with size $\DimLayer_i$ is defined as 
\begin{align}
\label{eq:RNN}
     \RNN_{i} &\colon \R^{\DimLayer_{i-1}+\DimHidden_{i}} \to \R^{\DimLayer_{i}}~, \nonumber \\[10pt]
     \Feature_{i-1}^{(t)} \mapsto \Feature_i^{(t)} &\coloneqq \hiddenState_{i}^{(t)} =\Activation_{i}^{(t)}(\WeightMatrixU_{i}\hiddenState_{i}^{(t-1)} + \WeightMatrixV_{i-1} \Feature_{i-1}^{(t)})~,\quad i = 1, \ldots, \NumLayers-1~.
\end{align}
Here, $\hiddenState_i^{(t)}$ denotes the hidden state of layer $i$  at time step $t=1,\ldots,T$, with arbitrary initial hidden state $h_i^{(0)} \in \R^{\DimHidden_i}$, where $\DimHidden_i$ is the dimension of the hidden state. The recurrent weight matrix $\WeightMatrixU_i \in \R^{\DimHidden_i \times \DimHidden_{i}}$ connects the previous hidden state $\hiddenState_{i}^{(t-1)}$ to the current time step, while the input weight matrix $\WeightMatrixV_{i-1} \in \R^{\DimLayer_i \times \DimLayer_{i-1}}$ processes the input features $\Feature_{i-1}^{(t)}$ from the previous layer. We assume that the layer output directly corresponds to the hidden state, which implies $\DimHidden_i = \DimLayer_i$.
\end{definition}
\begin{figure}[H]
\subfigure[Compact version]{\label{fig:RNN_compact}\includegraphics[width=0.49\textwidth]{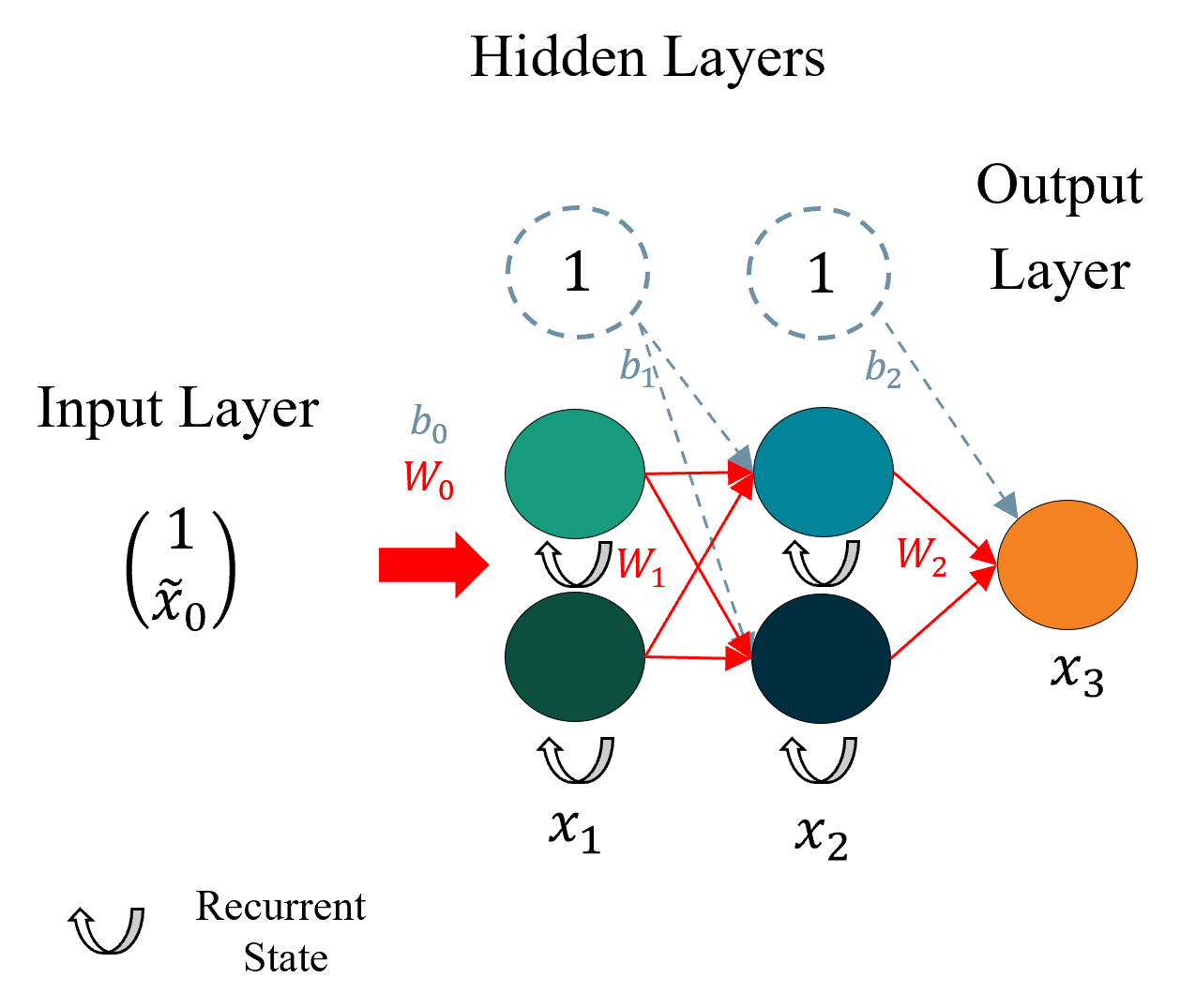}}
\hfill
\subfigure[Unfold version]{\label{fig:RNN_unfold}\includegraphics[width=0.49\textwidth]{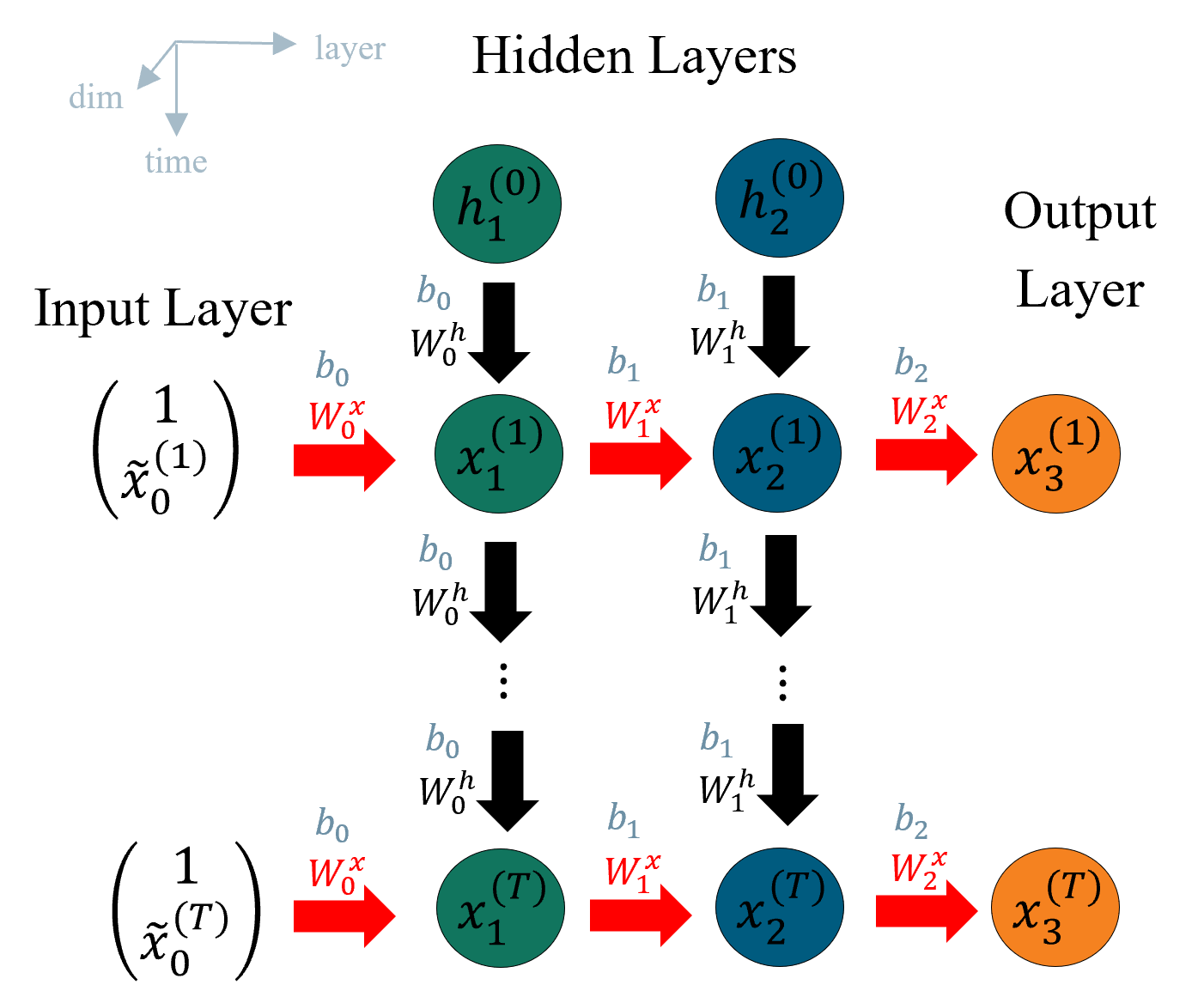}}
\caption{\Acl{rnn} with two hidden layers and two neurons each. The dotted circles and $B$ indicating the bias consideration. (a) Classical view with arrows indicating the recurrent neurons. (b) Unfolded network showing the recurrent states downwards.}
\label{fig:RNN}
\end{figure}
\noindent An example \ac{rnn} is shown in Figure~\ref{fig:RNN} in a compact and an unfolded view.
The bias term is taken into account as in the sections above by adding a $1$ to the initial feature vector $\Feature^{(t)}_0$ and the initial hidden state $\hiddenState^{(0)}_i$  and by adding an identity function to the activation function $\Activation_i$. Analogously, the weight matrix is expanded by $1$ at the upper left and filled with $0$.
Both the hidden state $\hiddenState^{(t)}_i$ and the feature vector $\Feature^{(t)}_i$ change over time, but the weight matrices $\WeightMatrixU_i$ and $\WeightMatrixV_i$ remain constant over time, though they differ based on whether they are applied to the hidden state vector or the feature vector.
The output of the recurrent layer can be rewritten by linearizing the activation function $\Activation_i$ analogous to the \ac{fcnn}
\begin{align*}
    \Feature_{i}^{(t)} &= \hiddenState_{i}^{(t)} = \SlopeIntercept_{i}^{(t)}( \SlopeIntercept_{i}^{(t)} (\WeightMatrixV_{i-1} \Feature_{i-1}^{(t)}+\WeightMatrixU_{i} \hiddenState_{i}^{(t-1)}))
\end{align*}
with activation matrix $\SlopeIntercept_{i}^{(t)} \in \R^{\DimLayer_i\times \DimLayer_i}$.
This can be formulated in dependence of the initial hidden state $\hiddenState_i^{(0)}$ at time $t=0$ using the notation ${\Feature_i=\left(\Feature_i^{(1)},\ldots,\Feature_i^{(T)}\right)^\top}\in\R^{\DimLayer_i\cdot T}$, ${\hiddenState_i=\left(\hiddenState_i^{(0)},\ldots,\hiddenState_i^{(0)}\right)^\top}\in\R^{\DimLayer_i\cdot T}$and $\SlopeIntercept_i=(\SlopeIntercept_i^{(1)},\ldots,\SlopeIntercept_i^{(T)})\in \R^{\DimLayer_i\times \DimLayer_i\cdot T}$ as
\begin{align*}
    \Feature_i = \SlopeIntercept_{i} & \left(\begin{pmatrix}
1 & &  &  {\multirow{2}{*}{\text{\Large{\textbf{0}}}}} &\\
 \prod_{k=1}^{1} \WeightMatrixU_{i} \SlopeIntercept_{i}^{(k)}   &\ddots &  & & \\
 \vdots  & &  & &\\
\prod_{k=1}^{T-1} \WeightMatrixU_{i} \SlopeIntercept_{i}^{(k)}   & \prod_{k=2}^{T-1} \WeightMatrixU_{i} \SlopeIntercept_{i}^{(k)} & \cdots &  1&
\end{pmatrix} \begin{pmatrix}
    \WeightMatrixV_{i-1} &  & \text{\Large{\textbf{0}}} \\
     & \ddots &  \\
    \text{\Large{\textbf{0}}} &  & \WeightMatrixV_{i-1}
\end{pmatrix}\Feature_{i-1}  \right. \\[10pt] \nonumber 
& \left. + \begin{pmatrix}
    1 &  & \text{\Large{\textbf{0}}}\\
     & \ddots & \\ 
    \text{\Large{\textbf{0}}} &  & \prod_{k=1}^{T-1}\WeightMatrixU_{i} \SlopeIntercept_{i}^{(k)}
\end{pmatrix}\begin{pmatrix}
    \WeightMatrixU_{i} &  & \text{\Large{\textbf{0}}} \\
     & \ddots &  \\
    \text{\Large{\textbf{0}}} &  & \WeightMatrixU_{i}
\end{pmatrix} \hiddenState_i \right).
\end{align*}
\begin{remark}(Product Operator)
When $k>T-1$, we consider the product operator to be $1$. 
\end{remark}
To simplify the subsequent computations and expressions we make the following assumption.
\begin{assumption}[Hidden State Initialization]
\label{ass:hiddenstate}
    For simplicity we assume that $\hiddenState_i^{(0)}$ is initialized as zero.
\end{assumption}
\noindent Therefore, the second part of the equation involving $\hiddenState_i^{(0)}$ can be omitted in the following. 
We define the effective weight matrices as
\begin{align*}
    \EffectiveMatrixH_{t,p,i}&=1 \quad && \text{with } p=t~,\\
    \EffectiveMatrixH_{t,p,i} &= \prod_{k=p}^{t-1}\WeightMatrixU_{i}\SlopeIntercept_{i}^{(k)} \quad && \text{with } p=t-1,\ldots,1~.\\
\end{align*}
Using these simplifies the previous notation to
\begin{align*}
   \Feature_i =\SlopeIntercept_i \left( \begin{pmatrix}
     \EffectiveMatrixH_{1,1,i} &  & \text{\Large{\textbf{0}}} \\
    \vdots & \ddots &  \\
    \EffectiveMatrixH_{T,1,i} & \cdots &  \EffectiveMatrixH_{T,T,i}
\end{pmatrix} \begin{pmatrix}
    \WeightMatrixV_{i-1} &  & \text{\Large{\textbf{0}}} \\
     & \ddots &  \\
    \text{\Large{\textbf{0}}} &  & \WeightMatrixV_{i-1}
\end{pmatrix}\Feature_{i-1} \right)~, 
\end{align*}
which can be combined into
\begin{align*}
& \MatrixR_{i-1}=\begin{pmatrix}
     \EffectiveMatrixH_{1,1,i}\WeightMatrixV_{i-1} &  & \text{\Large{\textbf{0}}} \\
    \vdots & \ddots &  \\
    \EffectiveMatrixH_{T,1,i}\WeightMatrixV_{i-1} & \cdots &  \EffectiveMatrixH_{T,T,i}\WeightMatrixV_{i-1}
\end{pmatrix}\in \R^{\DimLayer_i\cdot T\times \DimLayer_{i-1}\cdot T}
\end{align*}
Introducing the overall effective weight matrices
\begin{align}
\label{eq:rnn_effectivematrix}
    \EffectiveMatrixR_0 &\coloneqq \MatrixR_0 && \in \R^{\DimLayer_{1}\cdot T \times \DimLayer_0\cdot T}~,\\ \nonumber
    \EffectiveMatrixR_i &\coloneqq \MatrixR_i \EffectiveMatrixR_{i-1} && \in \R^{\DimLayer_{i+1}\cdot T \times \DimLayer_0\cdot T}~,\quad i=1,\ldots,\NumLayers-1~,
\end{align}
we can formulate the mapping of a \ac{rnn} with $\NumLayers$ layers similar to the \ac{fcnn}
\begin{align*}
    \Feature_i = \SlopeIntercept_i \EffectiveMatrixR_i\Feature_{0}~.
\end{align*}
\begin{remark}[Including Initial Hidden State]
If Assumption~\ref{ass:hiddenstate} does not hold, the derivation described above still holds when introducing a combined expression 
     \begin{align*}
     & \Feature^{h}_{i-1}= \begin{pmatrix}
        \hiddenState_i\\
        \Feature_{i-1}
    \end{pmatrix} \in \R^{\DimLayer_{i}\cdot T+\DimLayer_{i-1}\cdot T} 
    \end{align*} and 
    \begin{align*}
    \MatrixR_i = \begin{pmatrix} \mathbf{1} & 0 \\
        \MatrixRH_{i} & \MatrixRX_{i-1}
    \end{pmatrix}\in \R^{2\cdot\DimLayer_i\cdot T\times \DimLayer_{i}\cdot T+\DimLayer_{i-1}\cdot T}
\end{align*}
with $\MatrixRX_{i-1}$ being the previous used $\MatrixR_{i-1}$ and
\begin{align*}
   \MatrixRH_{i} = \begin{pmatrix}
     \EffectiveMatrixH_{1,1,i}\WeightMatrixU_{i-1} &  & \text{\Large{\textbf{0}}} \\
      & \ddots &  \\
   \text{\Large{\textbf{0}}} &  &  \EffectiveMatrixH_{T,1,i}\WeightMatrixU_{i-1}
\end{pmatrix}\in \R^{\DimLayer_i\cdot T\times \DimLayer_{i}\cdot T}~.
\end{align*}
This modification includes the contributions of both components---feature vector $\Feature$ and hidden state $\hiddenState$---within the equation.
\end{remark}

\subsubsection{Decision Tree Representation of Recurrent Neural Networks}
Analogous to the \ac{fcnn} we can build a \ac{dt} of the \ac{rnn} where every level in the \ac{dt} represents one neuron in a layer of the \ac{nn} at one time step $t=1,\ldots,T$. Thus, one decision rule in a node gives the activation state of one neuron at one time step. 
The depth of the \ac{dt} for an \ac{rnn} is the same as for a \ac{fcnn} multiplied by the number of time steps indicated by the additional dimension $T$ in the previously derived formulas in comparison to the \ac{fcnn}, i.e.
\begin{align}
        \Level_{ijt} = j + \sum_{\Tilde{i}=1}^{i-1} \sum_{\Tilde{t}=1}^{t}  (n_{\Tilde{i}}-1)~.
        \label{eq:RNNlevel}
\end{align}
The number of nodes in each level is again given by Equation~\eqref{eq:node}.

%% file: Chapter/4_Algorithm.tex
\section{Runtime Efficient Network to Tree Transformation (RENTT) Algorithm}
\label{sec:Algorithms}
While the theoretical foundations for transforming \acp{nn} into \acp{dt} have been established, direct implementation faces significant computational challenges. A complete transformation would generate a very large amout of nodes as shown in Equation~\eqref{eq:node}, rendering the approach intractable for practical applications. This section presents the \acf{rentt} algorithm, which achieves computational feasibility through sparse tree construction based on empirically observed activation patterns. By building only necessary decision paths, RENTT maintains mathematical equivalence while significantly reducing computational complexity.

\subsection{Implementation}
From an implementation perspective, the transformed decision tree is represented as a collection of node objects, each encapsulating the essential properties required for efficient traversal and prediction. Each node maintains:
\begin{itemize}
    \item Unique identifier (ID) following a binary tree numbering scheme where child nodes are assigned.
    \item Activation pattern encoding the path of neuron activations from root to current node.
    \item Effective weight matrix containing the effective linear weights and bias for decision-making.
    \item Sample memory storing training data points that flow through the node during construction.
\end{itemize}

The \ac{rentt} algorithm builds a sparse \ac{dt} using ante-hoc pruning. This method builds only the relevant paths of the \ac{dt}, selected by the occurring activation patterns obtained from the training data set $\mathcal{X}_0$. The implementation is shown in Algorithm~\ref{alg:transformation}. The process begins with the identification of activation patterns of the training samples. These patterns serve as a guide for pruning the \ac{dt}, enabling us to build only branches corresponding to used activation patterns. This ensures we retain solely those nodes that process at least one sample and thus contribute to the model's accuracy or efficiency. This selective approach not only conserves memory but also accelerates the transformation of the \ac{nn} into a \ac{dt} representation. The \ac{dt} can still be appended with additional necessary paths corresponding to new activation patterns not included in the training set. 

Our implementation of Algorithm~\ref{alg:transformation} is developed in Python, utilizing TensorFlow-based \acp{nn} to ensure comparability with the algorithm presented by~\cite{Ngu20}. It accommodates \acp{nn} for both classification (binary and multiclass) and regression tasks. The exact details of the implementation can be found in our code on GitHub\footnote{\url{https://github.com/HelenaM23/RENTT}}.

Regarding the applicability to \acp{cnn} and \acp{rnn}, we note the following.
\begin{remark}[\Ac{cnn} and \Ac{rnn} Implementation]
\label{remark:implementation}
Algorithm~\ref{alg:transformation} can be adapted for \acp{cnn} and \acp{rnn} through these adjustments:
\begin{enumerate}
    \item To transform \acp{cnn}, replace the set of weight matrices $\mathcal{W}$ with the set of filters $\mathcal{\Psi}=\{\Psi_0,\cdots,\Psi_{\NumLayers-1}\}$ and substitute  Equation~\eqref{eq:effectiveMatrix} with equations~\eqref{eq:cnn_effectiveMatrix} or~\eqref{eq:pooling_effectiveweightMatrix}. 
    \item To transform \acp{rnn}, additionally incorporate the set of hidden state weight matrices $\mathcal{W}^h=\{\WeightMatrixU_0,\cdots,\WeightMatrixU_{\NumLayers-1}\}$ and substitute Equation~\eqref{eq:effectiveMatrix} with Equation~\eqref{eq:rnn_effectivematrix}.
\end{enumerate}
\end{remark}

\begin{algorithm}[H]
    \caption{\acl{rentt} (RENTT)}\label{alg:transformation}    \raggedright \hspace*{\algorithmicindent}\textbf{Input:} Sets of input feature vectors $\mathcal{X}_0$, set of weight matrices $\mathcal{W}=\{\WeightMatrix_0, \ldots, \WeightMatrix_{\NumLayers - 1}\}$, and
    \\ \hspace*{\algorithmicindent} set of activation functions $\mathcal{A} = \{\Activation_{1}, \ldots, \Activation_{\NumLayers}\}$\\
    \hspace*{\algorithmicindent}\textbf{Output:} \ac{dt} $T$
        \begin{algorithmic}[1]
            \State Set $\EffectiveMatrix_0$ using Equation~\eqref{eq:effectiveMatrix}
            \State Create root node $q_0$ in \ac{dt} $T$ with ID, decision rule and samples in this node

            \For{$i=1,\ldots, \NumLayers- 1$}
                \State Initialize set of unique partial activation patterns $\mathcal{P}_i = \emptyset$ 
                \State Initialize set of effective matrices $\mathcal{D}_i = \emptyset$
                \For{$\Feature_0$ in $\mathcal{X}_0$} \Comment{Calculate used activation patterns}
                    \If{$i = 1$}
                        \State Calculate partial activation pattern $\ActivationPattern_{i}(\Feature_0)$
                    \Else 
                        \State Reuse last partial activation pattern $\ActivationPattern_{i-1}(\Feature_0)$ (see Definition~\ref{def:activation_pattern})
                    \EndIf
                    \State\parbox[t]{\dimexpr\linewidth-6em} {$\mathcal{P}_i \gets \mathcal{P}_i \cup \{\ActivationPattern_i(\Feature_0)\}$}
                \EndFor
                \For{$\leftindex_m\ActivationPattern_i$ in $\mathcal{P}_i$} \Comment{Pruning: only nodes for occupied activation patterns}
                    \State \parbox[t]{\dimexpr\linewidth-6em}{Calculate effective weight matrix $\leftindex_{m}\EffectiveMatrix_i$ associated to $\leftindex_m \ActivationPattern_i$ via Equation~\eqref{eq:effectiveMatrix}}
                    \State $\mathcal{D}_i \gets \mathcal{D}_i \cup \{\leftindex_{m}\EffectiveMatrix_i\}$
                    \State \parbox[t]{\dimexpr\linewidth-6em}{Append node $q_i$ associated to $\leftindex_m \ActivationPattern_i$ including ID, $\leftindex_m \ActivationPattern_i$, $\leftindex_{m}\EffectiveMatrix_i$ for decision rule and samples in this node to $T$}
                \EndFor
            \EndFor
        \If{$\Activation_\NumLayers$ is piecewise linear} \Comment{Add activation of last layer for regression tasks}
        \State Repeat line 4--16 for $i = N$
        \State \parbox[t]{\dimexpr\linewidth-6em} {Append leaf node $q_\NumLayers$ associated to $\leftindex_m \ActivationPattern_\NumLayers$ including ID, $\leftindex_m \ActivationPattern_\NumLayers$, $\leftindex_{m}\EffectiveMatrix_\NumLayers$ and samples in this node to $T$}
        \Else 
        \Comment{E.g., sigmoid or softmax activation for classification tasks}
        \State Use node $q_{\NumLayers-1}$ as leaf node and handle activation function $\Activation_\NumLayers$ separately 
        \EndIf
        \end{algorithmic}
\end{algorithm}

\subsection{Efficiency of the RENTT Algorithm}
\label{sec:Efficiency}
In this section, we present a series of experiments designed to evaluate the performance of Algorithm~\ref{alg:transformation} in terms of runtime and memory consumption, particularly as the complexity of the neural architecture increases. All these numerical experiments underline the theoretical findings of Section~\ref{sec:Theory}, that the reformulation of \acp{nn} into \acp{dt} is exact. This means that loss, accuracy and output of both \ac{nn} models and \ac{dt} are equivalent up to machine precision, as can be seen in Tables~\ref{tab:accuracy_comparison_reg} and~\ref{tab:accuracy_comparison_class} of the appendix.
While previous sections have laid out the theoretical foundations confirming the transformation's validity and accuracy, it is essential to substantiate these claims through rigorous testing that assesses computational efficiency.

We compare the runtime and memory consumption of \ac{rentt} with the algorithm ECDT provided by~\cite{Ngu20}, where the latter is only applicable for classification networks using \ac{relu} activation functions. 
Thus, we use the classification data sets Iris, Diabetes Class and Car Evaluation described in Appendix~\ref{a-sec:data-sets}. 
All experiments are performed on a workstation equipped with an Intel Core i9-14900K processor, NVIDIA RTX 4090 GPU with 24GB VRAM, and 96GB of system RAM.
We evaluate models with varying numbers of hidden neurons. For each configuration, ten models are trained and evaluated five times using different data samples to assess variability.
While model-to-model differences contribute more to variability than data sampling variations, the overall variability observed is negligible.

Due to hardware limitations, transformations for ECDT on networks with a larger number of hidden neurons than presented here could not be executed. Equation~\eqref{eq:node} reveals that the number of leaf nodes depends solely on the total number of neurons and the number of linear regions in the activation function, irrespective of their distribution across layers. Consequently, we observe that runtime and memory consumption remain consistent regardless of neuron distribution, so only the total number of hidden neurons is reported. In our analysis we employ a log-log plot, anticipating the fitting of a straight line, which indicates the power-law relationship $y=a\cdot x^b$ between $y$, the runtime or memory consumption, and $x$, the number of neurons. Exponent $b$ reflects the scaling behavior or growth rate of runtime or memory consumption as the number of neurons increases. To focus on the algorithms' limiting behavior and examine their asymptotic behavior, we fitted the line only for high numbers of neurons. In this context, we employ the Bachmann-Landau notation $\mathcal{O}$~\cite{Mala2022BigO} to describe how the algorithms perform. This approach allows assessing scalability and efficiency, particularly under conditions where hardware constraints limit practical execution for larger networks.

\subsubsection{Computational Cost: Reducing Runtime}
\begin{figure}[H]
    \centering
    \subfigure[Iris data set]{%
        \label{fig:fit_Iris}%
        \includegraphics[width=0.48\linewidth]{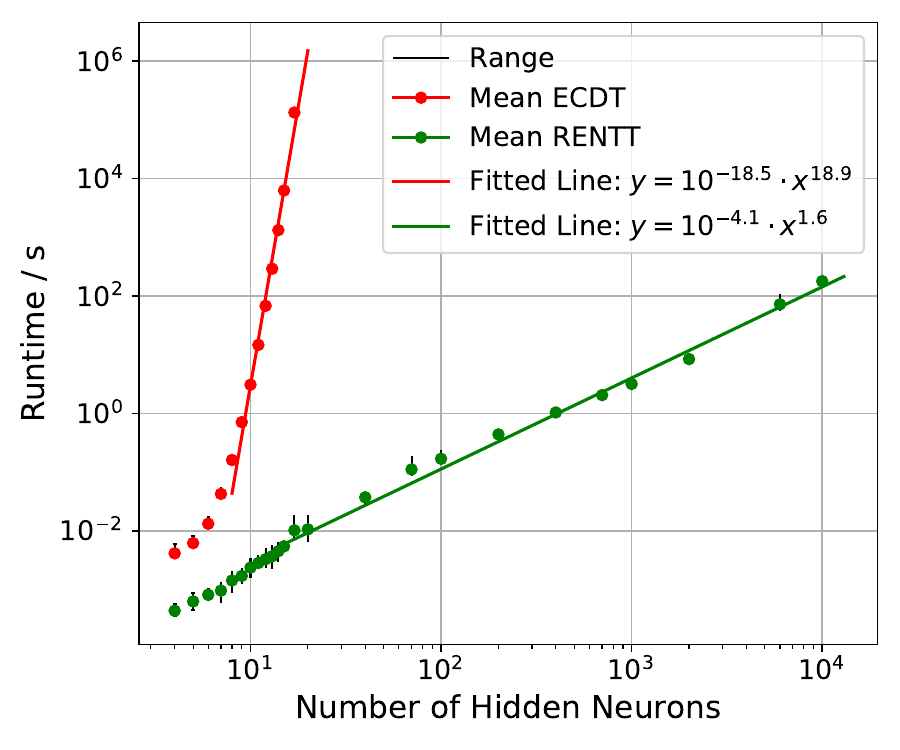}%
    }\hfill
    \subfigure[Diabetes Class data set]{%
        \label{fig:fit_Diabetes}%
        \includegraphics[width=0.48\linewidth]{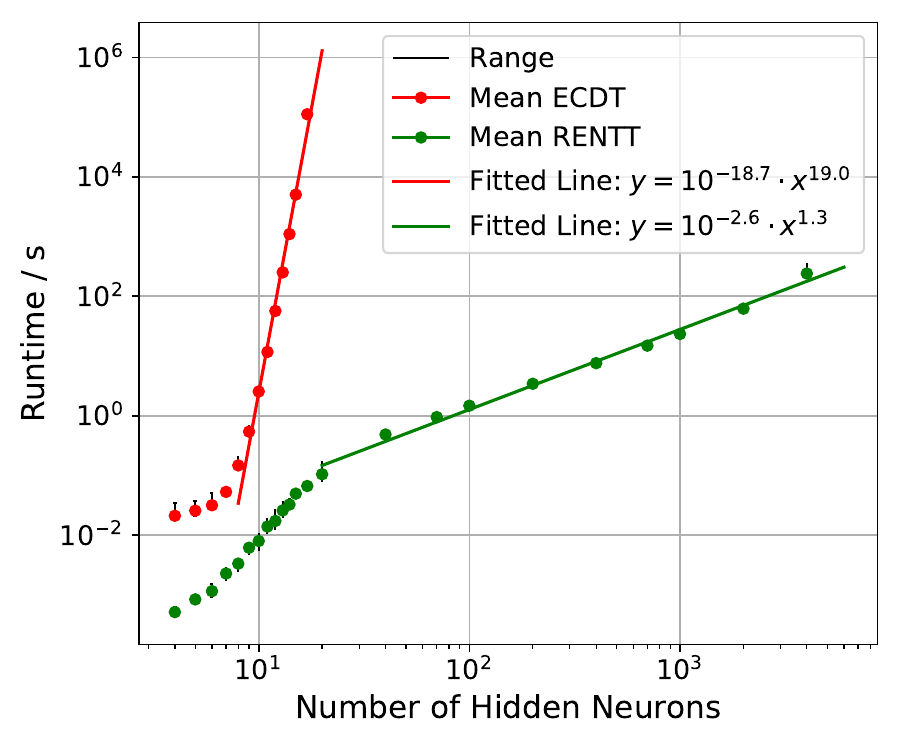}%
    }\hfill
    \subfigure[Car Evaluation data set]{%
        \label{fig:fit_CarEvaluation}%
        \includegraphics[width=0.48\linewidth]{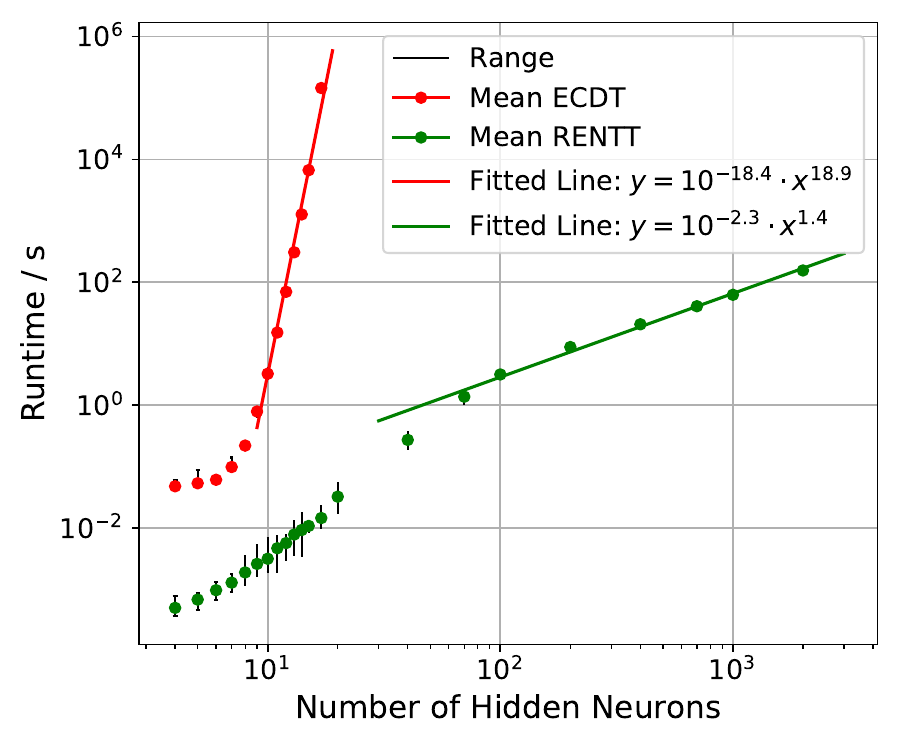}%
    }\hfill
    \caption{Computational time as a function of the number of hidden neurons for ECDT algorithm~\shortcite{Ngu20} (green) and our \ac{rentt} algorithm (red). Standard deviations are represented by the error bars, determined through the transformation of multiple models of the same size. Additional lines are fitted to determine the order of scaling for a large number of neurons.}
    \label{fig:runtime}
\end{figure}

In this section, we analyze the runtime experiments of our proposed algorithm \ac{rentt} compared to the ECDT algorithm~\cite{Ngu20}. The results are illustrated in Figure~\ref{fig:runtime}, which show the runtime performance across different data sets and configurations. The computational time is shown for both \ac{rentt} and ECDT as a function of the number of hidden neurons $x$.

Similar trends are observed for all data sets: the results indicate that \ac{rentt} consistently outperforms ECDT, particularly as the complexity of the \ac{nn} (i.e., the number of hidden neurons) increases. The fitted lines for both algorithms demonstrate that while ECDT's runtime grows exponentially with an order of magnitude of $\mathcal{O}\left(x^{18.9}\right)$ to $\mathcal{O}\left(x^{19.0}\right)$, \ac{rentt} exhibits significantly better scalability with $\mathcal{O}\left(x^{1.3}\right)$ to $\mathcal{O}\left(x^{1.5}\right)$. If hardware constraints were no limitations for $10^4$ hidden neurons, \ac{rentt} needs only $1.6\cdot10^2$~s to $1.9\cdot10^3$~s, which are around $3$~min to $30$~min, where ECDT takes $(1.3-1.9)\cdot 10^{57}$~s, which are around $6\cdot10^{49}$~years.

\subsubsection{Memory Consumption: Reducing Memory}
In this section, we evaluate the memory usage of the \ac{rentt} algorithm in comparison to the ECDT algorithm across multiple data sets. Memory efficiency is crucial when dealing with large \acp{nn}, as excessive memory usage can hinder practical applications, especially in resource-constrained environments.

\begin{figure}[H]
    \centering
    \subfigure[Iris data set]{%
        \label{fig:memory_Iris}%
        \includegraphics[width=0.49\linewidth]{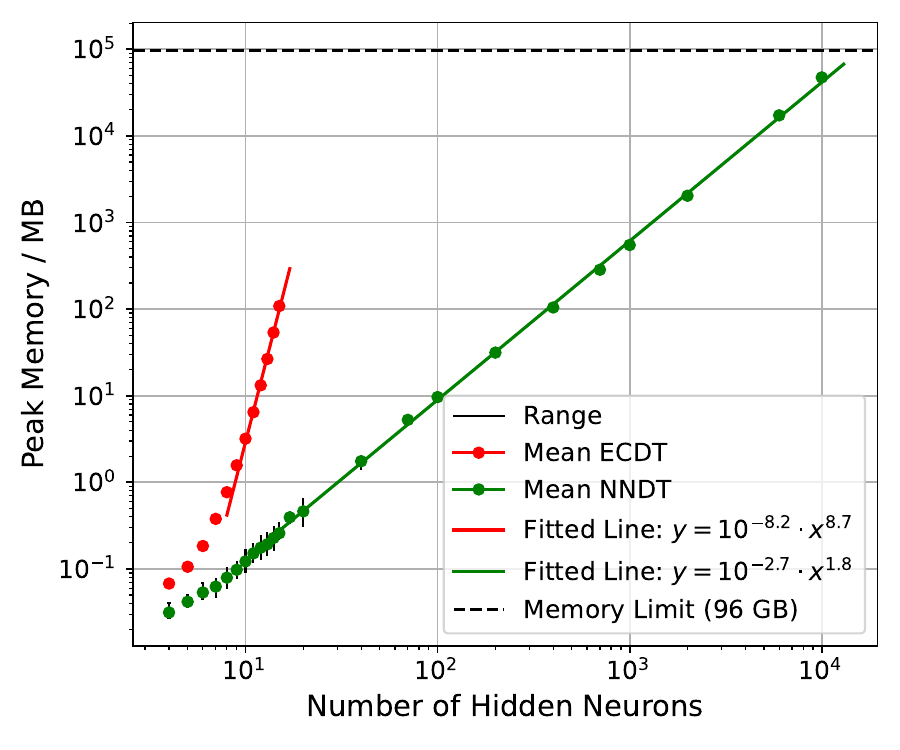}%
    }\hfill
    \subfigure[Diabetes Class data set]{%
        \label{fig:memory_Diabetes}%
        \includegraphics[width=0.49\linewidth]{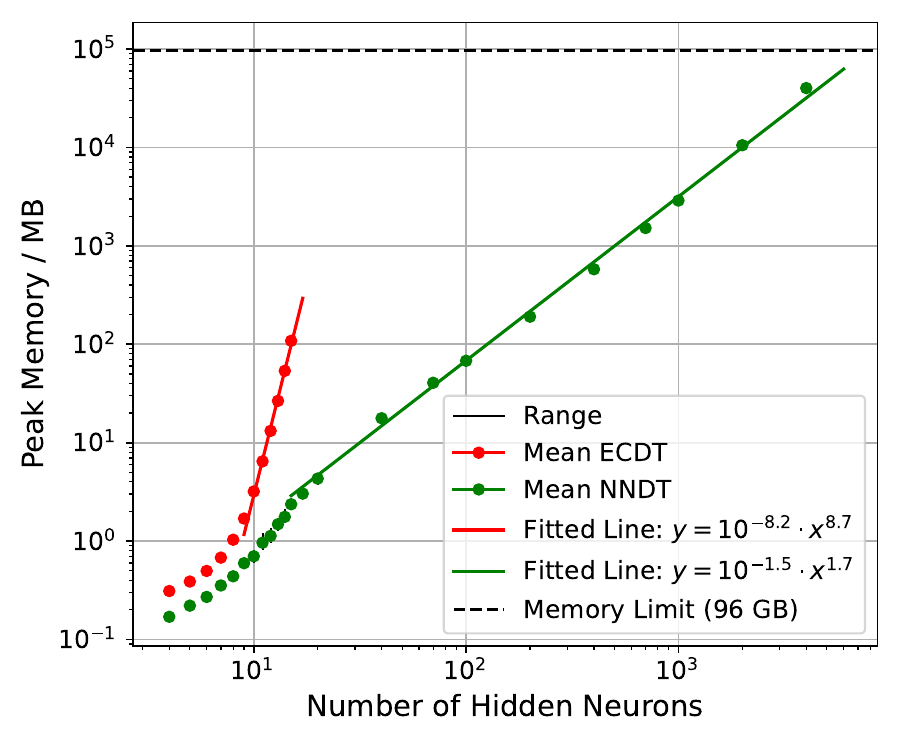}%
    }\hfill
    \subfigure[Car Evaluation data set]{%
        \label{fig:memory_CarEvaluation}%
        \includegraphics[width=0.49\linewidth]{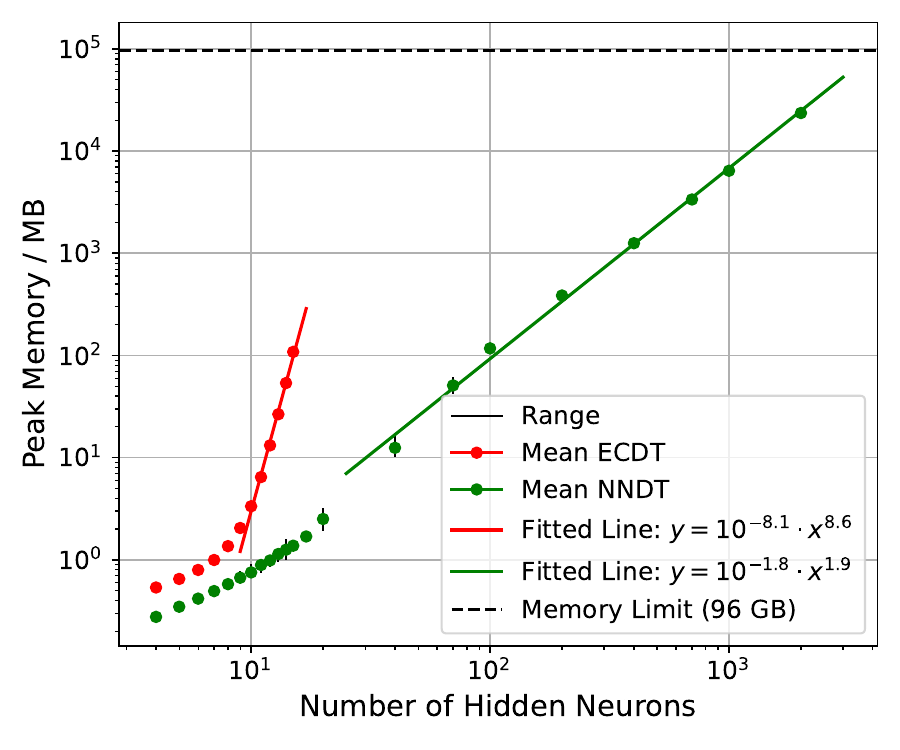}%
    }
    \caption{Maximum \ac{ram} consumption as a function of the number of hidden neurons for ECDT algorithm~\shortcite{Ngu20} (green) and our \ac{rentt} algorithm (red). Standard deviations are represented by the error bars, determined through the transformation of multiple models of the same size. Additional lines are fitted to determine the order of scaling for a large number of neurons.}
    \label{fig:memory}
\end{figure}

Figure~\ref{fig:memory} illustrates the memory consumption for both \ac{rentt} and ECDT as a function of the number of hidden neurons. The experiments were conducted under controlled conditions, with multiple transformations carried out to ensure reliability. Our hardware resources were limited to 96~GB \ac{ram}. Thus, the results of the memory consumption in Figure~\ref{fig:memory} are below the $10^5$~MB line. 
The results clearly show that \ac{rentt} exhibits significantly lower memory consumption than ECDT, particularly as the number of hidden neurons increases. ECDT has a scaling of $\mathcal{O}(x^{8.6})$ to $\mathcal{O}(x^{8.7})$, \ac{rentt} has a scaling of only $\mathcal{O}(x^{1.7})$ to $\mathcal{O}(x^{1.9})$. If hardware constraints were no limitations for $10^4$ hidden neurons, \ac{rentt} would need $31$~GB to $630$~GB \ac{ram}, where ECDT would take $2.0-4.0\cdot10^{23}$~GB, which are $2.0-4.0\cdot10^{20}$~TB. Although this memory consumption remains high in absolute terms, the improvement over ECDT is substantial. This limitation is further discussed in Section~\ref{sec:discussion-limitations}, but the strong difference highlights \ac{rentt}'s superior memory efficiency, making it a more suitable choice for handling complex \ac{nn} architectures.

This analysis is critical for understanding the practical implications of using \ac{rentt} for large-scale applications. By optimizing memory consumption, \ac{rentt} allows for the transformation of larger and more complex \acp{nn} into interpretable \acp{dt} without overwhelming system resources.

The experiments confirm that \ac{rentt} significantly outperforms ECDT in terms of computational efficiency and scalability across various data sets. As the complexity of \acp{nn} increases, the \ac{rentt} algorithm maintains manageable runtime and memory consumption, making it a practical solution for transforming complex \acp{nn} into interpretable \acp{dt}. 
The experiments demonstrate that our approach not only preserves the original \ac{nn} results as derived in the previous sections, but does so in a computationally efficient manner.

%% file: Chapter/5_FIExperiments.tex
\section{Feature Importance Definition (RENTT-FI)} \label{sec:fi-definition}
As previously shown, \acp{nn} can be transformed into equivalent \acp{dt}. 
As they not only produce the exact same output as the \acp{nn}, but also have the same ``inner'' logic, they provide a way to inspect \acp{nn} without the fidelity problems of other post-hoc XAI methods~\cite{bhaumik2022MLMShapLime,fresz2024XAImetric}.
But due to the \acp{dt} being multivariate and similarly complex to the \ac{nn}, this interpretability remains theoretical in nature without further ways to simplify the information contained in the \acp{dt}.
Different explanation types, at least for networks with piecewise linear activation functions in the last layer, could be created via the \acp{dt}---all of them with the benefit of exact calculations and thus, perfect correctness:
\begin{itemize}
    \item Counterfactual explanations: What would need to change in the input to receive a different outcome---either within the same input region (activation pattern) or across regions.
    \item Robustness, i.e., how close a decision is to the next activation pattern.
    \item Visualization of the activation patterns to show which regions in the input space would yield the same result for a given data sample.
    \item Local linear models of the \ac{nn} to yield intelligible local decision rules and surrogate models, respectively.
\end{itemize}
The derivation of these explanation types is left for future work. In the following, we focus on aggregating the \ac{dt} information by means of \acf{fi} explanations, as they closely reflect how \acp{dt} operate and can be presented without losing information for both local and regional cases.
This sort of  \ac{fi} will be called  \ac{rentt-fi} in the following.
\Ac{fi} explanations highlight the most ``important'' features in the input and how much they contributed to the model output.
Such explanations can either be local, i.e., show the most important features of a single input sample, or global, i.e., show the most important features over the entire input space.
Note that approaches in-between are also possible, as  \ac{fi} can also be calculated on a class-wise basis for classification tasks.

The term feature importance is often not clearly defined and multiple interpretations are possible---either as the feature effect $\FeatureEffect$ specifying the weighting of the different input features (how much the prediction changes for each unit increase in the feature), or the feature contribution $\FeatureContribution$, showing how much a feature contributes to the final prediction.
Most methods yield only one of these interpretations.
In contrast, our approach provides both: the effect $\FeatureEffect$, corresponding to the effective weight matrix $\EffectiveMatrix_{\NumLayers-1}$, and the contribution $\FeatureContribution$, as defined in the following.
We provide the feature direction $\FeatureDirection$ separately, indicating whether each feature contributes positively or negatively to the prediction.
Since \ac{rentt-fi} provides both interpretations, we will use this distinction in the following.

For a \ac{dt} and thus, the \ac{nn} in question, the local feature contribution of an input feature vector $\Feature_0$ regarding feature $j$ can be calculated as 
\begin{align}
    {\FeatureContribution_{\text{local}}}_{l,j}={\left[\leftindex_m\EffectiveMatrix_{\NumLayers-1}\right]}_{l,j}\cdot{\left[\Feature_0\right]}_j~,
\end{align}
Here, $\leftindex_m\EffectiveMatrix_{\NumLayers-1}$, including all $\Feature_0$ with same activation pattern, is the effective weight matrix of the output layer, as defined in Equation~\eqref{eq:effectiveMatrix}, corresponding to the input region $m$, that contains $\Feature_0$. This matrix is dependent on $\Feature_0$. It is provided in the leaf node and also indicates the feature effect. Often the weight matrix for the output layer is a column vector, which implies $l=0$. For multiclass classification involving $n$ classes $l=0,\ldots,n-1$, the feature contribution can be calculated separately for each class to create class-specific explanations.
The local feature contribution can now be aggregated by averaging it for each feature $j$ over all samples in the set of input feature vector $\mathcal{X}_0$ per input region $m$ into a regional feature contribution, such as
\begin{align}
    {\FeatureContribution_{\text{regional}}}_{l,j}^m=\frac{1}{\left|\mathcal{M}\right|}\sum_{\mathcal{M}}\lvert{\left[\leftindex_m\EffectiveMatrix_{\NumLayers-1}\right]}_{l,j}\cdot{\left[\Feature_0\right]}_j\rvert~,
\end{align}
with $\mathcal{M}=\{\Feature_0 \in \mathcal{X}_0\mid\ActivationPattern(\Feature_0) = \leftindex_m\ActivationPattern\}$.
The feature direction is calculated via
\begin{align}
    {\FeatureDirection_{\text{regional}}}_{l,j}^m=\sign\left(\frac{1}{\left|\mathcal{M}\right|}\sum_{\mathcal{M}} {\left[\leftindex_m\EffectiveMatrix_{\NumLayers-1}\right]}_{l,j}\cdot{\left[\Feature_0\right]}_j\right)~.
\end{align}
The regional feature effect is given by the effective weight matrix for this region $m$ per class $l$ and feature $j$ according to
\begin{align*}
    {\FeatureEffect_\text{regional}}_{l,j}^m={\left[\leftindex_m\EffectiveMatrix_{\NumLayers-1}\right]}_{l,j}~.
\end{align*}
Note that all samples within region $m$ share the same local feature effect.

We can also calculate an average feature contribution over all input regions to obtain a global feature contribution 
\begin{align*}
     {\FeatureContribution_{\text{global}}}_{l,j}=\frac{1}{|\ActivationPattern(\mathcal{X}_0)|}\sum_{m=1}^{|\ActivationPattern(\mathcal{X}_0)|}\frac{1}{\left|\mathcal{M}\right|}\sum_{\mathcal{M}} \lvert{\left[\leftindex_m\EffectiveMatrix_{\NumLayers-1}\right]}_{l,j}\cdot{\left[\Feature_0\right]}_j\rvert\\
     = \frac{1}{|\ActivationPattern(\mathcal{X}_0)|}\sum_{m=1}^{|\ActivationPattern(\mathcal{X}_0)|} {\FeatureContribution_{\text{regional}}}_{l,j}^m
\end{align*}
with feature direction
\begin{align}
    {\FeatureDirection_\text{global}}_{l,j}=\sign\left(\frac{1}{|\ActivationPattern(\mathcal{X}_0)|}\sum_{m=1}^{|\ActivationPattern(\mathcal{X}_0)|}\frac{1}{\left|\mathcal{M}\right|}\sum_{\mathcal{M}} {\left[\leftindex_m\EffectiveMatrix_{\NumLayers-1}\right]}_{l,j}\cdot{\left[\Feature_0\right]}_j\right)
\end{align}
and global feature effect for class $l$ and feature $j$
\begin{align*}
        {\FeatureEffect_\text{global}}_{l,j}=\frac{1}{|\ActivationPattern(\mathcal{X}_0)|}\sum_{m=1}^{|\ActivationPattern(\mathcal{X}_0)|} {\left[\leftindex_m\EffectiveMatrix_{\NumLayers-1}\right]}_{l,j}~.
\end{align*}

 
\section{Feature Importance Experiments}
\label{sec:FIexperiments}
In this section, we present comparisons of our \ac{fi} method with other methods commonly used in literature.
Our proposed \ac{rentt-fi} approach analyzes features at local, regional, and global scales, capturing both feature contributions and feature effects. Methods in literature provide only feature contributions while \ac{rentt-fi} also computes feature effects.
The transformation of \acp{nn} into equivalent \acp{dt} can be used to study different questions:
\begin{enumerate}
    \item For two functionally equivalent models, the \ac{nn} and its \ac{dt}, do the common  \ac{fi} methods produce the same  \ac{fi} values? \label{item:implementation-invariance}
    \item Do the common \ac{fi} methods and \ac{rentt-fi} show the expected results for simple data sets? Are their results similar for more complex ones? \label{item:FI-vs-RENTT}
    \item What are the performance (dis)advantages of the  \ac{fi} methods vs. \ac{rentt-fi}? \label{item:performance}
\end{enumerate}
While (\ref{item:implementation-invariance}) is more of a sanity check of the  \ac{fi} methods, also referred to as implementation-invariance~\cite{Nauta2022}, (\ref{item:FI-vs-RENTT}) tests the common  \ac{fi} methods for correctness, defined by them conforming to the ground truth \ac{fi} produced by \ac{rentt-fi}, and (\ref{item:performance}) refers to the implementation details and practical applicability of \ac{rentt-fi}.

\subsection{Common Feature Importance Methods}
\label{ssec:experiments-methods}
To comprehensively compare \ac{fi} methods across different datasets, we employ a diverse set of both local and global \ac{fi} methods, as described in the following. This approach enables us to capture various perspectives—from individual prediction explanations to model-wide importance—thereby providing a more robust foundation for our comparative analysis.

All methods are used with their default values, with the background data samples being set to at most 1000 training data samples to reduce runtime and memory complexity.
For the smaller data sets, all training data samples were used as background samples.

\subsubsection{Local Methods}
\label{sssec:experiments-local}
For the local \ac{fi} methods, we employ the toolbox dalex~\cite{dalex2021}. 
Three different local \ac{fi} methods are used, as described in the following.

One of the first \ac{xai} methods for machine learning decisions was \ac{lime}~\cite{Ribeiro2016LIME}. It uses simpler models to approximate the decision boundaries of more complex machine learning models, such as \acp{nn}.  Interpretable models, such as linear regression, serve as these simpler models to create  \ac{fi} explanations for data instances.

As a systematization of previous explanation methods, \ac{shap}~\cite{Lundberg2017SHAP} was proposed to provide local explanations by approximating Shapley Values, a game theoretic concept.
Compared to other explanation methods, this approach provides interesting theoretical properties: Local Accuracy, Missingness, and Consistency.
Local Accuracy means that the explanation conforms to the model within a small area around the data sample in question.
Missingness indicates that features missing in the original input have a  \ac{fi} of $0$, and Consistency describes that increasing the influence of a feature results in a corresponding increase in its \ac{fi}.
While useful in theory, \ac{shap} explanations are criticized, for example, for the resulting explanations being too complex for users~\cite{Kumar2020SHAPProblems}, \ac{fi} being attributed to features without impact on the result~\cite{Covert2021ExplainingByRemoving,Merrick2020ExplanationGame}, or for the actual implementation of the methods not holding the theoretical properties~\cite{Sundarajan2020ManyShapleyValues,slack2020fooling}.

The third local explanation method, \ac{bd}~\cite{Staniak2018BreakDown}, provides \ac{fi} values for a sample by computing the difference between the average model prediction and the average model prediction with the input features iteratively fixed to the specific feature values of the sample tested.
Note that, for additive models, this approach yields the same results as \ac{shap}, as noted in the original paper.

\subsubsection{Global Methods}
\label{sssec:experiments-global}
For the global \ac{fi} methods, the implementation of~\cite{Covert2020SAGEGit} is used.
Since not many different methods for estimating global \acp{fi} are available, only two are used here.
\ac{se}~\cite{Owen2014ShapleyEffects} quantify the model's sensitivity to each feature.
Related to \ac{se}, \ac{sage}~\cite{Covert2020SAGE} quantifies  \ac{fi} by estimating the predictive power each feature contributes. To account for feature interactions, the Shapley values are used.

\subsection{Feature Importance Comparison}
\label{ssec:FIexperiments-proceed}
We compare the local and global \ac{fi} methods described in Section~\ref{ssec:experiments-methods} on the data sets listed in Section~\ref{ssec:FIexperiments-datasets}. As before, all experiments are performed on a workstation equipped with an Intel Core i9-14900K processor, NVIDIA RTX 4090 GPU with 24GB VRAM, and 96GB of system RAM.
No significant differences between multiple training runs were observed.
As such, the results will be presented for only one model per data set for a clearer presentation. 

When comparing the different \acp{fi}, problems arise with their different interpretations, especially the so-called intercept.
The intercept can be interpreted as ``If no input feature provides any information, what is the expected model output?'' or ``How much of the input cannot be explained via the  \ac{fi}?''.
As such, it depends on the baseline used.
Per default, \ac{shap} and \ac{bd} use the output of the black box model when all input features are at their expected value as a baseline, resulting in the same intercept for all samples.
In contrast, \ac{lime} uses the bias of the local surrogate model, resulting in a sample-dependent intercept, while the intercept of \ac{rentt-fi} depends on the local input region in which the sample is located.
More precisely, the intercept of \ac{rentt} is the bias of the local linear model, which accurately describes the behavior of the neural network within that region.
When all  \ac{fi} values are summed together with this intercept, they reconstruct the final prediction of the model.
\ac{mae} could be used instead of \ac{rmse} for comparing  \ac{fi} methods, which would allow to compensate for this intercept offset, eliminating its influence on the comparison.
However, doing so would defeat the purpose: The \ac{mae} would only compare the final predictions (feature scores plus intercept) rather than understanding how individual features contribute differently across methods.
We choose to use \ac{rmse} without compensating for the intercept. This ensures a fair comparison because the intercept is an inherent part of how each \ac{fi} method generates its explanations. Removing it would alter the fundamental nature of what we are evaluating and potentially mask important differences between methods.

Additionally, Krippendorff's $\alpha$ will be used to evaluate the \ac{fi}-results~\cite{krippendorff2004alpha}.
This measure was proposed to evaluate the agreement between different psychometric tests, as in this context, results can be different depending on the specific test used and the person administering the tests.
Krippendorff's $\alpha$ was previously used in the context of \ac{xai}, mainly to evaluate whether different metrics for \ac{xai} methods agree in their results~\cite{tomsett2019sanityformetrics,fresz2024XAImetric}.
It provides values in [-1, 1], where $-1$ denotes structural disagreement between given results, $0$ denotes neither structural agreement nor disagreement and $1$ denotes perfect agreement between results.
Within this work, Krippendorff's $\alpha$ is used similar to the \ac{rmse} to evaluate whether the \ac{fi} methods agree in their results.
Two versions of $\alpha$ are used: Ordinal-scaled, measuring whether the ranking of the features based on the \ac{fi} is the same for the different methods, and interval-scaled, measuring whether the methods agree in the magnitude of the allocated \ac{fi}.
We use the version of~\cite{castro2017fastKrippendorff}.

In addition, we provide a \ac{re} to contextualize the mean differences against the typical magnitude of the \ac{fi} values.
As such, the \ac{re} between two methods is defined as the \ac{rmse} over all samples divided by the mean magnitude of both \ac{fi} across all samples and features on the data set. 

For local \ac{fi} methods, the \ac{rmse} between each \ac{fi} method pair will be computed samplewise and displayed as a violin plot, with the mean \ac{rmse} and standard deviation. This comparison will be done across entire data sets, to achieve a fair result.
The Krippendorff's $\alpha$ results are also one value for $\alpha$ per data point, once again presented as a violin plot.
Additionally, we provide the \ac{rmse} and \ac{re} as mean differences over all data points and include the standard deviation for both \ac{rmse} and \ac{re} as additional measures of variability.
For the local methods, the \ac{re} between two methods is defined as the \ac{rmse} of a sample divided by the mean \ac{fi} across all samples and features of both \ac{fi} methods on this data set.

Since the global \ac{fi} methods only provide one importance value per feature in the data set, the pairwise comparison of them via the \ac{rmse} is straightforward. 
Therefore, we provide the \ac{rmse} and \ac{re}, both with standard deviations, and also the Krippendorff's $\alpha$ results.
The full results for both local and global methods can be found in Appendix~\ref{a-sec:additional-FI-results}.

\subsection{Data Sets and Neural Network Training}
\label{ssec:FIexperiments-datasets}
To be able to evaluate XAI methods, data sets with available ground truth explanations can be helpful.
For this purpose, simple functions with known  \ac{fi}s are used, similar to previous publications with the same approach~\cite{Liu2005FIvsWhiteBox, SOBOL1998FIvsWhiteBox, Ishigami1990FIvsWhiteBox}.
This allows to identify differences between the expected results of the \ac{xai} methods and their actual results.
Note that evaluating XAI vs ground truth explanations may suffer from problems, especially if the model itself does not approximate the data perfectly and if the input space is high-dimensional, resulting in different models, and thus, explanations, being possible (e.g., for image data).
As such, an explanation might conform to the model in question, i.e., be correct, but may not be the expected explanation for a given data set.

The functions used here should be simple enough to enable the \acp{nn} to (almost) perfectly approximate the expected outputs, as shown by their accuracy in Table~\ref{tab:accuracy_comparison_reg} in Appendix~\ref{app:performance}. 
The following simple functions are used: 
\begin{itemize}
    \item Absolute:
    \begin{align}
        y_0 = |\Feature_{0}| \label{eq:fi_absolute}
    \end{align}
    \item Linear:
    \begin{align}
        y_0 = 4 \cdot {\Feature_{0,0}} + {\Feature_{0,1}} + 0.001\cdot \Feature_{0,2}~,\label{eq:fi_linear}
    \end{align} 
\end{itemize}
where $y_0$ denotes the ground truth label and $\Feature_{0,k}$ denotes feature $k$ of input sample $\Feature_0$.

While simple functions with known ground truth explanations can serve as sanity checks for  \ac{fi} methods, they and their corresponding models might lack the complexity to show significant differences between the methods.
Therefore, more complex data sets are used and the differences between methods quantitatively evaluated.
For regression tasks, the data sets Diabetes Reg and California Housing are used, for classification Iris, a different Diabetes data set (here called Diabetes Class), Wine Quality, Car Evaluation and Forest Cover Type.
A more detailed description of the data sets can be found in Appendix~\ref{a-sec:data-sets}.
The resulting \ac{fi} values of all methods directly depend on the scale of the input features. To achieve a fair comparison of the \ac{fi} methods, the input values for all data sets are normalized to $[0,1]$. 
Most of the previous data sets can be considered as ``toy'' data sets, all of them but the Forest Cover Type Data Set can be solved using a \ac{nn} with two hidden layers with eight neurons in the first and four in the second hidden layer.
For the Forest Cover Type data set, a network with 64 neurons in the first and 32 neurons in the second hidden layer was used.
For model training, all data sets were split into 80\% training data and 20\% testing data.
The model performance for the respective data sets is presented in Tables~\ref{tab:accuracy_comparison_reg} and~\ref{tab:accuracy_comparison_class} in Appendix~\ref{app:performance}.

\subsection{Results}
\label{ssec:experiments-results}
The experimental findings addressing the three key research questions are presented below.
We systematically examine implementation-invariance across functionally equivalent models, evaluate the alignment of common \ac{fi} methods with \ac{rentt-fi} as ground truth, and assess computational performance considerations.

\subsubsection{Differences within Feature Importance Methods between Decision Tree and Neural Network} \label{sec:fi-nn-vs-dt}
An important aspect of XAI methods is their implementation invariance~\cite{Nauta2022}, as also described in Section~\ref{sec:FIexperiments}.
A part of implementation invariance is that the result for \ac{fi} methods should be the same for functionally equivalent models.
Here, \ac{rentt} provides two equivalent models: the original \ac{nn} and the \ac{dt}.
As such, using the \ac{fi} methods on both should provide the same results, or with regard to the sample-based nature of the methods at least very similar results.
For the data sets described in Section~\ref{ssec:FIexperiments-datasets}, the comparison of the \acp{fi} between \ac{nn} and \ac{dt} for the local methods can be found in Figure~\ref{fig:FI_nndt_datasets_local} (Table~\ref{tab:nn_vs_dt_local_relative} in Appendix~\ref{a-ssec:nn-vs-dt}), the results for the global methods in Figure~\ref{fig:FI_nndt_datasets_global} (Table~\ref{tab:nn_vs_dt_global_relative} in Appendix~\ref{a-ssec:nn-vs-dt}).

For local \ac{fi} methods (Figure~\ref{fig:FI_nndt_datasets_local} and Table~\ref{tab:nn_vs_dt_local_relative}), the results reveal a clear pattern across the two data set types.
On regression tasks (Absolute, Linear, Diabetes Reg, California Housing), \ac{bd} produces the closest results, with \ac{re} values typically below $5\,\%$, seeming rather implementation-invariant for these tasks.
However, on classification tasks, \ac{bd} shows errors, with \ac{re} values frequently exceeding $200\,\%$ ($226\pm76\,\%$ for Iris, $225\pm55\,\%$ for Wine Quality, and $278\pm90\,\%$ for Car Evaluation).
\ac{lime} and \ac{shap} display similar patterns, but with worse \ac{re} values on regression tasks (typically $2-58\,\%$), with \ac{lime} performing slightly better than \ac{shap} on most data sets.
On classification tasks, both methods also exhibit high \ac{re} values ($14-277\,\%$), though \ac{shap} often shows slightly higher errors than \ac{lime}. 
Notably, the Diabetes Class data set represents an exception where all methods show relatively low \ac{re} values (\ac{bd}: $(5\pm2) \cdot 10^{-4}\,\%$, \ac{lime}: $14\pm4\,\%$, \ac{shap}: $26\pm29\,\%$).
This contrast between regression and classification performance suggests that the methodology of the \ac{fi} methods may be particularly sensitive to the discrete decision boundary characteristic of classification networks.
Another possible explanation may be that for classification tasks, only a limited amount of features receives weights $\gg0$, resulting in a very low average \ac{fi} across a data set and thus, high \ac{re} values.
\begin{figure}[H]
    \centering
    \subfigure[\ac{re} comparison for local \ac{fi} methods.]{%
        \label{fig:FI_nndt_datasets_local}%
        \includegraphics[width=0.44\textwidth]{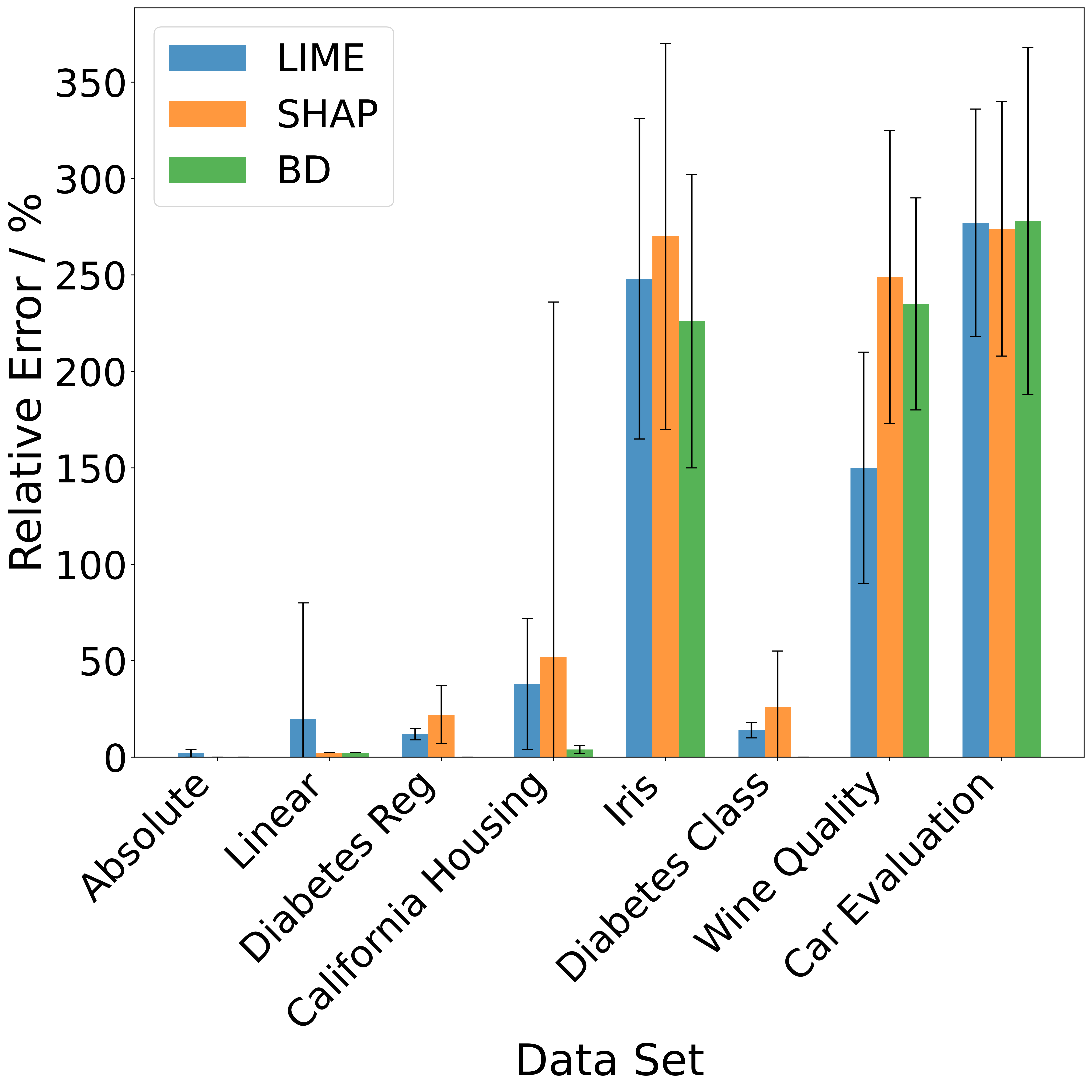}%
    }\hfill
    \subfigure[\ac{re} comparison for global \ac{fi} methods.]{%
        \label{fig:FI_nndt_datasets_global}%
        \includegraphics[width=0.44\textwidth]{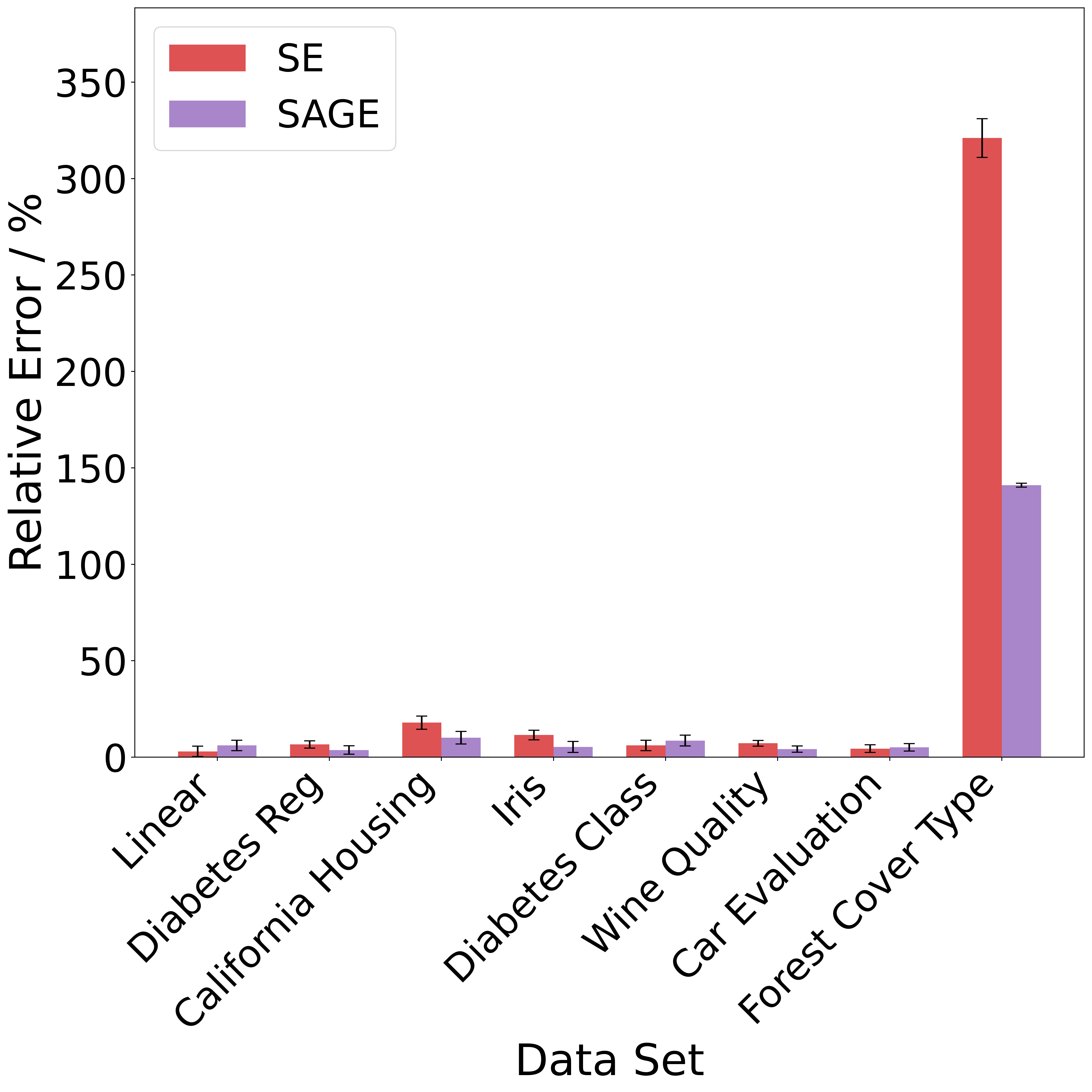}%
    }
    \caption{Comparative analysis of \ac{re} with standard deviation for local and global \ac{fi} methods when applied to functionally equivalent \ac{nn} and \ac{dt} models across all data sets where \ac{fi} calculations were possible. For the local methods on the Forest Cover Type data set, only \ac{rentt-fi} could be executed due to memory constraints.}
    \label{fig:FI_nndt_datasets_local_global}
\end{figure}

For the global \ac{fi} methods (Figure~\ref{fig:FI_nndt_datasets_global} and Table~\ref{tab:nn_vs_dt_global_relative}), both \ac{se} and \ac{sage} have a slightly lower \ac{re} for the classification tasks ($4.4\pm2.2\,\,\%$ to $11.5\pm2.5\,\%$ for \ac{se} and $4.2\pm1.6\,\%$ to $8.6\pm2.8\,\%$ for \ac{sage}) than for the regression tasks ($3.0\pm2.7\,\%$ to $17.9\pm3.4\,\%$ for \ac{se} and $3.7\pm2.2\,\%$ to $10.1\pm3.3\,\%$ for \ac{sage}).
The Forest Cover Type data set is an outlier with much higher \acp{re} ($321\pm10\,\%$ for \ac{se} and $141\pm1\,\%$ for \ac{sage}).


Overall, these results reveal significant implementation-dependence and underscore the lack of implementation-invariance of the current \ac{xai} methods, especially for classification tasks.
Only for \ac{bd} on regression tasks, consistently good results can be observed.
The more consistent performance of global methods suggests they may be less implementation-dependent across diverse problems, though their relative errors remain non-negligible.
Collectively, these findings show limitations in current XAI methods and underscore the value of \ac{rentt-fi}'s ground truth  \ac{fi} as a benchmark for method evaluation.

\subsubsection{Toy Example and Exploration of the Differences between LIME, SHAP, BD and RENTT-FI}\label{sec:fi-toy}
As described before, the simple functions defined in Section~\ref{ssec:FIexperiments-datasets} allow to explicitly compare the \ac{fi} methods with the ground truth.
For Equations~\eqref{eq:fi_absolute} and~\eqref{eq:fi_linear}, the small \acp{nn} approximate the corresponding functions almost perfectly, as shown via their Mean Squared Error (MSE) in Table~\ref{tab:accuracy_comparison_reg} in Appendix~\ref{app:performance}. As such, the explanation results for both functions can be compared to the ground truth parameters used in the functions themselves. 

\begin{figure}[H]
  \centering
  \includegraphics[width=0.63\linewidth]{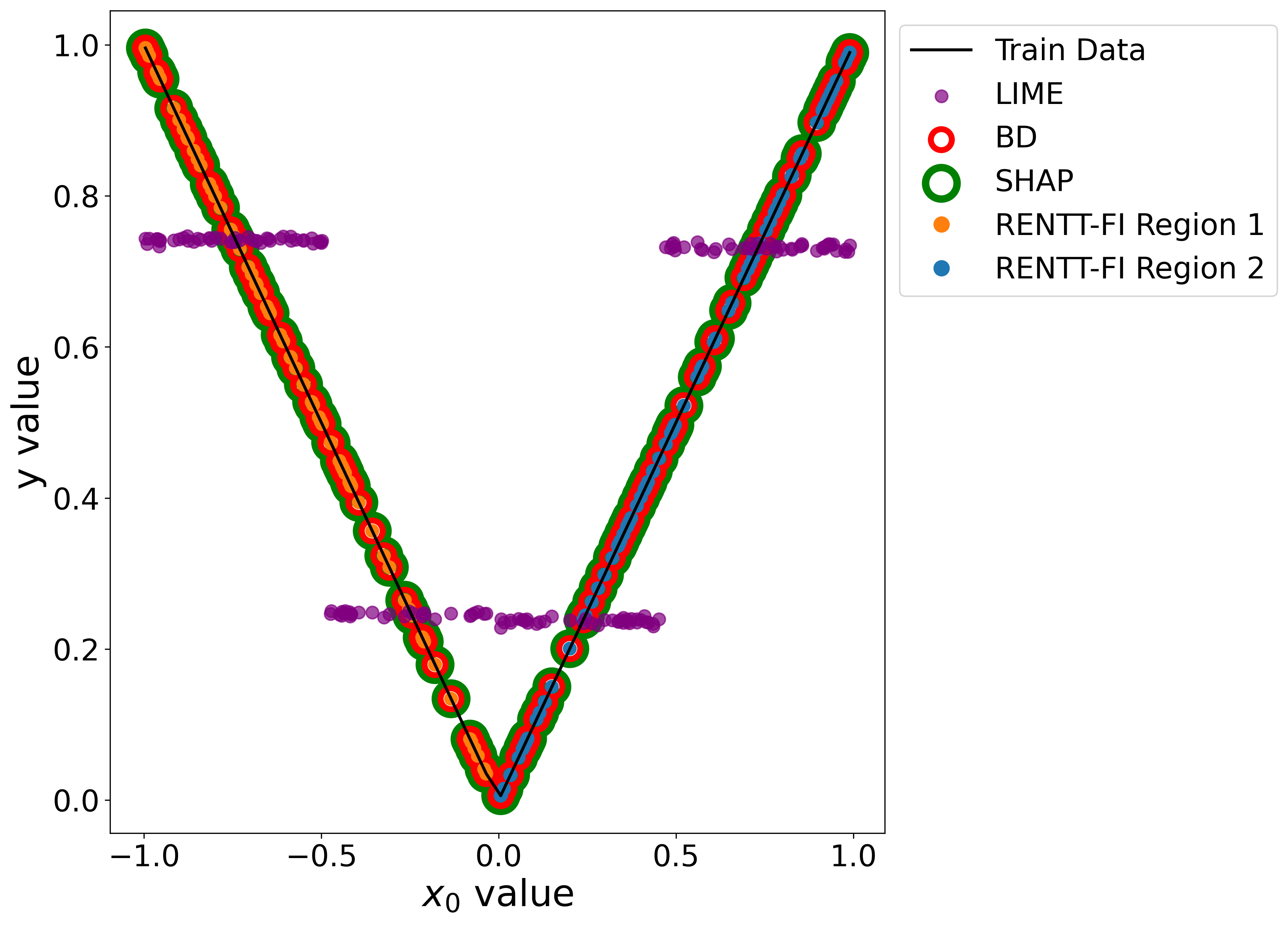}
  \caption{Training data generated via Function~\eqref{eq:fi_absolute} and prediction of the \ac{nn} (in blue/orange) with the weights given by \ac{rentt-fi}. The linear model for $x<0$ uses the effective weight $-1$ (orange), whereas the linear model for $x>0$ uses the weight $1$ (blue). Additionally, \ac{lime}- (red), \ac{bd}- and \ac{shap}-\acp{fi} (green) are shown, compensated for their intercept.}
  \label{fig:FI_abs_approximation}
\end{figure}

\begin{figure}[H]
  \centering
  \includegraphics[width=0.73\linewidth]{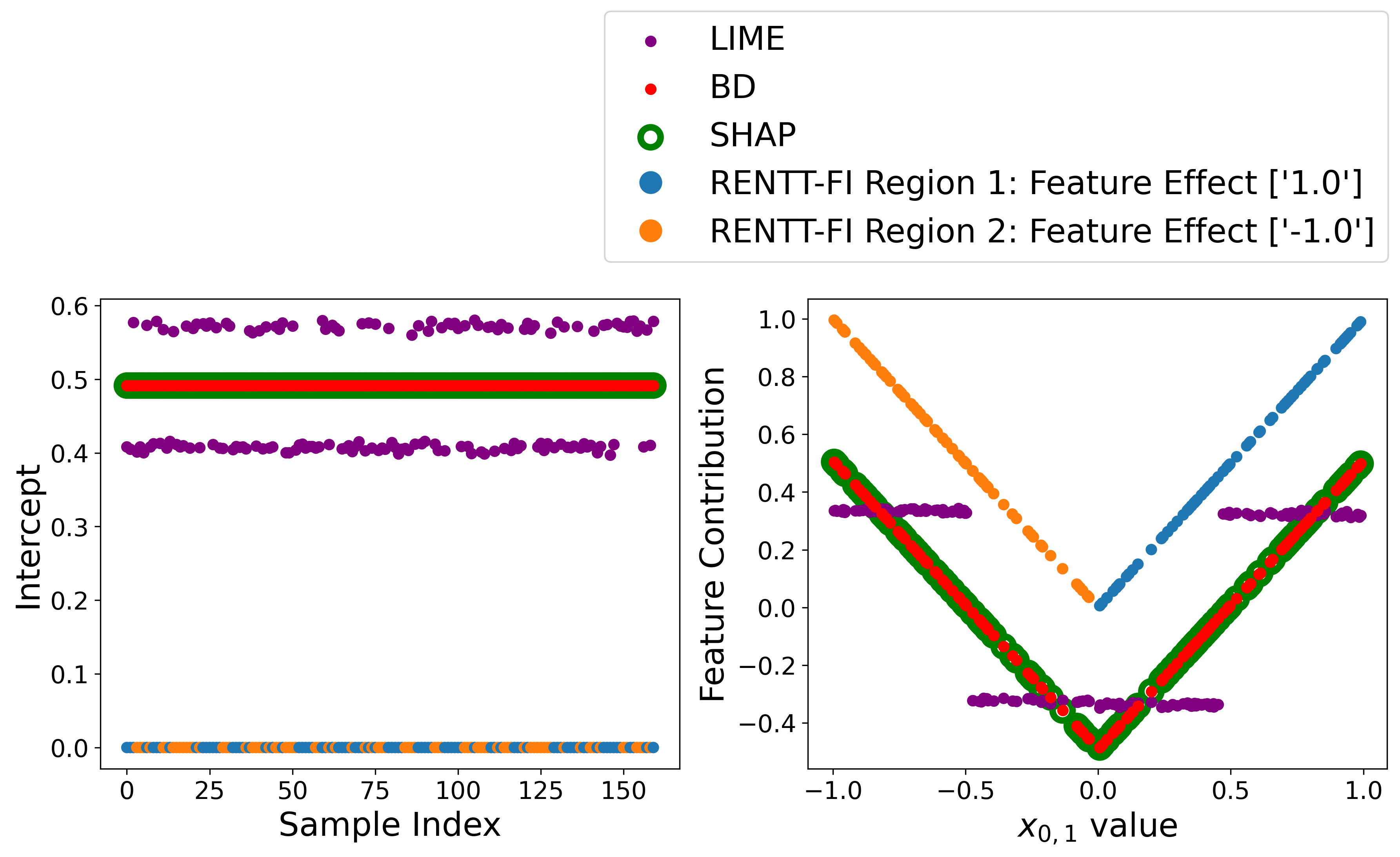}
  \caption{Intercept (left) and \acp{fi} (right) for the different methods for data generated via Function~\eqref{eq:fi_absolute}.}
  \label{fig:FI_abs_fi_intercept}
  \end{figure}

For the function given via Equation~\eqref{eq:fi_absolute}, one would expect the feature effect of $\Feature_{0}$ to be $-1$ for $\Feature_{0} < 0$ and $1$ for $\Feature_{0}>1$, representing $|\Feature_{0}|$. For this function, no meaningful global \ac{fi} or Krippendorff's $\alpha$ can be computed, since only one feature is present. 
The \ac{nn} approximation of the original function can be seen in Figure~\ref{fig:FI_abs_approximation}, together with the \acp{fi}. In the one-dimensional case, the \acp{fi} can be easily compensated for their intercept. As expected, the \ac{nn} provides a near-perfect approximation of the original function.
In addition, the \ac{fi} weights of the corresponding linear models of \ac{rentt-fi} conform to the expected effective weights of $-1$ for $x<0$ and $1$ for $x>0$.
Qualitatively, the \ac{shap} and \ac{bd} values agree with this, when compensated for the intercept.
Due to \ac{shap} and \ac{bd} providing an intercept of $0.5$ instead of $0$ (Figure~\ref{fig:FI_abs_fi_intercept}), these \acp{fi} have a constant offset compared to the \acp{fi} values of \ac{rentt-fi} and the ground truth values.
The \ac{fi} of \ac{lime}, based on the parameters used, results in plateaus (as shown in Figures~\ref{fig:FI_abs_approximation} and~\ref{fig:FI_abs_fi_intercept}), or in a hyperbolic tangent, exemplifying the parameter dependence of this method.


  \begin{figure}[H]
 \centering
 \includegraphics[width=1\linewidth]{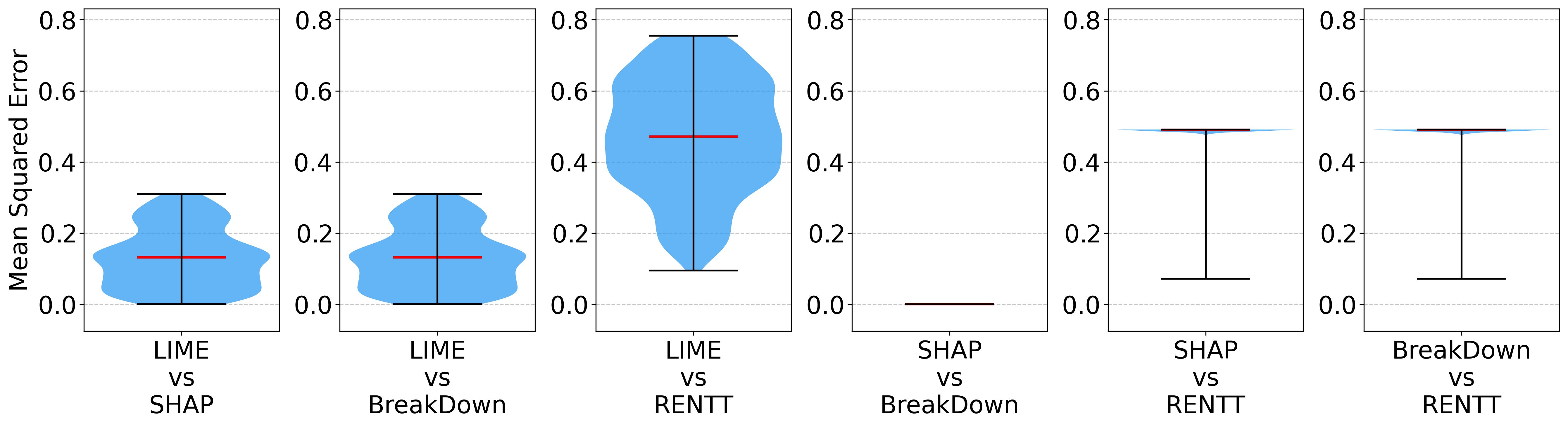}
 \caption{Violin plot of the \ac{rmse} between the different \ac{fi} methods for the data set generated via the function in Equation~\eqref{eq:fi_absolute}. Only the 95th percentile of samples is pictured.}
 \label{fig:FI_abs_rmse_violin}
\end{figure}

Quantitatively, these differences can also be seen in Figure~\ref{fig:FI_abs_rmse_violin}.
While \ac{shap}- and \ac{bd}-\acp{fi} are the same within computational accuracy, the difference between these two methods and the \ac{rentt-fi} is quite low and only dominated by the difference in the intercept (\ac{rmse} of 0.5).
The \ac{rmse} between \ac{lime} and \ac{shap} and \ac{bd} is substantially higher, ranging from $0$ to $0.35$ for the 95th percentile, with most samples having an \ac{rmse} below $0.2$.
The largest differences can be seen between \ac{rentt-fi} and \ac{lime}, with an \ac{rmse} between $0.1$ and $0.78$, once again showing the influence of the intercept.

\begin{figure}[H]
    \centering
    \subfigure[\ac{se}]{%
        \includegraphics[width=0.35\textwidth]{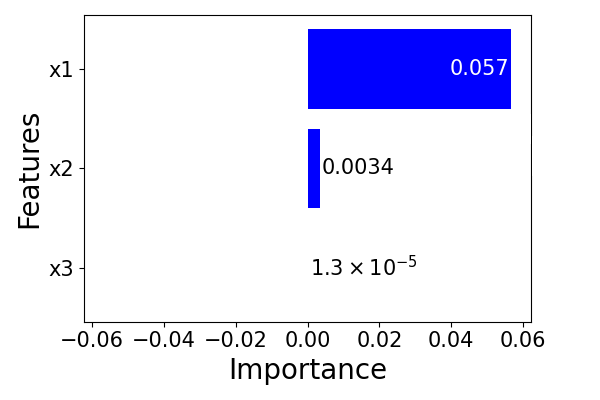}%
    }
    \subfigure[\ac{sage}]{%
        \includegraphics[width=0.34\textwidth]{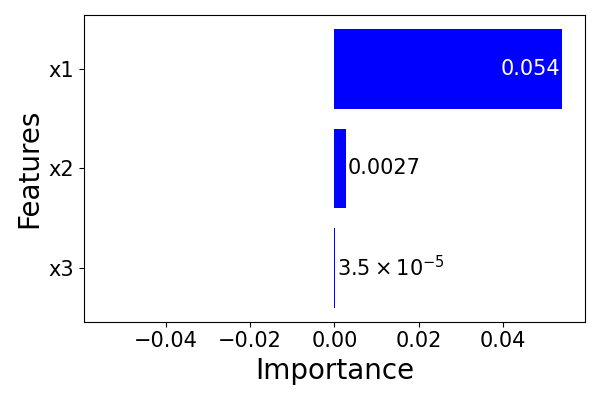}%
    }
    \\
    \subfigure[\ac{rentt-fi} feature contribution]{%
        \includegraphics[width=0.33\textwidth]{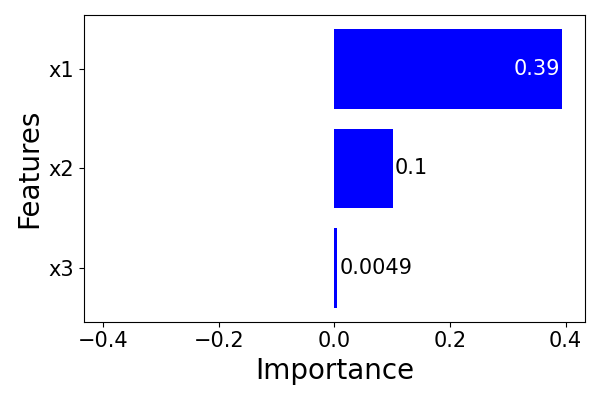}%
    }
    \subfigure[\ac{rentt-fi} feature effect]{%
        \includegraphics[width=0.33\textwidth]{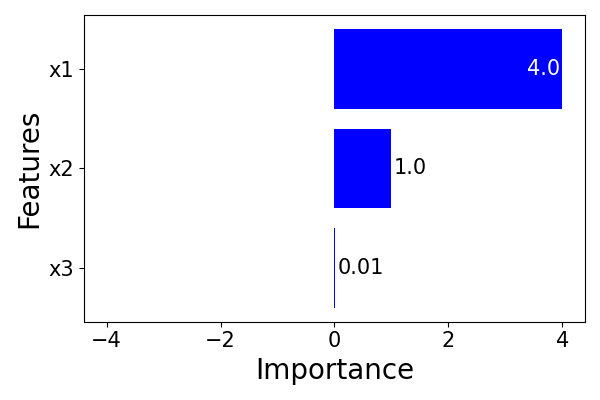}%
    }
    \caption{Global \ac{fi} by \ac{se}, \ac{sage} and \ac{rentt-fi} on an \ac{nn} trained on the data set generated via the function in Equation~\eqref{eq:fi_linear}.}
    \label{fig:FI_lin_global}
\end{figure}

\begin{figure}[H]
 \centering
 \includegraphics[width=0.7\linewidth]{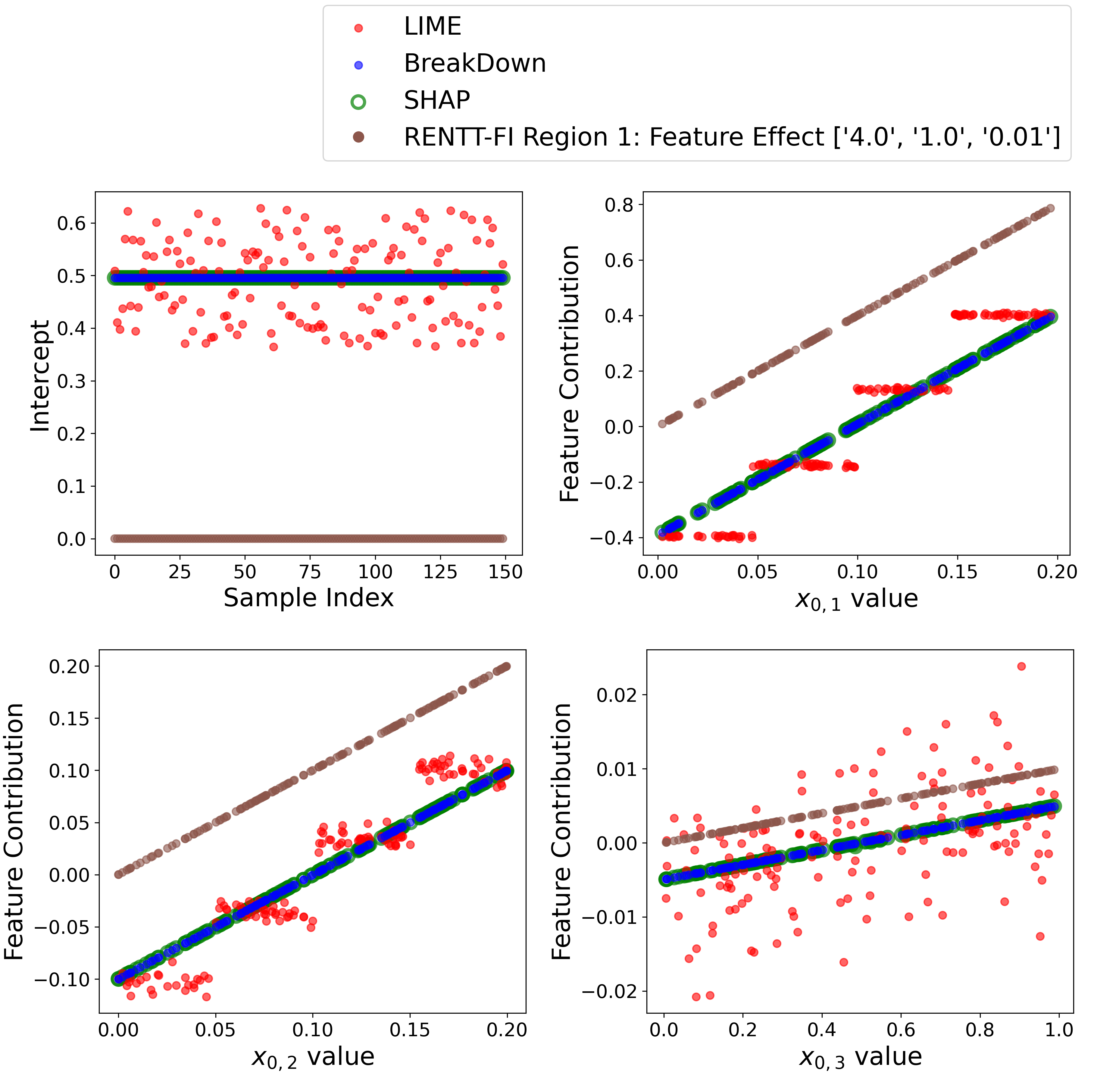}
 \caption{Intercept (top left) and \acp{fi} for the data set generated via the function in Equation~\eqref{eq:fi_linear}.}
 \label{fig:FI_lin_rmse}
\end{figure}
For the function given via Equation~\eqref{eq:fi_linear}, local and global \ac{fi} can be computed.
Here, the importance of the input features should reconstruct the initial function with the \ac{fi} of $\Feature_{0,0}$ being $4$, the \ac{fi} of $\Feature_{0,1}$ being $1$ and the \ac{fi} of $\Feature_{0,2}$ being $0.01$. This result is expected for both local and global \ac{fi} due to the linearity of the function used.

The global results for \ac{se}, \ac{sage}, and \ac{rentt-fi} are shown in Figure~\ref{fig:FI_lin_global}.
\Ac{rentt-fi} produces feature effect and contribution, where the contribution equals the feature effect multiplied by the average feature value.
In a more general case, this holds for all linear functions: Within a region, the contribution is the feature effect times the average feature value in this region. 
Both \ac{se} and \ac{sage} converge on different outcomes.
In these results, the influence of all features is widely underestimated ($0.057$ for $\Feature_{0,0}$ for \ac{se} and $0.054$ for \ac{sage}).
\ac{rentt-fi} produces the expected feature effect of $4$ for $\Feature_{0,0}$ and $0.39$ for the feature contribution, being close to the average value of $0.1$ times the effect with noise generated by the data sampling. 
Moreover, for \ac{se} and \ac{sage}, the relationships among the different \acp{fi} are misrepresented, as the most important feature receives a disproportionately higher share of the \ac{fi} compared to the function given via Equation~\eqref{eq:fi_linear}.
While providing similar values for Features $\Feature_{0,1}$ and $\Feature_{0,2}$, \ac{se} and \ac{sage} do not agree exactly, with an \ac{fi} of $3.4 \cdot 10^{-3}$ for \ac{se} and $2.7 \cdot 10^{-3}$ for \ac{sage} for $\Feature_{0,1}$ and $1.3 \cdot 10^{-5}$ and $3.5 \cdot 10^{-5}$ for $\Feature_{0,2}$ respectively.
For this data set, all global methods agree in their ranking of the important features (as shown in a value of $1.0$ for Krippendorff's $\alpha$ in Table~\ref{tab:FI_alpha_global_contribution} for all method pairs), while the differences in value result in an interval-scaled $\alpha$ of $0.998$ between \ac{se} and \ac{sage} and $0.112$ between \ac{rentt-fi} and \ac{se} and $0.104$ between \ac{rentt-fi} and \ac{sage}.
Similarly, as shown in Figure~\ref{fig:FI_RE_global_column}, the \ac{re} between \ac{se} and \ac{sage} is low ($7.7 \pm 2.7\,\%$), while the one between \ac{rentt-fi} and the other methods is significantly higher ($218 \pm 0.4\,\%$ for \ac{se} and \ac{rentt-fi}, $221 \pm 0.4\,\%$ for \ac{se} and \ac{rentt-fi}, Table~\ref{tab:rmse_relative_global} in Appendix~\ref{a-sec:additional-FI-results}).

 \begin{figure}[t]
     \centering
     \includegraphics[width=0.75\linewidth]{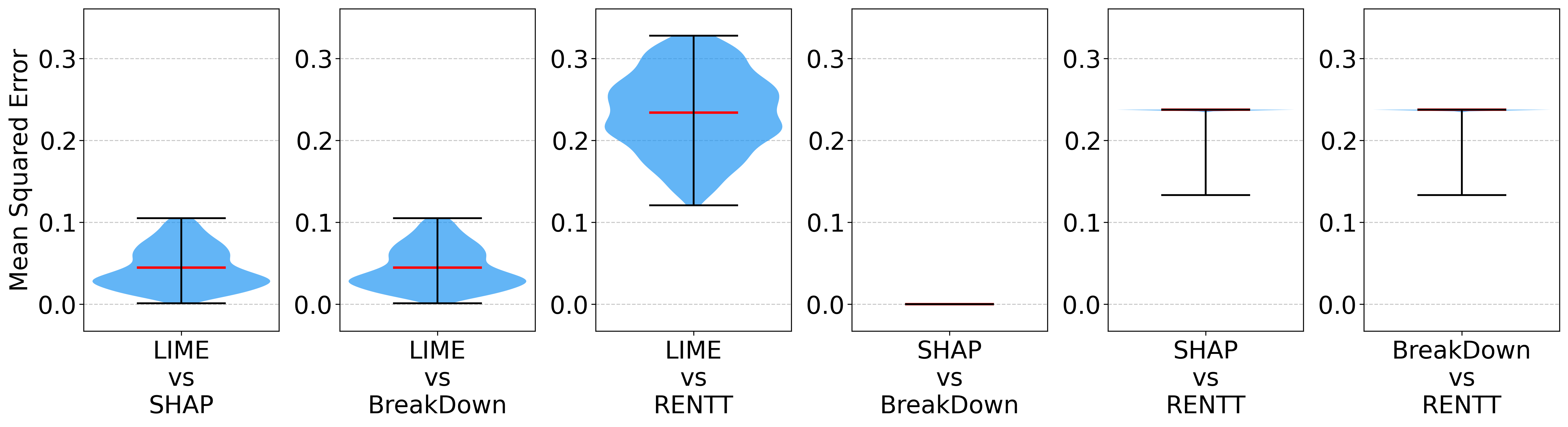}
     \caption{Violin plot of the \ac{rmse} between the different \ac{fi} methods for the data set generated via the function in Equation~\eqref{eq:fi_linear}. Only the 95th percentile of samples is pictured.}
     \label{fig:FI_reg_rmse_violin}
 \end{figure}

\begin{figure}[t]
    \centering
    \subfigure[Ordinal (ranking)]{%
        \label{subfig:FI_lin_krippendorf_ranking}%
        \includegraphics[width=0.75\textwidth]{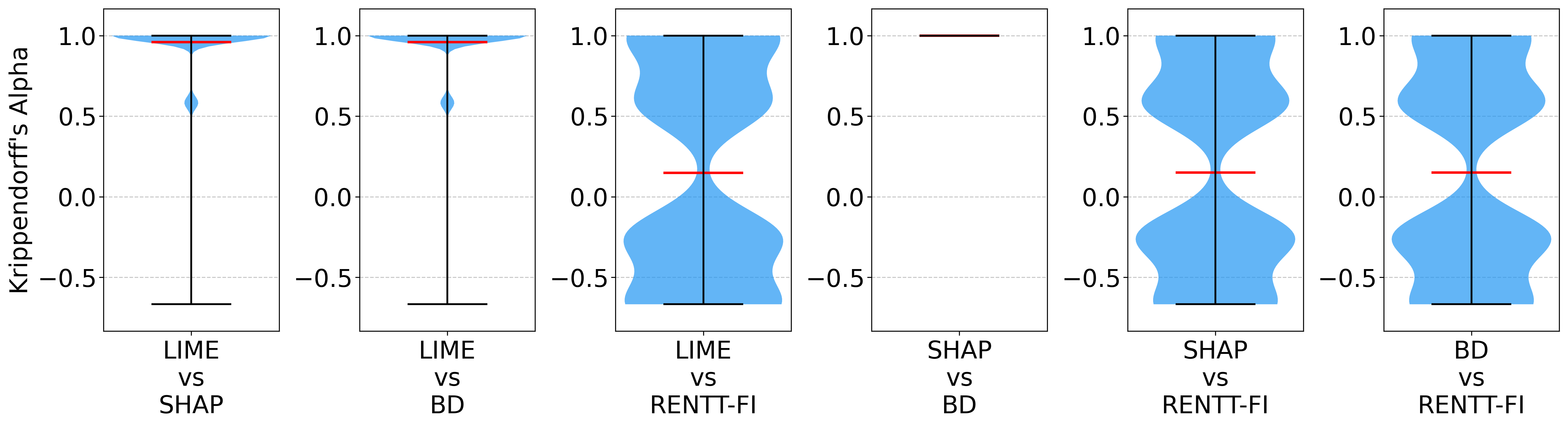}%
    }\hfill
    \subfigure[Interval (values)]{%
        \label{subfig:FI_lin_krippendorf_values}%
        \includegraphics[width=0.75\textwidth]{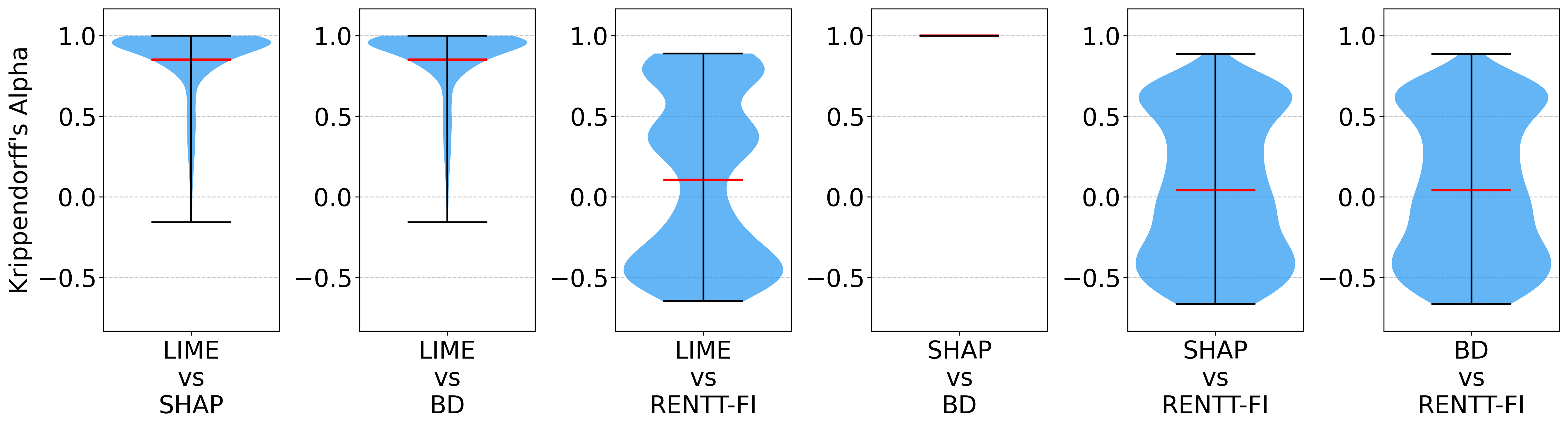}%
    }
    \caption{Violin plots of Krippendorff's $\alpha$, calculated samplewise between the different \ac{fi} methods for the data set generated via the function in Equation~\eqref{eq:fi_linear}. Only the 95th percentile of samples is pictured.}
    \label{fig:FI_lin_krippendorf_combined}
\end{figure}
The local \ac{fi} for the linear function given via Equation~\eqref{eq:fi_linear} yields largely the same findings as for the absolute function (Figure~\ref{fig:FI_reg_rmse_violin}):
\ac{shap}, \ac{bd} and \ac{rentt-fi} agree, with the offset in the \ac{rmse} between \ac{shap}/\ac{bd} and \ac{rentt-fi} created by the intercept (Figure~\ref{fig:FI_lin_rmse}).
This offset dominates the \ac{rmse}. 
A similar trend is observed with \ac{lime}, as \ac{lime} and \ac{shap}/\ac{bd} are closer together with an \ac{rmse} below $0.11$, whereas \ac{lime} and \ac{rentt-fi} yield \ac{rmse}-values between $0.11$ and $0.34$. 
For this data set, the differences between the evaluation methods become interesting:
The \ac{rmse} (Figure~\ref{fig:FI_reg_rmse_violin}), Krippendorff's $\alpha$ ordinal-scaled (Figure~\ref{subfig:FI_lin_krippendorf_ranking}) and interval-scaled (Figure~\ref{subfig:FI_lin_krippendorf_values}) show that \ac{lime}, \ac{shap} and \ac{bd} mostly agree in their ranking of the features, with the value agreement being slightly worse.
In contrast, \ac{rentt-fi} seems to provide quite different \ac{fi}-rankings and values with neither clear structural agreement or disagreement.
The explanation for this can be found once again within the intercept differing between the \ac{fi} methods (Figure~\ref{fig:FI_lin_rmse}).
The higher intercept of the \ac{fi} methods leads to lower \ac{fi}-values, since all methods roughly approximate the \ac{nn}-output via the sum of the \acp{fi} and the intercept (Figure~\ref{fig:FI_lin_rmse}).
This results in lower agreement in the interval-scaled $\alpha$-values, while not being reflected in the ordinal-scaled $\alpha$-values.


\subsubsection{Differences between Existing Methods and RENTT-FI across Different Data Sets} \label{sec:fi-method-comparison}
\begin{figure}[H]
    \centering
    \subfigure[Local \ac{fi} methods.]{%
        \label{fig:FI_RE_local_column}%
        \includegraphics[width=0.6\textwidth]{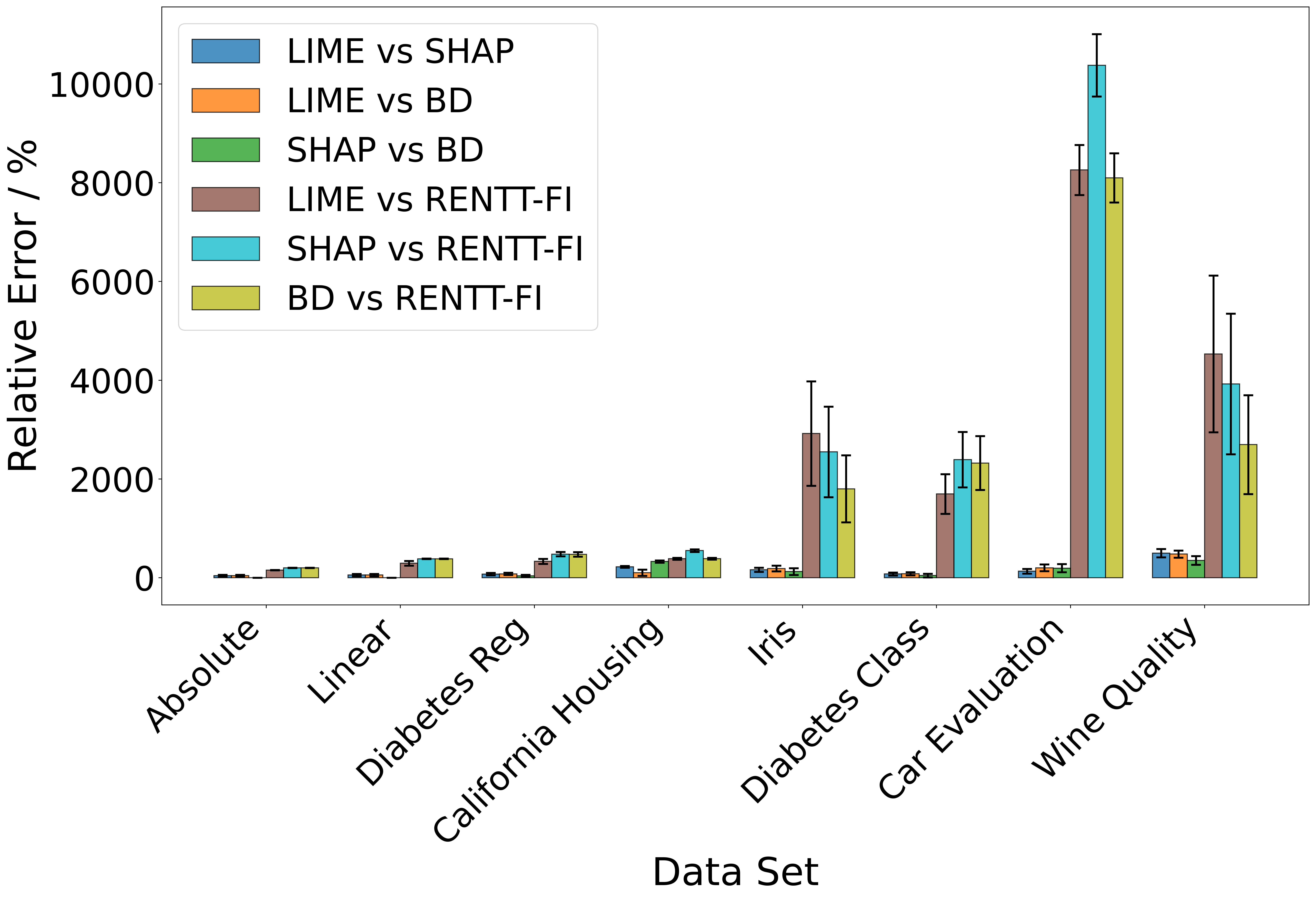}%
    }\hfill
    \subfigure[Global \ac{fi} methods.]{%
        \label{fig:FI_RE_global_column}%
        \includegraphics[width=0.39\textwidth]{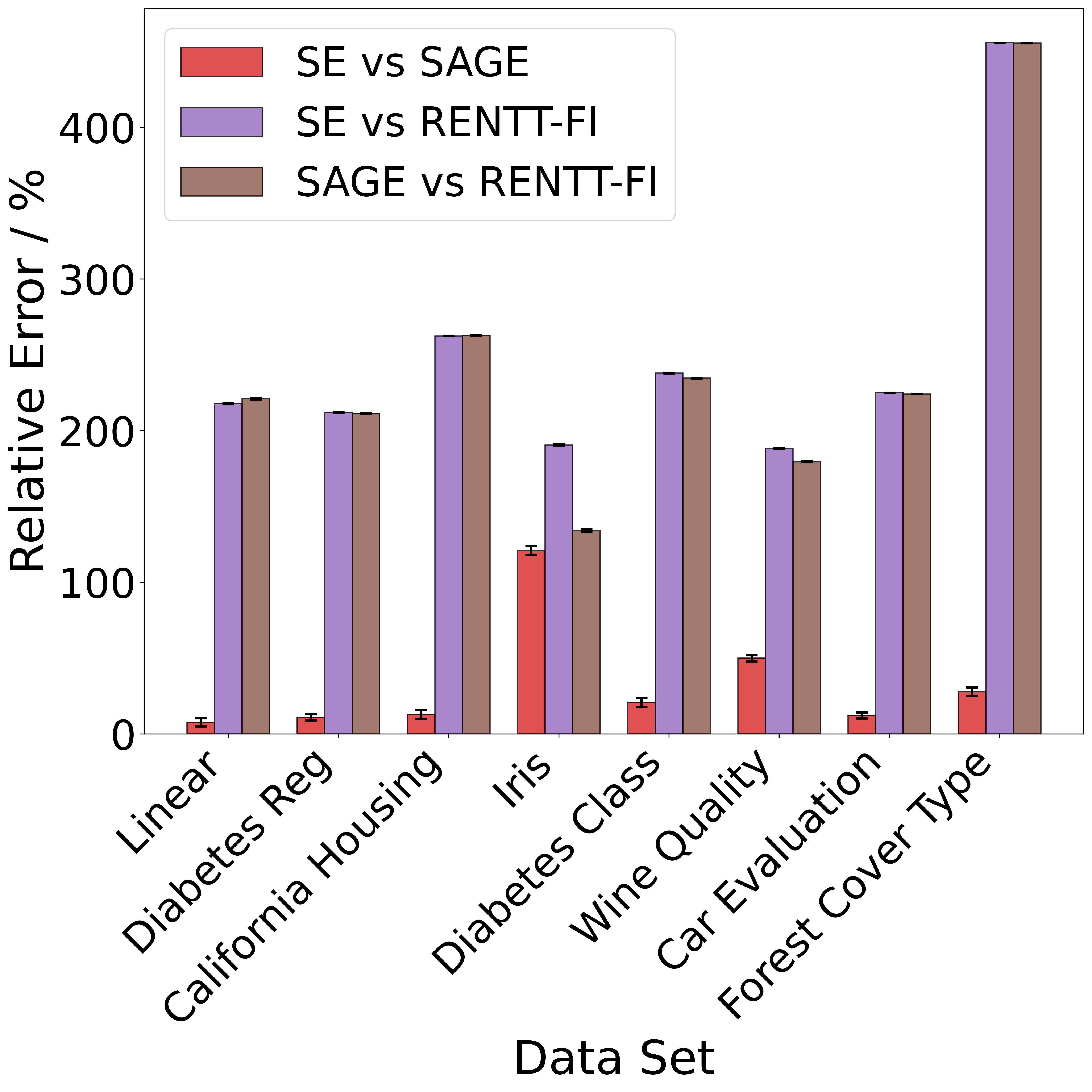}%
    }
    \caption{Comparative analysis of \ac{re} with standard deviation for local and global \ac{fi} methods across all data sets where \ac{fi} calculations were possible.}
    \label{fig:FI_RE_local_global}
\end{figure}

To evaluate whether existing \ac{fi} methods conform to ground truth explanations, \ac{shap}, \ac{lime} and \ac{bd} can be quantitatively compared to \ac{rentt-fi} across multiple data sets.
Figure~\ref{fig:FI_RE_local_global} as well as Tables~\ref{tab:rmse_relative_local_reg} and~\ref{tab:rmse_relative_local_cla} in Appendix~\ref{a-sec:additional-FI-results} present the \ac{rmse} and \ac{re} between the different local methods for regression and classification tasks, respectively.
For the local methods, it can be observed that for the regression tasks \ac{shap} and \ac{bd} produce the most similar results on the smallest data sets, with an \ac{re} of ($5.0 \pm 8.0) \cdot10^{-6}\,\%$ for the absolute function, $0.04 \pm 1.0\,\%$ for the linear one and $37 \pm 21\%$ for Diabetes Reg. 
As data set complexity increases (California Housing), \ac{lime} and \ac{bd} become the closest pair, with  \acp{re}  ranging around $100\,\%$.
Notably, \ac{rentt-fi} consistently produces substantially different values compared to all approximation methods, with  \acp{re}  between $(153 \pm 3) - (549 \pm 27)\,\%$.

In the classification tasks, all local methods generate significantly different results (lowest for \ac{shap}/\ac{bd} on Diabetes Class with $44 \pm 37\,\%)$, but a consistent pattern emerges: \ac{lime}, \ac{shap}, and \ac{bd} are substantially more similar to each other than any of them is to \ac{rentt-fi}.
The  \acp{re}  between  all methods but \ac{rentt-fi} range from $(44 \pm 37)-(496\pm84)\,\%$, while errors between these methods and and \ac{rentt-fi} reach $(1,696\pm401)-(10,379\pm632)\,\%$.
These differences are highly data set-dependent, with the Car Evaluation data set showing particularly large discrepancies for \ac{rentt-fi} (\ac{re} of $(8099\pm497)-(10,379\pm632)\,\%$).

For all used data sets, visualizations of Krippendorff's $\alpha$ and the \ac{rmse}, similar to Figures~\ref{fig:FI_reg_rmse_violin} and~\ref{fig:FI_lin_krippendorf_combined}, can be found in Appendix~\ref{a-sec:additional-FI-results}.
They mainly consolidate the previous findings of the agreement between \ac{fi}-methods being data set dependent with a trend towards \ac{shap} and \ac{bd} agreeing the most. 

For the global \ac{fi} methods, \ac{se} and \ac{sage} produce values that are orders of magnitude smaller than \ac{rentt-fi} (typically 0.001-0.03 vs 0.1-7.5 in Tables~\ref{tab:FI_global_reg} and~\ref{tab:FI_global_cla}).
Visualizations of the \ac{fi} values, similar to Figure~\ref{fig:FI_lin_global}, can be found in Appendix~\ref{a-sec:additional-FI-results}, in Figures~\ref{fig:FI_diabetes_reg_global} and~\ref{fig:FI_California_global} for the regression tasks and Figures~\ref{fig:FI_iris_global},~\ref{fig:FI_diabetes_global},~\ref{fig:FI_Wine_global},~\ref{fig:FI_CarEvaluation_global}, \ref{fig:FI_Covertype_global_contribution} and~\ref{fig:FI_Covertype_global_effect} for the classification tasks.
While \ac{se} and \ac{sage} show reasonable to high agreement in both sign and relative magnitude, meaningful systematic (dis)agreement ($|\alpha| > 0.7$) with \ac{rentt-fi} values (Table~\ref{tab:FI_alpha_global_contribution}) depends on the data set in question.
For the Linear, Diabetes Reg, Iris, Diabetes Class and Wine Quality, all methods agree in their feature ranking (ordinal $|\alpha| > 0.75$), but not in their \ac{fi} values (interval $|\alpha| < 0.6$).
The \ac{re} in Figure~\ref{fig:FI_RE_global_column} follows the same pattern of higher differences between \ac{rentt-fi} and \ac{sage}/\ac{se}.
For the global methods, the \ac{re} does not consistently increase or decrease between classification and regression tasks.
For the most complex data set, the Forest Cover Type, the \ac{re} is significantly higher ($455.3 \pm 0.0006\%$ between \ac{rentt-fi} and \ac{se}/\ac{sage}) than for the other data sets (at most $262.9 \pm 0.2\%$ between \ac{rentt-fi} and \ac{se}/\ac{sage} for California Housing).

These results demonstrate that while existing approximation methods show internal consistency patterns (with \ac{bd} and \ac{shap} aligning on small data sets), all diverge substantially from the ground truth explanations provided by \ac{rentt-fi}.
This divergence seems to increase with data set complexity.

\begin{table}[H]
    \caption{Pairwise Krippendorff's $\alpha$ reliability coefficients for global \ac{fi} methods.}
    \centering
    \begin{tabular}{|l|l|c|c|}
        \hline
        \textbf{Data Set} & \textbf{Comparison} & \textbf{$\alpha$ (Ordinal)} & \textbf{$\alpha$ (Interval)} \\
        \hline
        \multirow{3}{*}{Linear} & SE vs SAGE & 1.000 & 0.998 \\
         & SE vs RENTT-FI & 1.000 & 0.112 \\
         & SAGE vs RENTT-FI & 1.000 & 0.104 \\
        \hline
        \multirow{3}{*}{Diabetes Reg} & SE vs SAGE & 0.988 & 0.995 \\
         & SE vs RENTT-FI & 0.878 & 0.093 \\
         & SAGE vs RENTT-FI & 0.855 & 0.097 \\
        \hline
        \multirow{3}{*}{California Housing} & SE vs SAGE & 0.978 & 0.996 \\
         & SE vs RENTT-FI & 0.196 & 0.022 \\
         & SAGE vs RENTT-FI & 0.107 & 0.022 \\
        \hline
        \hline
        \multirow{3}{*}{Iris} & SE vs SAGE & 1.000 & 0.652 \\
         & SE vs RENTT-FI & 0.825 & 0.332 \\
         & SAGE vs RENTT-FI & 0.825 & 0.592 \\
        \hline
        \multirow{3}{*}{Diabetes Class} & SE vs SAGE & 0.978 & 0.989 \\
         & SE vs RENTT-FI & 0.978 & 0.082 \\
         & SAGE vs RENTT-FI & 0.955 & 0.098 \\
        \hline
        \multirow{3}{*}{Wine Quality} & SE vs SAGE & 0.971 & 0.891 \\
         & SE vs RENTT-FI & 0.829 & -0.503 \\
         & SAGE vs RENTT-FI & 0.769 & -0.452 \\
        \hline
        \multirow{3}{*}{Car Evaluation} & SE vs SAGE & 1.000 & 0.987 \\
         & SE vs RENTT-FI & 0.005 & 0.037 \\
         & SAGE vs RENTT-FI & 0.005 & 0.039 \\
        \hline
        \multirow{3}{*}{Forest Cover Type} & SE vs SAGE & 0.965 & 0.994 \\
         & SE vs RENTT-FI & -0.410 & 0.010 \\
         & SAGE vs RENTT-FI & -0.437 & 0.010 \\
        \hline
    \end{tabular}
    \label{tab:FI_alpha_global_contribution}
\end{table}

\begin{figure}[H]
    \centering
    \subfigure[Local \ac{fi} methods comparison.]{%
        \label{fig:FI_runtime_local}%
        \includegraphics[width=0.49\textwidth]{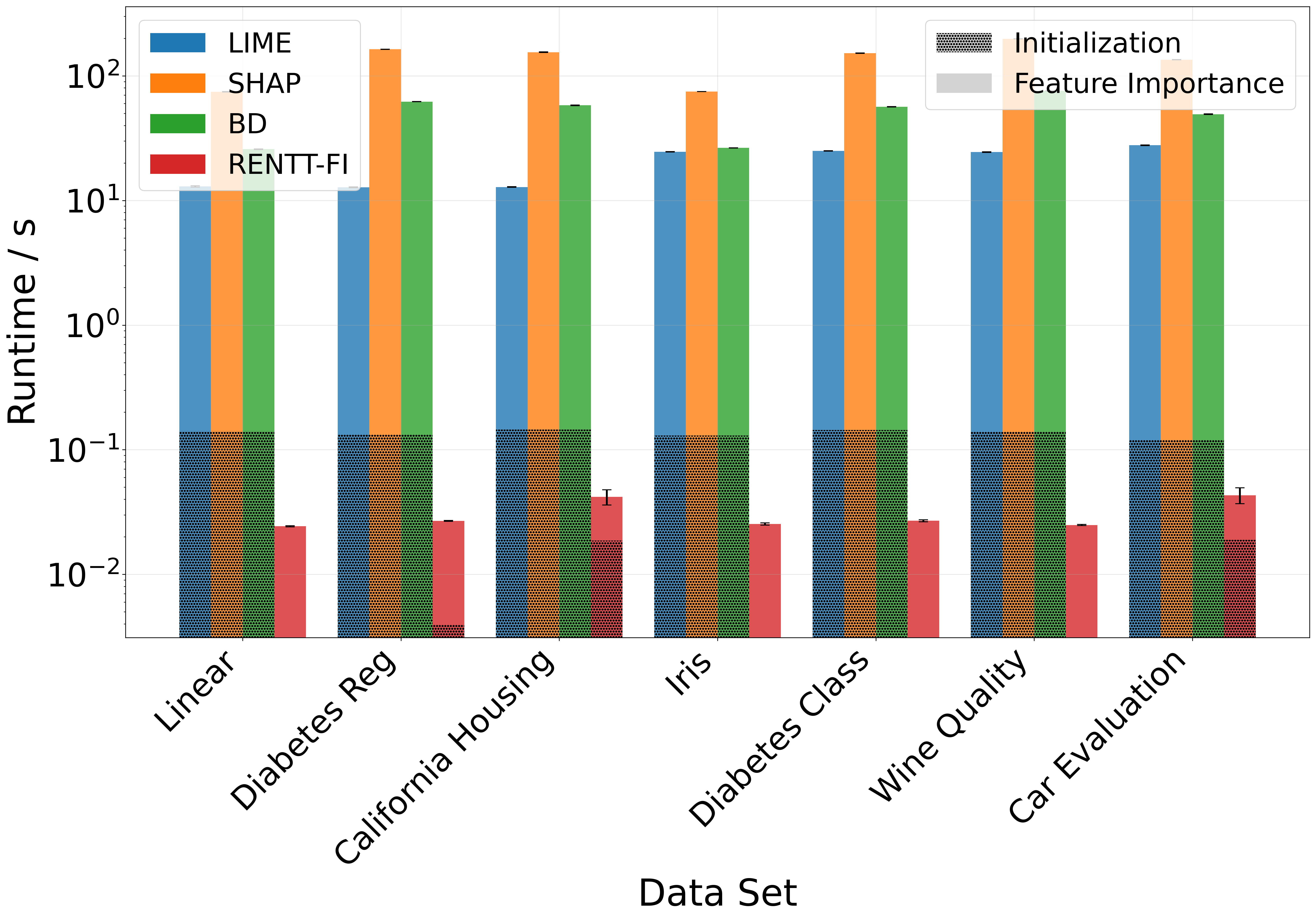}%
    }\hfill
    \subfigure[Global \ac{fi} methods comparison.]{%
        \label{fig:FI_runtime_global}%
        \includegraphics[width=0.48\textwidth]{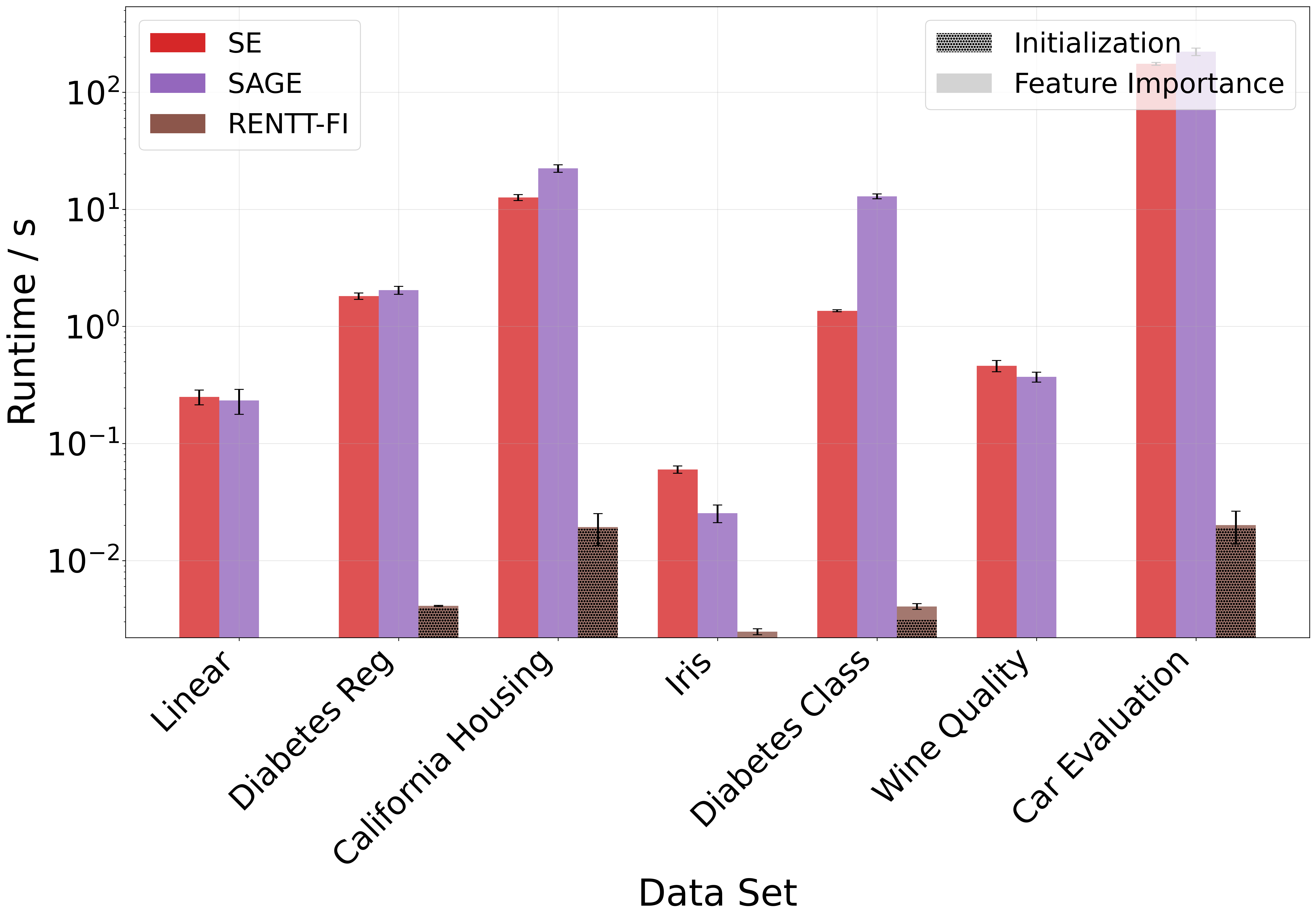}%
    }
    \caption{Runtime comparison of \ac{fi} methods. The hatched area indicates the initialization of the method---explainer initialization or in case of \ac{rentt-fi} the transformation of the \ac{nn} into \ac{dt}.}
    \label{fig:fi_runtime_overview}
\end{figure}

\subsubsection{Performance Comparison between existing Methods and RENTT-FI}\label{ssec:fi-runtime}
In this section, we explore the effectiveness of our \ac{fi} calculations, comparing its runtime against established methods such as \ac{lime}, \ac{shap}, and others. By providing both global and local \ac{fi} insights, we aim to illustrate how  \ac{rentt-fi} enhances interpretability without sacrificing performance.

The runtime comparison presented in Figure~\ref{fig:fi_runtime_overview} demonstrates the superior computational efficiency of \ac{rentt-fi} across all evaluated data sets and both local and global \ac{fi} methods. 
As illustrated in Figure~\ref{fig:FI_runtime_local}, \ac{rentt-fi} consistently achieves the fastest execution times for local explanations, producing these faster by several orders of magnitude compared to \ac{lime}, \ac{shap} and \ac{bd}. 
 \ac{rentt-fi} achieves runtime improvements of approximately 300-500× faster than \ac{lime}, 1,800-7,500× faster than \ac{shap}, and 600-2,900× faster than \ac{bd}. Remarkably, \ac{rentt-fi} maintains sub-millisecond execution times (0.02-0.04 seconds) regardless of data set complexity, while traditional methods require tens to hundreds of seconds.

Similarly, Figure~\ref{fig:FI_runtime_global} reveals that \ac{rentt-fi} maintains its computational advantage for global \ac{fi} calculations, substantially outperforming \ac{se} and \ac{sage} methods. 
The consistent performance across diverse data sets validates the scalability and robustness of our approach.
The detailed numerical results supporting these visualizations are provided in Appendix~\ref{a-ssec:runtime} (Tables~\ref{tab:runtime_local} and~\ref{tab:runtime_global}).

\subsection{Discussion of the Results}
This chapter presented a comprehensive evaluation of \ac{fi} methods through systematic experiments comparing \ac{rentt-fi} with established explanation methods.
The analysis revealed limitations in current XAI approaches that have significant implications for practical applications. 

Most methods failed the implementation-invariance test, producing significantly different \ac{fi} values for functionally equivalent models (\ac{nn} and \ac{dt}), especially in classification tasks (Section~\ref{sec:fi-nn-vs-dt}).
These findings demonstrate that while existing methods exhibit internal consistency patterns under specific conditions, they systematically fail to capture the true \ac{fi} as determined by the network's mathematical structure (Section~\ref{sec:fi-toy}).
The divergence from ground truth explanations reveals fundamental limitations in current XAI approaches that cannot be resolved through methodological refinements alone, highlighting the critical need for mathematically grounded explanation techniques like \ac{rentt-fi}.

For local methods, it can be observed that \ac{shap} and \ac{bd} showed strong agreement on smaller data sets (Absolute, Linear, Diabetes Reg), \ac{lime} and \ac{bd} became the most similar methods on larger data sets (California Housing), but with rather high \ac{re} (Section~\ref{sec:fi-method-comparison}).
All methods diverged substantially from \ac{rentt-fi}, with \acp{re} values consistently exceeding $150\,\%$ across data sets and reaching up to $10,400\,\%$ in classification tasks.
This divergence increased with data set complexity.
For the global explanation methods, \ac{se} and \ac{sage} produced values orders of magnitude smaller than \ac{rentt-fi} with no meaningful overlap or common structure in actual importance values.

One important consideration is the computational cost distribution for different explanation methods (Section~\ref{ssec:fi-runtime}). For \ac{rentt-fi}, the transformation process represents the most computationally expensive component, depending on network size with minimal cost for small networks. Subsequently, calculating Feature Effects for each sample and determining feature contributions is relatively inexpensive, with costs scaling primarily with the number of features. As such, the relative computational burden of \ac{rentt-fi} decreases when explaining many samples of a data set.
In contrast, methods such as \ac{lime}, \ac{shap}, and \ac{bd} require an initialization phase where the explanation infrastructure is prepared---including creating model wrappers and storing reference data for baseline comparisons. While this initialization cost varies by implementation, the subsequent \ac{fi} calculations represent the more computationally expensive phase for these methods.
Our runtime analysis demonstrate a significant improvement in \ac{fi} methodology: \ac{rentt-fi} reduces the computational effort while maintaining correctness. \Ac{rentt-fi} provides explanability insights with lower computational overhead compared to existing approaches.

%% file: Chapter/7_Conclusion.tex
\section{Discussion and Limitations} 
\label{sec:discussion-limitations}
In this section, we discuss the current limitations of our methodology and outline promising directions for future development.

The transformation's exactness of \ac{rentt} fundamentally depends on Assumption~\ref{ass:activationfunction}, which restricts the approach to piecewise linear activation functions when 100\,\% correctness is an essential requirement.
Some commonly used activation functions in modern deep learning, such as sigmoid or tanh, cannot be directly transformed without approximation. While piecewise linear approximation is possible, this creates a trade-off between correctness and computational cost---more linear segments improve approximation quality but increase the tree size exponentially. 

Memory complexity represents the most critical bottleneck for practical applications of \ac{rentt}, especially when dealing with larger network architectures. Despite significant improvements over existing methods like ECDT---with memory scaling at $\mathcal{O}(x^{1.7})$ to $\mathcal{O}(x^{1.9})$ versus $\mathcal{O}(x^{8.6})$ to $\mathcal{O}(x^{8.7})$---absolute memory requirements remain substantial. Networks with $10^4$ hidden neurons require 31-630 GB of RAM, limiting applicability on standard hardware.
The choice of activation functions critically impacts memory efficiency, as even inherently piecewise linear functions with more than two linear regions increase computational complexity. According to Equation~\eqref{eq:node}, the number of decision tree nodes scales with  the number of linear regions $\LinearRegions_i$ of the activation function in the layer $i$ (as specified in Assumption 3.1.3) as an exponent.
For optimal performance, ReLU-family activation functions (ReLU, Leaky ReLU, etc.) should be used, which have only two linear regions. This approach fortunately aligns with common practices in modern deep learning where ReLU and its variants dominate. When other activation functions are necessary, the number of approximation segments should be balanced against available computational resources carefully. 
To overcome this limitation while enhancing runtime efficiency, future work should extend beyond the pruning techniques used in our approach by integrating parallelization strategies for distributed computation, optimizing data structures for memory efficiency, and developing enhanced model serialization methods for effective resource management. By integrating these improvements, the transformation framework would be capable of processing larger datasets and more complex models with substantially reduced execution time and memory overhead, thereby achieving better scalability.

We noticed for the \ac{fi} calculations that \ac{rentt-fi} has lower computational costs on smaller to medium-sized networks compared to existing approaches. An important research question emerges regarding the \ac{nn} size and sample number threshold beyond which \ac{rentt-fi}'s transformation costs surpass those of conventional \ac{fi} methods to determine the optimal \ac{nn} size ranges and sample numbers  for \ac{rentt-fi} deployment.

To reduce runtime and memory complexity, we use ante-hoc pruning of the \acp{dt}.
This pruning removes activation patterns that are not used throughout the relevant data set (here training and testing data).
For these data sets, the transformation with pruning is exact, resulting in the exact same predictions for all data samples.
A key limitation arises when unseen data samples during deployment trigger activation patterns that were inactive during training. These patterns, having been pruned from the \ac{dt}, create missing pathways that prevent proper inference. While such pathways could be dynamically added to the \ac{dt}, this would necessitate extracting weights from the original \ac{nn}, which must therefore also be retained in memory.
To prevent such issues, a more complex and less efficient pruning could be used, e.g., by only removing activation patterns that are impossible to reach.
Such a pruning approach would allow for perfect exactness even for unseen data samples, but would increase memory and runtime complexity significantly.

Currently, the implementation of this transformation is limited to \acp{fcnn}. Expanding the implementation to include \acp{cnn} and \acp{rnn} as described in Remark~\ref{remark:implementation} will extend the applicability of the approach to a wider range of neural network architectures, ensuring that diverse models can benefit from enhanced interpretability and explainability.

\ac{rentt} transforms \acp{nn} into equivalent \acp{dt}.
Ideally, this would result in directly interpretable models.
But due to the resulting decision trees being multivariate, they are too complex to understand.
As such, further simplifications are necessary to use these trees as explanations.
We chose to represent the \acp{dt} via \ac{fi}, as this was computationally efficient, a common explanation type and did not result in a loss of correctness.
Further explanation types enabled by the \acp{dt} (see Section~\ref{sec:fi-definition}) could further improve the interpretability of \acp{nn}.

As commented on throughout this paper, some problems arise when comparing different \ac{fi} methods and \ac{xai} methods in general.
While \ac{rentt-fi} separates feature contribution and feature effect clearly, this distinction is often not as clear for other approaches.
Especially for the global methods, it is not specified whether their results should be interpreted as effect or contribution.
We chose to compare the feature contribution of \ac{rentt-fi} to \ac{se} and \ac{sage}, to maintain consistency with the local methods. 
Depending on the specific question a user is interested in, the feature effect might provide a clearer interpretation for the global case.

As an extension to the existing work of evaluating \ac{xai} methods via \ac{rentt-fi}, metrics for the correctness of \ac{xai} could also be evaluated using \ac{rentt-fi}.
Previously, such metrics have been shown to disagree systematically~\cite{tomsett2019sanityformetrics}.
With \ac{rentt-fi}, it can be assessed which correctness metric best reflects the ground truth \ac{fi}.
In cases where \ac{rentt-fi} is not feasible due to runtime or memory complexity, the metric that best agreed with \ac{rentt-fi} for simpler models or data sets could be used to choose between feasible \ac{xai} methods.

While theoretical benefits for \ac{xai} methods are desirable, explanations are directed towards humans.
As such, their main goals depend on downstream impact, e.g, by empowering humans to spot incorrect model decisions or provide new modes of scientific discovery.
In general, many \ac{xai} papers lack user studies to evaluate their proposed methods~\cite{suh2025noXAIStudies}.
For \ac{rentt-fi}, a user study could provide interesting insights into how the created explanations are perceived and whether their theoretical soundness create benefits for downstream goals.

The local linear models underlying our transformation represent an underutilized resource for deeper insights into neural network behavior. While our current applications focus primarily on \ac{fi} calculation, these models could enable broader explainability applications that remain unexplored, such as counterfactual analysis to generate ``what-if'' scenarios.
Further exploration of the local linear models used for this transformation could provide deeper insights into the decision-making processes of neural networks. By examining these models in detail, we can enhance explainability, offering more granular perspectives on how decisions are made within the network. 
Additionally, local linear models have potential applications beyond \ac{xai}, such as improving robustness. Understanding decision boundaries and gradients near these boundaries could be crucial for developing models that are resistant to adversarial attacks and other perturbations.

Within \acp{nn}, usually different activation patterns are not tracked.
Consequently, it is difficult to specify whether a model is sufficiently trained or whether all input regions contain adequate data samples to ensure robust decisions.
The \ac{rentt}-\acp{dt} automatically track the number of data samples corresponding to each activation pattern.
This enables the identification of underutilized and thus, possibly undertrained activation patterns, for which more data samples would improve decision robustness.
In the same vein, the pruning method applied to the decision trees could be utilized: If pruning yields a simpler representation of the \acp{nn}, this representation could be used for inference to decrease runtime and memory complexity.
Our experiments reveal that non-linear neural networks sometimes defaulted to a linear representation in the \acp{dt}, indicating that the used models might have been too complex for the data set in question.

\section{Conclusion}
\label{sec:conclusion}
The conversion of \acp{nn} into \acp{dt} offers a practical solution for improving model interpretability and explanability. By transforming \acp{nn} including \acp{fcnn}, \acp{cnn} with convolution and pooling layers and \acp{rnn} with general activation functions into multivariate \acp{dt}, this approach maintains the original network's functionalities while making the decision-making process more understandable.
This allows to circumvent the often stated tradeoff between accuracy and interpretability.
The linearization of \acp{nn}, especially with piecewise linear activation functions, simplifies network operations into linear transformations, enabling their mapping to \ac{dt} structures.
To address scalability challenges, efficient implementation techniques and ad hoc pruning methods are used to manage large networks. These strategies reduce memory consumption and runtime, making the transformation suitable for large-scale applications. 
This method effectively enhances explainability in \ac{nn} architectures by incorporating feature importance calculations globally, regionally, and locally. This provides valuable insights into the influence of input features on model decisions, aiding in understanding model behavior.
This allows for downstream applications such as ensuring fairness of models by showing the influence of sensitive attributes such as age or gender. 
In conclusion, transforming \acp{nn} into \acp{dt} allows performance and interpretability, making it easier to analyze and understand complex models. This method contributes to the goal of making \acp{nn} more accessible and reliable for diverse applications, providing essential tools for diagnostics and feature importance analysis. As \acp{nn} are increasingly utilized in critical domains, the ability to interpret and trust their decisions is vital for their effective deployment.

%% file: Chapter/Appendix_cleaned.tex
\newpage


\section{Data Sets} \label{a-sec:data-sets}
For the different experiments, we used a set of common data sets.
Where possible and not stated otherwise, the data sets from scikit-learn were used.
For regression tasks, we used these data sets:
\begin{itemize}
    \item The Diabetes Reg data set contains data collected from 442 patients, with each patient having ten features related to diabetes progression one year after baseline. The features include Age, Sex, Body Mass Index, Average Blood Pressure and six blood serum measurements. From these inputs, the disease progression within a year should be predicted as a continuous value.
    \item The California Housing Data Set contains information about housing prices in California, based on data collected from various block groups in the state. Overall, 20,640 samples with eight features are present, with the features Median Income (MedInc), House Age, Latitude, Longitude, Population (all related to one block group), Average Number of Rooms (AveRooms), Average Number of Bedrooms (AveBedrms), Average Number of Occupants (AveOccup). From these, the median house value within a block group should be predicted.
\end{itemize}
Since \ac{rentt-fi} can also be used for classification tasks, further data sets were included for evaluation:
\begin{itemize} 
    \item The Iris Data Set consists of 150 samples of iris flowers, each belonging to one of three species: Iris setosa, Iris versicolor, and Iris virginica. The data set includes four features measured for each sample: sepal length, sepal width, petal length, and petal width, all measured in centimeters. Based on these features, the species of flower should be predicted.
    \item A number of different Diabetes data sets exist, also for classification. The one used here (called Diabetes Class) contains 768 samples with eight features such as Number of Pregnancies, Plasma Glucose Concentration, Body Mass Index and Age. The task is a binary classification of whether a patient will be tested positive for diabetes. The version of kaggle is used\footnote{\url{https://www.kaggle.com/datasets/mathchi/diabetes-data-set}}, which is a subset of an older data set created by~\cite{Smith1988Diabetes}.
    \item The Wine Quality Data Set consists of 178 samples of wine, each described by 13 features derived from chemical analyses, such as Alcohol Content, Ash and Magnesium. Based on these features, the samples can be categorized into three different wine quality types, low, medium, and high.
    \item The Car Evaluation Data Set contains 1,728 data samples of cars which are sorted into four categories regarding their acceptability: unacceptable, acceptable, good and very good. For this, six features can be used, which contain information about the price of the car, maintenance costs, number of doors, person capacity, luggage boot size and a safety rating. Here, the version by~\cite{bohanec1988CarEvaluation} is used.
    \item Since \ac{rentt} is constructed to also work for larger and more complex \ac{nn}, the Forest Cover Type data set is also used, which requires a larger \ac{nn} than the other data sets to achieve acceptable accuracy. It contains information on the types of forest cover for 30x30m cells in the Rocky Mountain region of the United States~\cite{Blackard1998CoverType}. It includes 581,012 samples with 54 features describing various properties of the areas, such as elevation, distances to hydrology and soil type. With these, one of seven classes of forest cover type should be predicted, with classes such as spruce-fir, lodgepole pine and ponderosa pine.
    \end{itemize}

\section{Performance of the Neural Network Models on the Data Sets}
\label{app:performance}
The results for the performance of the \ac{nn} are shown in Table~\ref{tab:accuracy_comparison_reg} for the regression performance and Table~\ref{tab:accuracy_comparison_class} for the classification performance. 
Both tables show that there are no differences in performance between \ac{nn} and \ac{dt}, which provides evidence that the theoretical equivalence of both models also holds in practice.

\begin{table}[H]
\centering
\caption{Number of neurons in the hidden layer of the \ac{nn}, performance of \ac{nn} and \ac{dt} on the regression data sets and number of nodes of the pruned decision trees.}
\label{tab:accuracy_comparison_reg}
\begin{tabular}{|c|p{2cm}|c|c|c|}
\hline
\textbf{Data Set} & \textbf{Neurons in Hidden Layers}& \textbf{\Ac{nn} MSE}  & \textbf{\Ac{dt} MSE} & \textbf{Number of Nodes} \\
\hline
Absolute & $8,\ 4$ & $3.43\cdot 10^{-10}$ & $3.43\cdot 10^{-10}$ & $39$ \\
Linear & $8,\ 4$ & $2.53\cdot 10^{-7}$ & $2.53\cdot 10^{-7}$ & $48$\\
Diabetes Reg & $8,\ 4$ & $0.03$ & $0.03$ & $201$ \\
California Housing & $8,\ 4$ & $0.015$ & $0.015$ & $159$\\
\hline
\end{tabular}
\end{table}

\begin{table}[H]
\centering
\caption{Number of neurons in the hidden layer of the \ac{nn}, performance of \ac{nn} and \ac{dt} on the classification data sets and number of nodes of the pruned decision trees.}
\label{tab:accuracy_comparison_class}
\begin{tabular}{|c|p{2cm}|c|c|c|}
\hline
\textbf{Data Set} & \textbf{Neurons in Hidden Layers}& \textbf{\Ac{nn} Acc/ \%}  & \textbf{\Ac{dt} Acc/ \%} & \textbf{Number of Nodes} \\
\hline
Iris & $8,\ 4$ & $77$ & $77$ & $114$ \\
Diabetes Class & $8,\ 4$ & $75$ &  $75$ & $169$\\
Wine Quality & $8,\ 4$ & $92$  & $92$ & $68$ \\
Car Evaluation & $8,\ 4$ & $94$ & $94$ & $398$  \\
Forest Cover Type & $64,\ 32$ & $88$ & $88$ & $7,638,684$ \\
\hline
\end{tabular}
\end{table}

\section{Global Feature Importance Values}
In Tables~\ref{tab:FI_global_reg} and~\ref{tab:FI_global_cla}, the \ac{fi} values for the global explanation methods are presented, for regression tasks in Table~\ref{tab:FI_global_reg} and for classification tasks in Table~\ref{tab:FI_global_cla}.
Both tables contain the values for the global methods \ac{se} and \ac{shap} both on a \ac{nn} and on an equivalent \ac{dt} generated via \ac{rentt} and the \ac{rentt}-\ac{fi} results.
\begin{landscape}

\begin{table}[H]
    \caption{\ac{fi} results for the global methods between \ac{nn} and \ac{dt} on regression data.}
    \label{tab:FI_global_reg}
    \centering
    \begin{tabular}{|c|c|c|c|c|c|c|}
    \hline
    \textbf{Data Set} & \textbf{Feature} & \textbf{\ac{se} \Ac{dt}} & \textbf{\ac{se} \Ac{nn}} & \textbf{\ac{sage} \Ac{dt}} & \textbf{\ac{sage} \Ac{nn}} & \textbf{\ac{rentt}-\ac{fi}} \\
\hline
\multirow{3}{*}{linear}
& $x_{1}$ & $0.055 \pm 0.001$ & $0.055 \pm 0.001$ & $0.055 \pm 0.001$ & $0.053 \pm 0.001$ & $0.39$ \\
& $x_2$ & $0.0035 \pm 0.0005$ & $0.0045 \pm 0.0005$ & $0.0025 \pm 0.0005$ & $0.0029 \pm 0.0004$ & $0.1$ \\
& $x_3$ & $(0.3 \pm 2.5) \cdot 10^{-5}$ & $(2.8 \pm 2.5) \cdot 10^{-5}$ & $(0.05 \pm 2.5) \cdot 10^{-5}$ & $(2.0 \pm 2.1) \cdot 10^{-5}$ & $0.005$ \\
\hline     \multirow{10}{*}{}
& age & $(2.0 \pm 0.6) \cdot 10^{-4}$ & $(2.5 \pm 0.5) \cdot 10^{-4}$ & $(1.2 \pm 0.7) \cdot 10^{-4}$ & $(2.1 \pm 0.8) \cdot 10^{-4}$ & $-0.052$ \\
& sex & $(1.3 \pm 1.2) \cdot 10^{-4}$ & $(1.0 \pm 1.2) \cdot 10^{-4}$ & $(2.4 \pm 1.5) \cdot 10^{-4}$ & $(3.5 \pm 1.6) \cdot 10^{-4}$ & $-0.085$ \\
& bmi & $0.0085 \pm 0.0002$ & $0.0087 \pm 0.0002$ & $0.0098 \pm 0.0003$ & $0.0096 \pm 0.0003$ & $0.12$ \\
& bp & $0.0056 \pm 0.0002$ & $0.0055 \pm 0.0002$ & $0.0054 \pm 0.0002$ & $0.0054 \pm 0.0002$ & $0.12$ \\
Diabetes & s1 & $(-6.1 \pm 0.4) \cdot 10^{-4}$ & $(-5.2 \pm 0.4) \cdot 10^{-4}$ & $(-5.5 \pm 0.5) \cdot 10^{-4}$ & $(-5.9 \pm 0.5) \cdot 10^{-4}$ & $-0.037$ \\
Reg & s2 & $(-6.6 \pm 0.7) \cdot 10^{-4}$ & $(-6.2 \pm 0.8) \cdot 10^{-4}$ & $(-9.9 \pm 0.9) \cdot 10^{-4}$ & $(-9.5 \pm 1.1) \cdot 10^{-4}$ & $-0.069$ \\
& s3 & $0.0017 \pm 0.0001$ & $0.0019 \pm 0.0001$ & $0.0017 \pm 0.0001$ & $0.0015 \pm 0.0001$ & $-0.044$ \\
& s4 & $0.0036 \pm 0.0001$ & $0.0036 \pm 0.0001$ & $0.0034 \pm 0.0001$ & $0.0034 \pm 0.0001$ & $0.051$ \\ 
& s5 & $0.0101 \pm 0.0003$ & $0.0095 \pm 0.0002$ & $0.0096 \pm 0.0002$ & $0.0098 \pm 0.0003$ & $0.19$ \\
& s6 & $0.0014 \pm 0.0001$ & $0.0014 \pm 0.0001$ & $0.0013 \pm 0.0001$ & $0.0014 \pm 0.0001$ & $0.045$ \\
\hline
& MedInc & $0.0268 \pm 0.0002$ & $0.0263 \pm 0.0002$ & $0.0270 \pm 0.0002$ & $0.0266 \pm 0.0002$ & $0.28$ \\
& HouseAge & $0.00129 \pm 0.00005$ & $0.00137 \pm 0.00005$ & $(9.6 \pm 0.5) \cdot 10^{-4}$ & $(8.7 \pm 0.5) \cdot 10^{-4}$ & $0.11$ \\
& AveRooms & $0.0054 \pm 0.0008$ & $0.0076 \pm 0.0009$ & $0.0078 \pm 0.0009$ & $0.0073 \pm 0.0009$ & $-0.073$ \\
California & AveBedrms & $-0.0069 \pm 0.0008$ & $-0.0092 \pm 0.0009$ & $-0.0101 \pm 0.0009$ & $-0.0094 \pm 0.0009$ & $0.079$ \\
Housing & Population & $(-1.4 \pm 0.3) \cdot 10^{-4}$ & $(-1.7 \pm 0.3) \cdot 10^{-4}$ & $(-1.8 \pm 0.2) \cdot 10^{-4}$ & $(-2.1 \pm 0.2) \cdot 10^{-4}$ & $0.014$ \\
& AveOccup & $0.0032 \pm 0.0001$ & $0.0033 \pm 0.0001$ & $0.0025 \pm 0.0001$ & $0.0025 \pm 0.0001$ & $-0.15$ \\
& Latitude & $0.0107 \pm 0.0003$ & $0.0117 \pm 0.0003$ & $0.0079 \pm 0.0003$ & $0.0094 \pm 0.0003$ & $-0.36$ \\
& Longitude & $(1.2 \pm 3.3) \cdot 10^{-4}$ & $(-5.6 \pm 2.9) \cdot 10^{-4}$ & $0.0014 \pm 0.0003$ & $(3.6 \pm 3.0) \cdot 10^{-4}$ & $-0.66$ \\
\hline
\end{tabular}
\end{table}

\begin{table}[H]
    \caption{\ac{fi} results for the global methods between \ac{nn} and \ac{dt} on classification data.}
    \label{tab:FI_global_cla}
    \centering
    \begin{tabular}{|c|c|c|c|c|c|c|}
    \hline
    \textbf{Data Set} & \textbf{Feature} & \textbf{\ac{se} \Ac{dt}} & \textbf{\ac{se} \Ac{nn}} & \textbf{\ac{sage} \Ac{dt}} & \textbf{\ac{sage} \Ac{nn}} & \textbf{\ac{rentt}-\ac{fi}} \\
   \hline     \multirow{4}{*}{Iris}
& sepal length & $-0.0052 \pm 0.0004$ & $-0.0054 \pm 0.0003$ & $-0.013 \pm 0.001$ & $-0.013 \pm 0.001$ & $-0.052$ \\
& sepal width & $0.030 \pm 0.002$ & $0.029 \pm 0.002$ & $0.014 \pm 0.006$ & $0.029 \pm 0.005$ & $-0.62$ \\
& petal length & $0.040 \pm 0.001$ & $0.040 \pm 0.001$ & $0.101 \pm 0.003$ & $0.097 \pm 0.003$ & $0.17$ \\
& petal width & $0.186 \pm 0.004$ & $0.201 \pm 0.004$ & $0.46 \pm 0.01$ & $0.46 \pm 0.01$ & $0.89$ \\
\hline
& Pregnancies & $0.0040 \pm 0.0004$ & $0.0045 \pm 0.0003$ & $0.0051 \pm 0.0004$ & $0.0056 \pm 0.0004$ & $0.095$ \\
& Glucose & $0.108 \pm 0.003$ & $0.104 \pm 0.003$ & $0.112 \pm 0.003$ & $0.118 \pm 0.003$ & $1.37$ \\
& Blood Pressure & $-0.0046 \pm 0.0006$ & $-0.0042 \pm 0.0007$ & $-0.0069 \pm 0.0007$ & $-0.0065 \pm 0.0007$ & $-0.61$ \\
Diabetes & SkinThickness & $0.0057 \pm 0.0003$ & $0.0058 \pm 0.0004$ & $0.0054 \pm 0.0004$ & $0.0050 \pm 0.0004$ & $0.17$ \\
Class & Insulin & $-0.0014 \pm 0.0005$ & $-0.0014 \pm 0.0005$ & $(-2.0 \pm 0.5) \cdot 10^{-4}$ & $-0.0016 \pm 0.0005$ & $-0.13$ \\
& BMI & $0.033 \pm 0.001$ & $0.033 \pm 0.001$ & $0.037 \pm 0.002$ & $0.039 \pm 0.002$ & $0.59$ \\
& Pedigree & $0.0099 \pm 0.0008$ & $0.0100 \pm 0.0008$ & $0.0099 \pm 0.0009$ & $0.0105 \pm 0.0009$ & $0.21$ \\
& Age & $0.042 \pm 0.002$ & $0.044 \pm 0.002$ & $0.041 \pm 0.002$ & $0.040 \pm 0.002$ & $0.35$ \\
\hline
  & alcohol & $(-8.1 \pm 3.5) \cdot 10^{-4}$ & $(-7.2 \pm 3.2) \cdot 10^{-4}$ & $-0.0059 \pm 0.0006$ & $-0.0048 \pm 0.0004$ & $0.31$ \\
& malic acid & $0.024 \pm 0.001$ & $0.022 \pm 0.001$ & $0.022 \pm 0.002$ & $0.025 \pm 0.001$ & $0.38$ \\
& ash & $0.0026 \pm 0.0007$ & $0.0063 \pm 0.0009$ & $-0.018 \pm 0.002$ & $-0.014 \pm 0.001$ & $0.65$ \\
& alcalinity & $0.035 \pm 0.002$ & $0.037 \pm 0.001$ & $0.050 \pm 0.002$ & $0.049 \pm 0.002$ & $1.01$ \\
& magnesium & $(-4.0 \pm 3.2) \cdot 10^{-4}$ & $(8.1 \pm 3.1) \cdot 10^{-4}$ & $0.0031 \pm 0.0005$ & $0.0027 \pm 0.0004$ & $0.22$ \\
Wine & total phenols & $0.034 \pm 0.001$ & $0.032 \pm 0.001$ & $0.039 \pm 0.002$ & $0.038 \pm 0.002$ & $0.42$ \\
Quality & flavanoids & $0.135 \pm 0.003$ & $0.125 \pm 0.003$ & $0.161 \pm 0.004$ & $0.162 \pm 0.003$ & $0.98$ \\
& nonflavanoids & $0.050 \pm 0.002$ & $0.049 \pm 0.002$ & $0.035 \pm 0.003$ & $0.039 \pm 0.003$ & $0.76$ \\
& PACs & $0.013 \pm 0.0008$ & $0.014 \pm 0.0007$ & $0.015 \pm 0.001$ & $0.015 \pm 0.001$ & $0.43$ \\
& color intensity & $0.031 \pm 0.002$ & $0.032 \pm 0.002$ & $0.029 \pm 0.002$ & $0.032 \pm 0.002$ & $0.56$ \\
& hue & $0.066 \pm 0.002$ & $0.068 \pm 0.002$ & $0.085 \pm 0.003$ & $0.085 \pm 0.002$ & $0.56$ \\
& od280/od315 & $0.170 \pm 0.004$ & $0.165 \pm 0.004$ & $0.200 \pm 0.005$ & $0.197 \pm 0.005$ & $1.24$ \\
& proline & $0.102 \pm 0.003$ & $0.098 \pm 0.003$ & $0.190 \pm 0.005$ & $0.184 \pm 0.004$ & $1.07$ \\
    \hline
   & buying price & $0.108 \pm 0.004$ & $0.103 \pm 0.004$ & $0.124 \pm 0.004$ & $0.124 \pm 0.004$ & $6.43$ \\
& maintenance & $0.094 \pm 0.003$ & $0.085 \pm 0.003$ & $0.100 \pm 0.004$ & $0.101 \pm 0.004$ & $5.63$ \\
Car & doors & $0.0023 \pm 0.0005$ & $0.0016 \pm 0.0005$ & $0.0042 \pm 0.0008$ & $0.0034 \pm 0.0009$ & $-0.67$ \\
Evaluation & persons & $0.183 \pm 0.005$ & $0.187 \pm 0.005$ & $0.189 \pm 0.005$ & $0.175 \pm 0.005$ & $2.02$ \\
& lug boot & $0.028 \pm 0.002$ & $0.028 \pm 0.002$ & $0.041 \pm 0.002$ & $0.039 \pm 0.002$ & $-2.14$ \\
& safety & $0.216 \pm 0.005$ & $0.218 \pm 0.005$ & $0.225 \pm 0.005$ & $0.226 \pm 0.005$ & $-3.47$ \\
\hline
    \end{tabular}
\end{table}
\begin{table}[H]
    \caption{\ac{fi} results for the global methods between \ac{nn} and \ac{dt} on classification data.}
    \label{tab:FI_global_cla2}
    \centering
    \begin{tabular}{|c|c|c|c|c|c|c|}
    \hline
    \textbf{Data Set} & \textbf{Feature} & \textbf{\ac{se} \Ac{dt}} & \textbf{\ac{se} \Ac{nn}} & \textbf{\ac{sage} \Ac{dt}} & \textbf{\ac{sage} \Ac{nn}} & \textbf{\ac{rentt}-} \\
    & & & & & & \textbf{\ac{fi}} \\
   \hline
    & Elevation & $(9.8 \pm 6.6)  \cdot 10^{-4}$ & $-0.0011 \pm 0.0004$ & $-0.52 \pm 0.02$ & $0.001 \pm 0.001$ & $183.7$ \\
    & Aspect & $(6.3 \pm 6.8)  \cdot 10^{-4}$ & $0.002 \pm 0.001$ & $-0.44 \pm 0.02$ & $0.004 \pm 0.001$ & $-10.2$ \\
    & Slope & $-0.022 \pm 0.003$ & $0.017 \pm 0.003$ & $-0.37 \pm 0.02$ & $0.022 \pm 0.003$ & $8.1$ \\
    & Hor. Dist. Hydrology & $0.006 \pm 0.001$ & $-0.001 \pm 0.001$ & $-0.51 \pm 0.02$ & $0.004 \pm 0.002$ & $-19.3$ \\
    & Vert. Dist. Hydrology & $-0.005 \pm 0.003$ & $0.019 \pm 0.004$ & $-0.24 \pm 0.02$ & $0.030 \pm 0.003$ & $25.2$ \\
    & Hor. Dist. Roadways & $0.011 \pm 0.003$ & $0.0037 \pm 0.001$ & $-0.80 \pm 0.03$ & $0.002 \pm 0.001$ & $-51.3$ \\
    & Hillshade 9am & $(-9.1 \pm 2.4)  \cdot 10^{-4}$ & $0.006 \pm 0.003$ & $0.015 \pm 0.003$ & $0.0027 \pm 0.0007$ & $58.1$ \\
    & Hillshade Noon & $(10.0 \pm 0.6) \cdot 10^{-4}$ & $(-2.4 \pm 0.54)  \cdot 10^{-4}$ & $-0.39 \pm 0.02$ & $(-2.0 \pm 0.5)  \cdot 10^{-4}$ & $-37.3$ \\
    & Hillshade 3pm & $0.066 \pm 0.004$ & $0.026 \pm 0.004$ & $-0.80 \pm 0.03$ & $0.025 \pm 0.004$ & $20.1$ \\
    & Hor. Dist. Fire Points & $0.004 \pm 0.003$ & $0.010 \pm 0.002$ & $-0.73 \pm 0.03$ & $0.011 \pm 0.002$ & $-36.9$ \\
    & Wilderness Area 1 & $0.012 \pm 0.001$ & $0.017 \pm 0.003$ & $-0.56 \pm 0.02$ & $0.025 \pm 0.003$ & $-38.9$ \\
    & Wilderness Area 2 & $0.042 \pm 0.001$ & $0.060 \pm 0.004$ & $-0.43 \pm 0.02$ & $0.068 \pm 0.004$ & $-6.0$ \\
    & Wilderness Area 3 & $0.0037 \pm 0.0005$ & $0.0081 \pm 0.002$ & $-0.066 \pm 0.009$ & $0.012 \pm 0.002$ & $-29.1$ \\
    & Wilderness Area 4 & $-0.0053 \pm 0.0006$ & $0.022 \pm 0.004$ & $0.006 \pm 0.008$ & $0.029 \pm 0.004$ & $-3.4$ \\
   Forest & Soil Type 1 & $-0.013 \pm 0.002$ & $-0.011 \pm 0.004$ & $-0.25 \pm 0.02$ & $-0.016 \pm 0.004$ & $-1.3$ \\
   Cover & Soil Type 2 & $-0.005 \pm 0.002$ & $0.011 \pm 0.004$ & $-0.19 \pm 0.02$ & $0.020 \pm 0.003$ & $-1.9$ \\
    Type & Soil Type 3 & $0.121 \pm 0.003$ & $0.047 \pm 0.004$ & $-0.92 \pm 0.03$ & $0.053 \pm 0.005$ & $-0.8$ \\
    & Soil Type 4 & $-0.005 \pm 0.002$ & $0.007 \pm 0.002$ & $-0.39 \pm 0.02$ & $0.010 \pm 0.002$ & $-1.6$ \\
    & Soil Type 5 & $(1.0 \pm 6.1)  \cdot 10^{-4}$ & $-0.0057 \pm 0.001$ & $-0.34 \pm 0.02$ & $-0.011 \pm 0.002$ & $-1.6$ \\
    & Soil Type 6 & $-0.0011 \pm 0.0004$ & $0.0085 \pm 0.002$ & $-0.06 \pm 0.01$ & $0.014 \pm 0.002$ & $-1.7$ \\
    & Soil Type 7 & $(-8.9 \pm 6.9)  \cdot 10^{-4}$ & $0.0035 \pm 0.002$ & $-0.41 \pm 0.02$ & $0.005 \pm 0.002$ & $0.0$ \\
    & Soil Type 8 & $0.0 \pm 0.0$ & $(-3.8 \pm 2.0)  \cdot 10^{-4}$ & $0.0 \pm 0.0$ & $(-3.0 \pm 1.0)  \cdot 10^{-4}$ & $-0.19$ \\
    & Soil Type 9 & $(2.8 \pm 5.9)  \cdot 10^{-4}$ & $-0.007 \pm 0.002$ & $-0.21 \pm 0.02$ & $-0.004 \pm 0.001$ & $-0.55$ \\
    & Soil Type 10 & $0.011 \pm 0.0009$ & $0.0042 \pm 0.001$ & $-0.14 \pm 0.01$ & $0.009 \pm 0.002$ & $-9.9$ \\
    & Soil Type 11 & $-0.0031 \pm 0.0006$ & $0.0017 \pm 0.0005$ & $-0.23 \pm 0.01$ & $0.0020 \pm 0.0004$ & $-2.4$ \\
    & Soil Type 12 & $0.086 \pm 0.002$ & $0.052 \pm 0.004$ & $-0.44 \pm 0.02$ & $0.045 \pm 0.005$ & $-10.2$ \\
    & Soil Type 13 & $0.007 \pm 0.001$ & $0.008 \pm 0.002$ & $-0.066 \pm 0.01$ & $0.010 \pm 0.002$ & $-7.9$ \\
    & Soil Type 14 & $0.0056 \pm 0.0006$ & $(-1.5 \pm 10)  \cdot 10^{-4}$ & $-0.49 \pm 0.02$ & $(-3.0 \pm 10)  \cdot 10^{-4}$ & $-0.022$ \\
\hline
    \end{tabular}
\end{table}

\begin{table}[H]
    \caption{\ac{fi} results for the global methods between \ac{nn} and \ac{dt} on classification data.}
    \label{tab:FI_global_cla3}
    \centering
    \begin{tabular}{|c|c|c|c|c|c|c|}
    \hline
    \textbf{Data Set} & \textbf{Feature} & \textbf{\ac{se} \Ac{dt}} & \textbf{\ac{se} \Ac{nn}} & \textbf{\ac{sage} \Ac{dt}} & \textbf{\ac{sage} \Ac{nn}} & \textbf{\ac{rentt}-\ac{fi}} \\
   \hline
    & Soil Type 15 & $-0.0015 \pm 0.0004$ & $0.001 \pm 0.001$ & $-0.05 \pm 0.01$ & $0.004 \pm 0.002$ & $0.0$ \\
       & Soil Type 16 & $(2.6 \pm 0.2)  \cdot 10^{-4}$ & $(-5.1 \pm 0.71)  \cdot 10^{-4}$ & $-0.37 \pm 0.02$ & $(-6.0 \pm 0.70)  \cdot 10^{-4}$ & $-0.51$ \\
    & Soil Type 17 & $(1.20 \pm 0.05)  \cdot 10^{-4}$ & $(-5.1 \pm 5.0)  \cdot 10^{-5}$ & $-0.16 \pm 0.01$ & $(0.6 \pm 6.0)  \cdot 10^{-5}$ & $-0.41$ \\
    & Soil Type 18 & $(0.8 \pm 4.0)  \cdot 10^{-4}$ & $(6.6 \pm 9.0)  \cdot 10^{-4}$ & $0.0034 \pm 0.001$ & $(-7.0 \pm 6.0)  \cdot 10^{-4}$ & $0.47$ \\
    & Soil Type 19 & $0.007 \pm 0.001$ & $-0.0072 \pm 0.003$ & $-0.38 \pm 0.02$ & $-0.007 \pm 0.003$ & $-0.80$ \\
    & Soil Type 20 & $(5.7 \pm 0.24)  \cdot 10^{-4}$ & $(-7.3 \pm 1.6)  \cdot 10^{-4}$ & $-0.096 \pm 0.009$ & $(-6.0 \pm 0.80)  \cdot 10^{-4}$ & $-4.6$ \\
    & Soil Type 21 & $(0.59 \pm 6.2)  \cdot 10^{-4}$ & $0.010 \pm 0.003$ & $-0.27 \pm 0.02$ & $0.006 \pm 0.002$ & $-0.27$ \\
    & Soil Type 22 & $0.0 \pm 0.0$ & $0.0 \pm 0.0$ & $0.0 \pm 0.0$ & $0.0 \pm 0.0$ & $-13.9$ \\
    & Soil Type 23 & $0.0025 \pm 0.0006$ & $0.008 \pm 0.001$ & $-0.074 \pm 0.01$ & $0.010 \pm 0.001$ & $-19.9$ \\
    & Soil Type 24 & $-0.016 \pm 0.003$ & $0.040 \pm 0.004$ & $-0.11 \pm 0.03$ & $0.046 \pm 0.004$ & $-8.5$ \\
    & Soil Type 25 & $0.0033 \pm 0.001$ & $0.014 \pm 0.002$ & $-0.089 \pm 0.01$ & $0.014 \pm 0.002$ & $-0.38$ \\
    & Soil Type 26 & $0.0 \pm 0.0$ & $0.0 \pm 0.0$ & $0.0 \pm 0.0$ & $0.0 \pm 0.0$ & $15.9$ \\
  Forest  & Soil Type 27 & $-0.0011 \pm 0.0003$ & $-0.0026 \pm 0.003$ & $0.16 \pm 0.01$ & $-0.001 \pm 0.003$ & $-0.26$ \\
 Cover   & Soil Type 28 & $-0.001 \pm 0.001$ & $0.028 \pm 0.003$ & $-0.32 \pm 0.02$ & $0.023 \pm 0.003$ & $-0.73$ \\
  Type  & Soil Type 29 & $0.0076 \pm 0.0008$ & $0.001 \pm 0.001$ & $-0.63 \pm 0.02$ & $0.004 \pm 0.002$ & $-20.9$ \\
    & Soil Type 30 & $0.0 \pm 0.0$ & $(1.7 \pm 3.0)  \cdot 10^{-4}$ & $0.0 \pm 0.0$ & $(-9.0 \pm 4.3)  \cdot 10^{-5}$ & $-6.76$ \\
    & Soil Type 31 & $0.043 \pm 0.002$ & $0.338 \pm 0.008$ & $-0.83 \pm 0.03$ & $0.316 \pm 0.008$ & $-7.9$ \\
    & Soil Type 32 & $0.0022 \pm 0.0006$ & $0.019 \pm 0.002$ & $-0.16 \pm 0.01$ & $0.029 \pm 0.003$ & $-23.9$ \\
    & Soil Type 33 & $0.0019 \pm 0.0002$ & $0.0099 \pm 0.001$ & $-0.032 \pm 0.006$ & $0.0098 \pm 0.002$ & $-14.0$ \\
    & Soil Type 34 & $0.043 \pm 0.003$ & $0.019 \pm 0.003$ & $-0.61 \pm 0.02$ & $0.022 \pm 0.003$ & $-0.45$ \\
    & Soil Type 35 & $(3.0 \pm 1.2)  \cdot 10^{-4}$ & $-0.004 \pm 0.002$ & $-0.026 \pm 0.004$ & $-0.002 \pm 0.001$ & $0.088$ \\
    & Soil Type 36 & $0.0 \pm 0.0$ & $(5.0 \pm 3.0)  \cdot 10^{-5}$ & $0.0 \pm 0.0$ & $(-9.0 \pm 3.0)  \cdot 10^{-5}$ & $0.0$ \\
    & Soil Type 37 & $(-6.7 \pm 5.5)  \cdot 10^{-4}$ & $-0.0038 \pm 0.001$ & $-0.27 \pm 0.02$ & $-0.004 \pm 0.002$ & $-0.011$ \\
    & Soil Type 38 & $-0.011 \pm 0.003$ & $0.012 \pm 0.004$ & $-0.094 \pm 0.02$ & $0.012 \pm 0.003$ & $-3.1$ \\
    & Soil Type 39 & $0.029 \pm 0.004$ & $0.026 \pm 0.004$ & $-0.57 \pm 0.03$ & $0.022 \pm 0.004$ & $1.08$ \\
    & Soil Type 40 & $0.166 \pm 0.004$ & $0.064 \pm 0.005$ & $-0.83 \pm 0.02$ & $0.058 \pm 0.005$ & $-0.67$ \\
    \hline
    \end{tabular}
\end{table}
\end{landscape}

\section{Feature Importance Results between Neural Networks and Decision Tree for Local and Global Methods}
\label{a-ssec:nn-vs-dt}

\begin{figure}[htb]
    \centering
    \includegraphics[width=0.5\linewidth]{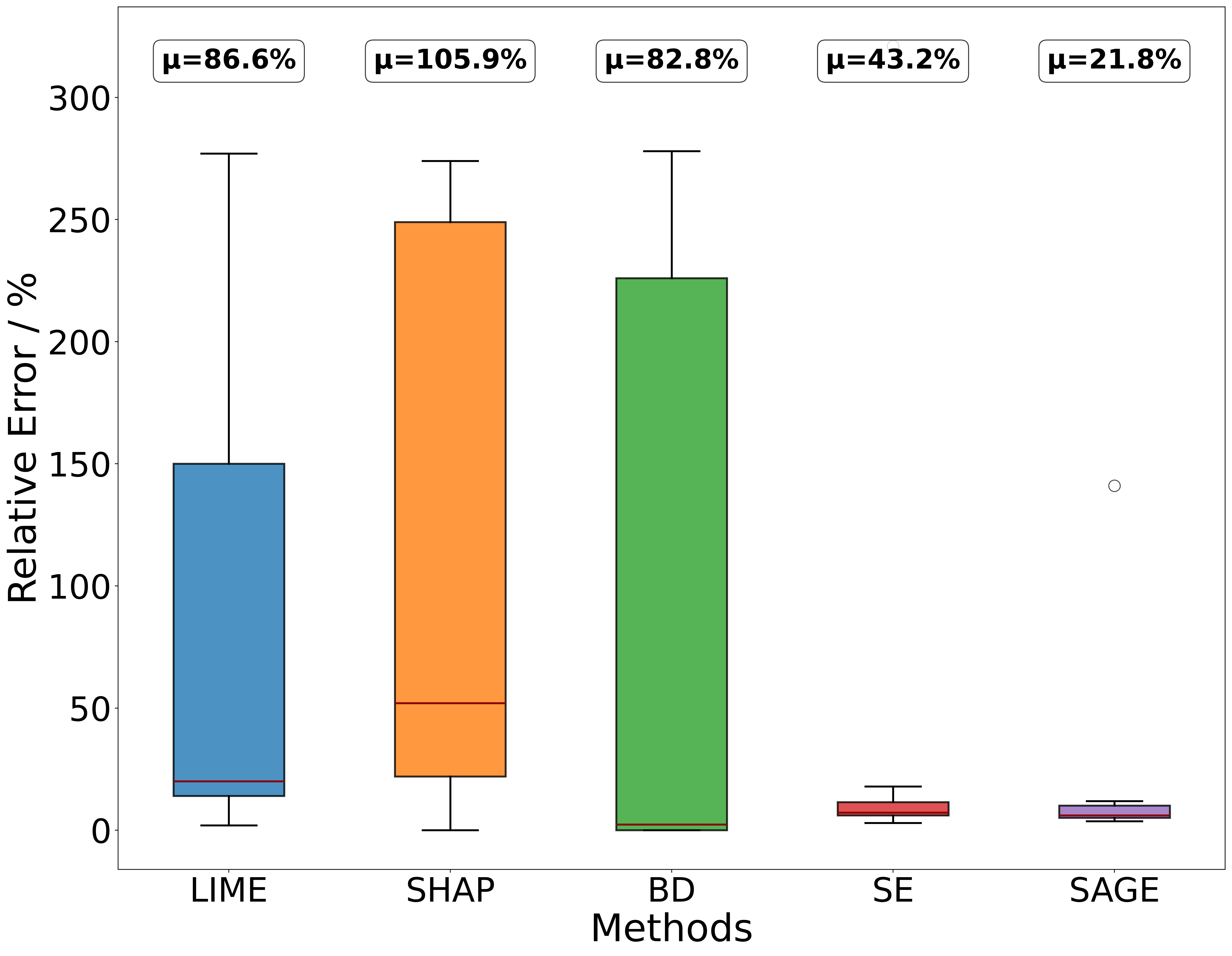}
    \caption{Comparison of \ac{re} for \ac{fi} methods when applied to functionally equivalent \ac{nn} and \ac{dt} across multiple data sets.}
    \label{fig:FI_nndt_average}
\end{figure}

\begin{figure}[htb]
    \centering
    \subfigure[Regression vs classification tasks, mean across all global and local methods.]{%
        \label{fig:FI_nndt_boxplot_reg_vs_class_sub}%
        \includegraphics[width=0.45\textwidth]{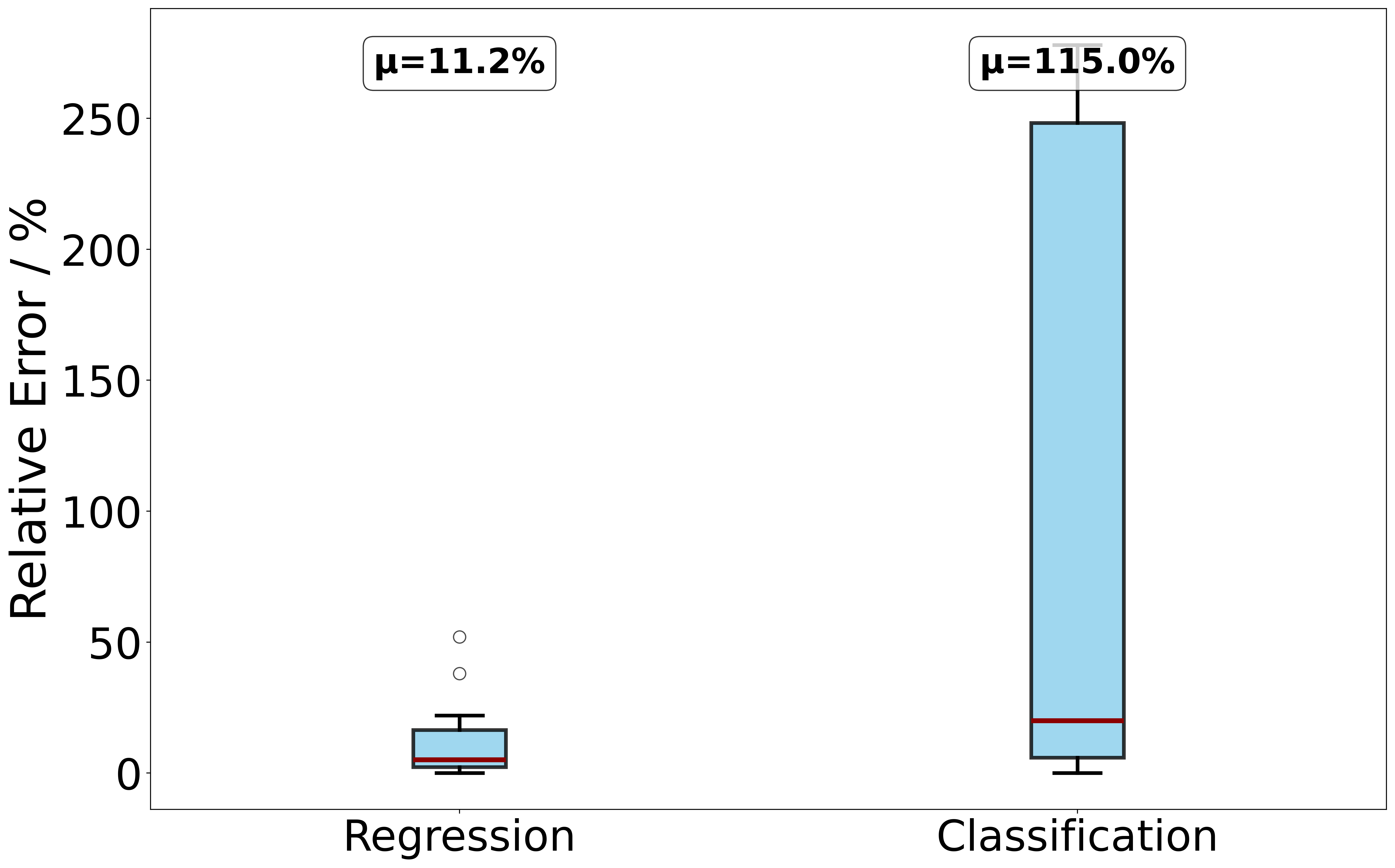}%
    }\hfill
    \subfigure[Local vs global \ac{fi} methods, mean across all tasks.]{%
        \label{fig:FI_nndt_boxplot_local_vs_global_sub}%
        \includegraphics[width=0.45\textwidth]{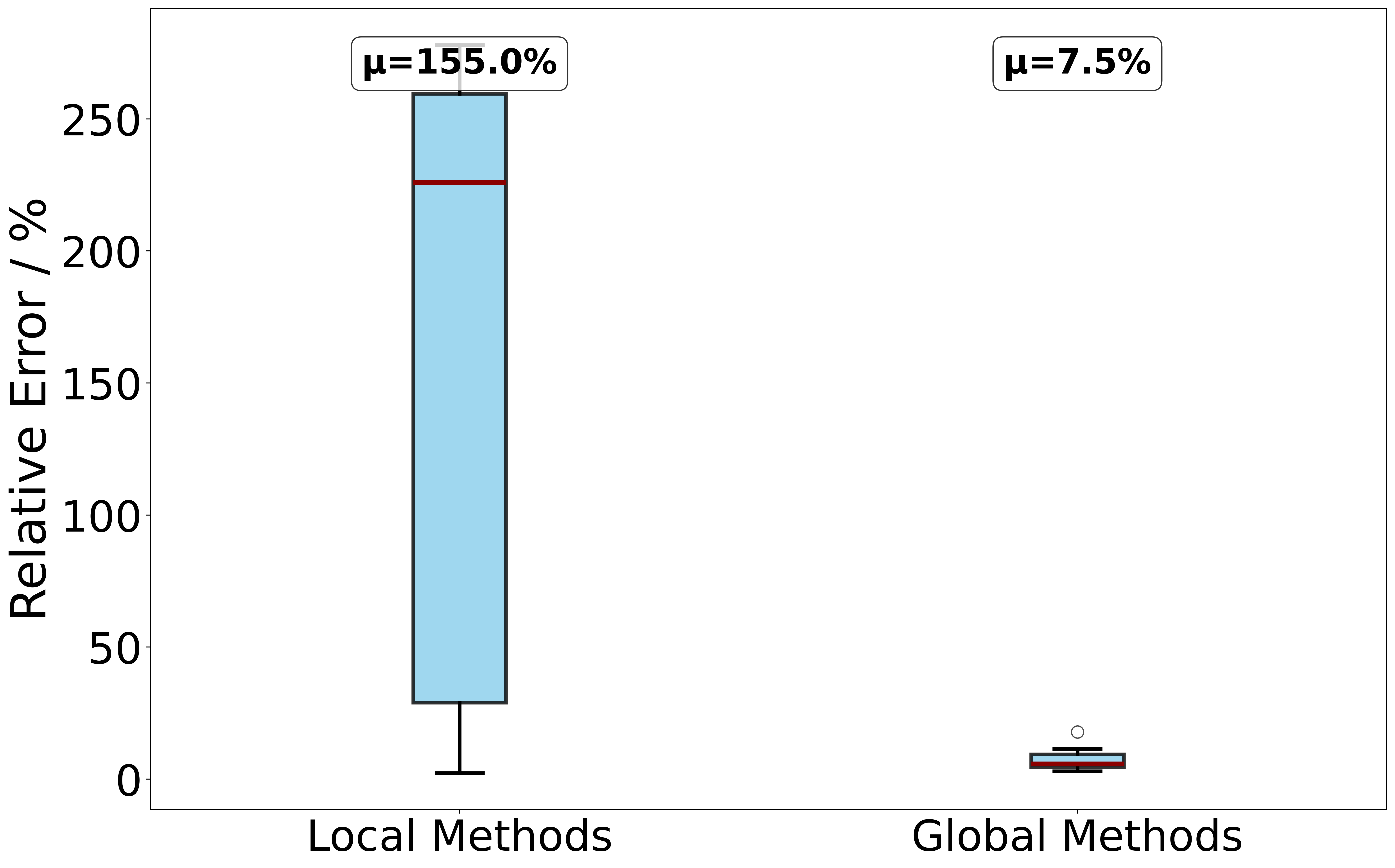}%
    }
    \caption{Boxplot summary comparing mean \ac{re} for \ac{fi} methods when applied to functionally equivalent \ac{nn} and \ac{dt} models across all methods and data sets.}
    \label{fig:FI_nndt_boxplot}
\end{figure}

This section provides detailed quantitative comparisons of feature importance values between neural networks and their equivalent decision trees. The analysis uses Root Mean Squared Error (RMSE) and Relative Error (RE) to quantify the differences between the methods.
Figures~\ref{fig:FI_nndt_average} and~\ref{fig:FI_nndt_boxplot} underpin the results of Section~\ref{sec:fi-nn-vs-dt}: The local \ac{fi} methods show significantly higher \ac{re} values than the global methods, while these still have non-negligible \acp{re} (Figures~\ref{fig:FI_nndt_average} and~\ref{fig:FI_nndt_boxplot_local_vs_global_sub}).
Additionally, the \ac{re} is in general lower for the regression tasks than for the classification tasks (Figure~\ref{fig:FI_nndt_boxplot_reg_vs_class_sub}).
The results in Tables~\ref{tab:nn_vs_dt_local_relative} and~\ref{tab:nn_vs_dt_global_relative} further support the findings in Section~\ref{sec:fi-nn-vs-dt} by demonstrating that LIME, SHAP, BD, SE, and SAGE produce similar but not identical feature importance values when applied to neural networks versus decision trees.

\begin{landscape}
\begin{table}[H]
    \caption{ \ac{rmse}  and  \ac{re} between \ac{nn} and \ac{dt} for local methods \ac{nn} vs \ac{dt}} \label{tab:nn_vs_dt_local_relative}
    \centering
    \begin{tabular}{|c|c|c|c|c|c|c|}
        \hline
        \multirow{2}{*}{\textbf{Data Set}} & \multicolumn{2}{c|}{\textbf{ \ac{lime} \ac{nn} vs \ac{dt}}} & \multicolumn{2}{c|}{\textbf{ \ac{shap} \ac{nn} vs \ac{dt}}} & \multicolumn{2}{c|}{\textbf{BD \ac{nn} vs \ac{dt}}} \\
        \cline{2-7}
         & \textbf{ \ac{rmse} } & \textbf{RE / \%} & \textbf{ \ac{rmse} } & \textbf{RE / \%} & \textbf{ \ac{rmse} } & \textbf{RE / \%} \\
        \hline
        Absolute & \(0.008 \pm 0.006\) & \(2 \pm 2\) & \((3 \pm 2) \cdot 10^{-7}\) & \((1.4 \pm 0.8) \cdot 10^{-4}\) & \((3 \pm 2) \cdot 10^{-7}\) & \((1.4 \pm 0.8) \cdot 10^{-4}\) \\
        
        Linear & \(0.02 \pm 0.06\) & \(20 \pm 60\) & \(0.002 \pm 0.0\) & \(2.3 \pm 0.0\) & \(0.002 \pm 0.0\) & \(2.3 \pm 0.0\) \\
        
        Diabetes Reg & \(0.005 \pm 0.001\) & \(12 \pm 3\) & \(0.006 \pm 0.004\) & \(22 \pm 15\) & \((3 \pm 1) \cdot 10^{-8}\) & \((10 \pm 4) \cdot 10^{-5}\) \\
        
        
        California Housing & \(0.05 \pm 0.04\) & \(38 \pm 34\) & \(0.04 \pm 0.16\) & \(52 \pm 184\) & \(0.005 \pm 0.002\) & \(4 \pm 2\) \\
        \hline
        
        Iris & \(0.06 \pm 0.02\) & \(248 \pm 83\) & \(0.07 \pm 0.03\) & \(270 \pm 100\) & \(0.09 \pm 0.03\) & \(226 \pm 76\) \\
        
        Diabetes Class & \(0.010 \pm 0.003\) & \(14 \pm 4\) & \(0.01 \pm 0.02\) & \(26 \pm 29\) & \((3 \pm 1) \cdot 10^{-7}\) & \((5 \pm 2) \cdot 10^{-4}\) \\
        
        Wine Quality & \(0.03 \pm 0.01\) & \(150 \pm 60\) & \(0.06 \pm 0.02\) & \(249 \pm 76\) & \(0.08 \pm 0.02\) & \(225 \pm 55\) \\
        
        Car Evaluation & \(0.20 \pm 0.04\) & \(277 \pm 59\) & \(0.16 \pm 0.04\) & \(274 \pm 66\) & \(0.20 \pm 0.07\) & \(278 \pm 90\) \\
        
        Forest Cover Type & - & - & - & - & - & - \\
        \hline
    \end{tabular}
\end{table}

\begin{table}[H]
       \centering
    \caption{ \ac{rmse}  and  \ac{re} between \ac{dt} and \ac{nn}  \ac{fi} values for \ac{se} and \ac{sage} methods.} \label{tab:nn_vs_dt_global_relative}
    \begin{tabular}{|c|c|c|c|c|}
        \hline
         \multirow{2}{*}{\textbf{Data Set}} & \multicolumn{2}{c|}{\textbf{SE \ac{nn} vs \ac{dt}}} & \multicolumn{2}{c|}{\textbf{ \ac{sage} \ac{nn} vs \ac{dt}}}\\
         \cline{2-5}
         & \textbf{ \ac{rmse} } & \textbf{RE / \%} & \textbf{ \ac{rmse} } & \textbf{RE / \%} \\
        
        \hline
        Linear & $(6.0 \pm 1.0) \cdot 10^{-4}$ & $3.0 \pm 2.7$ & $0.0012 \pm 0.001$ & $6.1 \pm 2.7$ \\
        Diabetes Reg & $(2.0 \pm 0.0) \cdot 10^{-4}$ & $6.6 \pm 1.9$ & $(1.0 \pm 0.0) \cdot 10^{-4}$ & $3.7 \pm 2.2$ \\
        California Housing & $0.0012 \pm 0.000$ & $17.9 \pm 3.4$ & $(7.0 \pm 0.0) \cdot 10^{-4}$ & $10.1 \pm 3.3$ \\
        \hline
        Iris & $0.0075 \pm 0.002$ & $11.5 \pm 2.5$ & $0.0078 \pm 0.004$ & $5.3 \pm 2.9$ \\
        Diabetes Class & $0.0016 \pm 0.001$ & $6.1 \pm 2.7$ & $0.0023 \pm 0.001$ & $8.6 \pm 2.8$ \\
        Wine Quality & $0.0037 \pm 0.001$ & $7.2 \pm 1.5$ & $0.0028 \pm 0.001$ & $4.2 \pm 1.6$ \\
        Car Evaluation & $0.0046 \pm 0.002$ & $4.4 \pm 2.0$ & $0.0058 \pm 0.002$ & $5.1 \pm 1.9$ \\
        Forest Cover Type & $0.0463 \pm 0.001$ & $321 \pm 10$ & $0.4221 \pm 0.002$ & $141 \pm 1$ \\
        \hline
    \end{tabular}
\end{table}
\end{landscape}

\section{Additional Results for the Comparison of the Feature Importance Methods} \label{a-sec:additional-FI-results}
This section provides comprehensive comparisons between different feature importance methods. The analysis includes both local methods (LIME, SHAP, BD, RENTT-FI) and global methods (SE, SAGE, RENTT-FI) evaluated using RMSE, relative error metrics and Krippendorff's alpha in ordinal and interval manner.

The results show that traditional methods (LIME, SHAP, BD) generally produce more similar feature importance values to each other, but differ significantly from the ground truth revealing method RENTT-FI.
\begin{table}[hbt]
    \caption{Pairwise \ac{rmse} and \ac{re} for the local \ac{fi} methods for the regression tasks.}
    \centering
    \begin{tabular}{|c|l|c|c|}
        \hline
        \textbf{Data Set} & \textbf{Comparison} & \textbf{ \ac{rmse} } & \textbf{RE / \%} \\
        \hline
         & \ac{lime} vs \ac{shap} & \(0.13 \pm 0.06\) & \(40 \pm 18\) \\
         & \ac{lime} vs \ac{bd} & \(0.13 \pm 0.06\) & \(40 \pm 18\) \\
        Absolute & \ac{lime} vs  \ac{rentt}-\ac{fi} & \(0.51 \pm 0.01\) & \(153 \pm 3\) \\
         & \ac{shap} vs \ac{bd} & \((1.2 \pm 2.0) \cdot 10^{-9}\) & \((5.0 \pm 8.0) \cdot 10^{-6}\) \\
         & \ac{shap} vs  \ac{rentt}-\ac{fi} & \(0.49 \pm 0.01\) & \(199 \pm 4\) \\
         & \ac{bd} vs  \ac{rentt}-\ac{fi} & \(0.49 \pm 0.01\) & \(199 \pm 4\) \\
        \hline
         & \ac{lime} vs \ac{shap} & \(0.06 \pm 0.03\) & \(55 \pm 23\) \\
         & \ac{lime} vs \ac{bd} & \(0.06 \pm 0.03\) & \(55 \pm 23\) \\
        Linear & \ac{lime} vs  \ac{rentt}-\ac{fi} & \(0.33 \pm 0.05\) & \(293 \pm 48\) \\
         & \ac{shap} vs \ac{bd} & \((3.7 \pm 82) \cdot 10^{-6}\) & \(0.04 \pm 1.0\) \\
         & \ac{shap} vs  \ac{rentt}-\ac{fi} & \(0.326 \pm 0.005\) & \(383 \pm 6\) \\
         & \ac{bd} vs  \ac{rentt}-\ac{fi} & \(0.326 \pm 0.005\) & \(383 \pm 6\) \\
        \hline
         & \ac{lime} vs \ac{shap} & \(0.030 \pm 0.010\) & \(73 \pm 24\) \\
         & \ac{lime} vs \ac{bd} & \(0.032 \pm 0.009\) & \(78 \pm 23\) \\
        Diabetes & \ac{lime} vs  \ac{rentt}-\ac{fi} & \(0.14 \pm 0.02\) & \(332 \pm 51\) \\
         Reg & \ac{shap} vs \ac{bd} & \(0.011 \pm 0.006\) & \(37 \pm 21\) \\
         & \ac{shap} vs  \ac{rentt}-\ac{fi} & \(0.135 \pm 0.013\) & \(478 \pm 45\) \\
         & \ac{bd} vs  \ac{rentt}-\ac{fi} & \(0.136 \pm 0.013\) & \(474 \pm 45\) \\
        \hline
         & \ac{lime} vs \ac{shap} & \(0.26 \pm 0.02\) & \(221 \pm 18\) \\
         & \ac{lime} vs \ac{bd} & \(0.12 \pm 0.08\) & \(100 \pm 63\) \\
        California & \ac{lime} vs  \ac{rentt}-\ac{fi} & \(0.46 \pm 0.02\) & \(384 \pm 19\) \\
        Housing & \ac{shap} vs \ac{bd} & \(0.280 \pm 0.019\) & \(328 \pm 22\) \\
         & \ac{shap} vs  \ac{rentt}-\ac{fi} & \(0.47 \pm 0.02\) & \(549 \pm 27\) \\
         & \ac{bd} vs  \ac{rentt}-\ac{fi} & \(0.47 \pm 0.02\) & \(384 \pm 18\) \\
        \hline
    \end{tabular}
    \label{tab:rmse_relative_local_reg}
\end{table}

\begin{table}[H]
    \caption{Pairwise \ac{rmse} and \ac{re} for the local \ac{fi} methods for the classification tasks.}
    \centering
    \begin{tabular}{|c|l|c|c|}
        \hline
        \textbf{Data Set} & \textbf{Comparison} & \textbf{ \ac{rmse} } & \textbf{RE / \%} \\
        \hline
         & \ac{lime} vs \ac{shap} & \(0.038 \pm 0.010\) & \(160 \pm 42\) \\
         & \ac{lime} vs \ac{bd} & \(0.044 \pm 0.014\) & \(186 \pm 58\) \\
        Iris & \ac{lime} vs  \ac{rentt}-\ac{fi} & \(0.70 \pm 0.25\) & \(2921 \pm 1058\) \\
         & \ac{shap} vs \ac{bd} & \(0.034 \pm 0.019\) & \(124 \pm 70\) \\
         & \ac{shap} vs  \ac{rentt}-\ac{fi} & \(0.69 \pm 0.25\) & \(2548 \pm 917\) \\
         & \ac{bd} vs  \ac{rentt}-\ac{fi} & \(0.69 \pm 0.26\) & \(1800 \pm 678\) \\
        \hline
         & \ac{lime} vs \ac{shap} & \(0.057 \pm 0.023\) & \(76 \pm 30\) \\
         & \ac{lime} vs \ac{bd} & \(0.059 \pm 0.024\) & \(79 \pm 32\) \\
        Diabetes & \ac{lime} vs  \ac{rentt}-\ac{fi} & \(1.28 \pm 0.30\) & \(1696 \pm 401\) \\
        Class & \ac{shap} vs \ac{bd} & \(0.023 \pm 0.020\) & \(44 \pm 37\) \\
         & \ac{shap} vs  \ac{rentt}-\ac{fi} & \(1.27 \pm 0.30\) & \(2391 \pm 561\) \\
         & \ac{bd} vs  \ac{rentt}-\ac{fi} & \(1.27 \pm 0.30\) & \(2323 \pm 545\) \\
        \hline
         & \ac{lime} vs \ac{shap} & \(0.095 \pm 0.034\) & \(132 \pm 47\) \\
         & \ac{lime} vs \ac{bd} & \(0.143 \pm 0.049\) & \(200 \pm 68\) \\
        Car & \ac{lime} vs  \ac{rentt}-\ac{fi} & \(5.92 \pm 0.36\) & \(8259 \pm 506\) \\
        Evaluation & \ac{shap} vs \ac{bd} & \(0.110 \pm 0.047\) & \(193 \pm 83\) \\
         & \ac{shap} vs  \ac{rentt}-\ac{fi} & \(5.93 \pm 0.36\) & \(10379 \pm 632\) \\
         & \ac{bd} vs  \ac{rentt}-\ac{fi} & \(5.93 \pm 0.36\) & \(8099 \pm 497\) \\
        \hline
         & \ac{lime} vs \ac{shap} & \(0.109 \pm 0.018\) & \(496 \pm 84\) \\
         & \ac{lime} vs \ac{bd} & \(0.105 \pm 0.016\) & \(479 \pm 72\) \\
        Wine & \ac{lime} vs  \ac{rentt}-\ac{fi} & \(0.99 \pm 0.35\) & \(4532 \pm 1587\) \\
        Quality & \ac{shap} vs \ac{bd} & \(0.088 \pm 0.022\) & \(351 \pm 88\) \\
         & \ac{shap} vs  \ac{rentt}-\ac{fi} & \(0.99 \pm 0.36\) & \(3926 \pm 1424\) \\
         & \ac{bd} vs  \ac{rentt}-\ac{fi} & \(0.98 \pm 0.37\) & \(2696 \pm 1002\) \\
        \hline
    \end{tabular}
    \label{tab:rmse_relative_local_cla}
\end{table}

\begin{table}[H]
    \caption{Pairwise \ac{rmse} and \ac{re} for the global \ac{fi} methods for the regression and classification tasks.} \label{tab:rmse_relative_global}
    \centering
    \begin{tabular}{|c|l|c|c|}
        \hline
        \textbf{Data Set} & \textbf{Comparison} & \textbf{RMSE} & \textbf{RE / \%} \\
        \hline
         & SE vs SAGE & $0.0015\pm0.0005$ & $7.7\pm2.7$ \\
         Linear & SE vs RENTT-FI & $0.201\pm0.0004$ & $218\pm0.4$ \\
         & SAGE vs RENTT-FI & $0.203\pm0.0004$ & $221\pm0.4$ \\
        \hline
         Diabetes & SE vs SAGE & $0.0004\pm0.00007$ & $11.0\pm2$ \\
         Reg & SE vs RENTT-FI & $0.0896\pm0.00004$ & $212.1\pm0.1$ \\
        & SAGE vs RENTT-FI & $0.0894\pm0.00005$ & $211.4\pm0.1$ \\
        \hline
         California& SE vs SAGE & $0.0010\pm0.0002$ & $13.0\pm3$ \\
         Housing& SE vs RENTT-FI & $0.293\pm0.0002$ & $262.5\pm0.2$ \\
         & SAGE vs RENTT-FI & $0.293\pm0.0002$ & $262.9\pm0.2$ \\
        \hline
        \hline
         & SE vs SAGE & $0.133\pm0.003$ & $121\pm3$ \\
         Iris& SE vs RENTT-FI & $0.478\pm0.001$ & $190.6\pm0.5$ \\
         & SAGE vs RENTT-FI & $0.391\pm0.003$ & $134\pm1$ \\
        \hline
         Diabetes & SE vs SAGE & $0.0057\pm0.0007$ & $20.9\pm3$ \\
         Class & SE vs RENTT-FI & $0.555\pm0.0005$ & $238.0\pm0.2$ \\
         & SAGE vs RENTT-FI & $0.5500\pm0.0005$ & $234.7\pm0.2$ \\
        \hline
        Wine & SE vs SAGE & $0.0288\pm0.0009$ & $50.0\pm2$ \\
        Quality & SE vs RENTT-FI & $0.669\pm0.0005$ & $188.2\pm0.2$ \\
         & SAGE vs RENTT-FI & $0.652\pm0.0007$ & $179.5\pm0.2$ \\
        \hline
         Car& SE vs SAGE & $0.013\pm0.002$ & $12.2\pm2$ \\
         Evaluation& SE vs RENTT-FI & $3.93\pm0.001$ & $225.0\pm0.08$ \\
         & SAGE vs RENTT-FI & $3.93\pm0.002$ & $224.2\pm0.09$ \\
        \hline
         Forest & SE vs SAGE & $0.0052\pm0.0005$ & $27.9\pm2.8$ \\
         Cover & SE vs RENTT-FI & $30.1\pm0.0004$ & $455.3\pm0.006$ \\
        Type & SAGE vs RENTT-FI & $30.1\pm0.0004$ & $455.3\pm0.006$ \\
        \hline
    \end{tabular}
\end{table}

\begin{figure}[H]
    \centering
    \subfigure[\ac{se} feature contribution]{%
        \includegraphics[height=0.23\textheight]{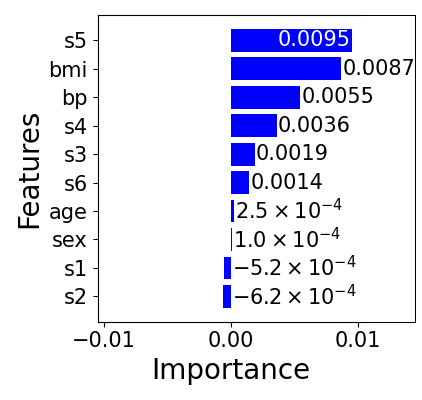}%
    }\hfill
    \subfigure[\ac{sage} feature contribution]{%
        \includegraphics[height=0.23\textheight]{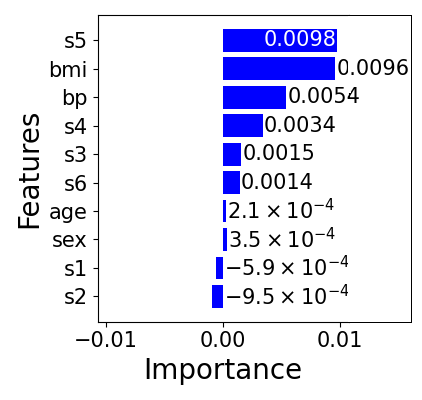}%
    }\\
    \subfigure[\ac{rentt}-\ac{fi} feature contribution]{%
        \includegraphics[height=0.23\textheight]{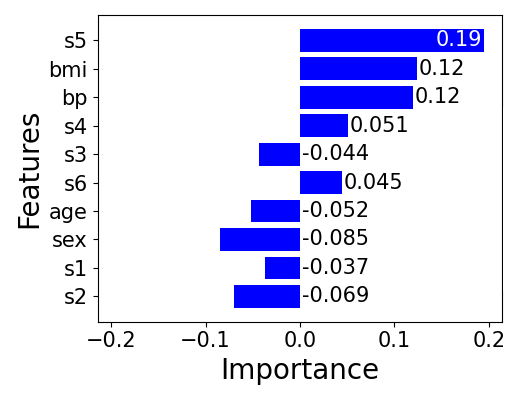}%
    }\hfill
    \subfigure[\ac{rentt}-\ac{fi} feature effect]{%
        \includegraphics[height=0.23\textheight]{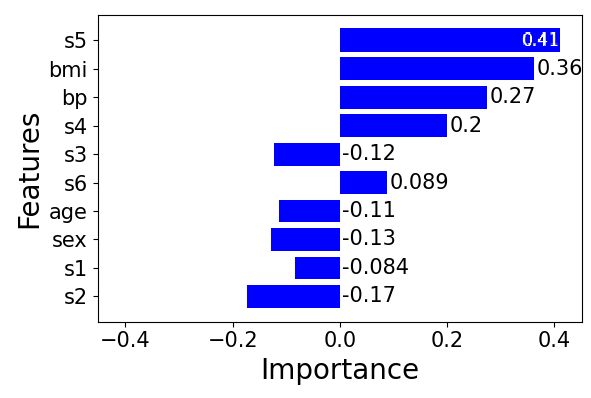}%
    }
    \caption{Global \ac{fi} by \ac{se}, \ac{sage} and \ac{rentt} on an \ac{nn} trained on the Diabetes Reg data set.}
    \label{fig:FI_diabetes_reg_global}
\end{figure}

\begin{figure}[H]
    \centering
    \includegraphics[width=1\linewidth]{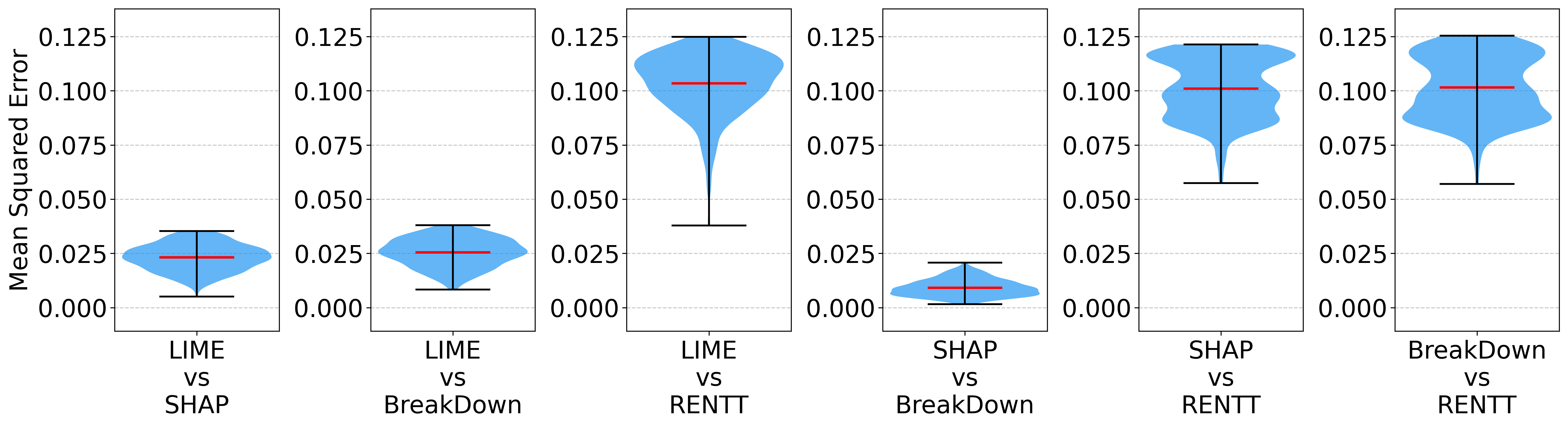}
    \caption{Violin plot of the \ac{rmse} between the different \ac{fi}-methods for the Diabetes Reg data set. Only the 95th percentile of samples is pictured.}
    \label{fig:FI_reg_rmse_violin_diabetes}
\end{figure}

\begin{figure}[H]
    \centering
    \subfigure[Ordinal Krippendorff's $\alpha$ (ranking)]{%
        \includegraphics[width=0.99\textwidth]{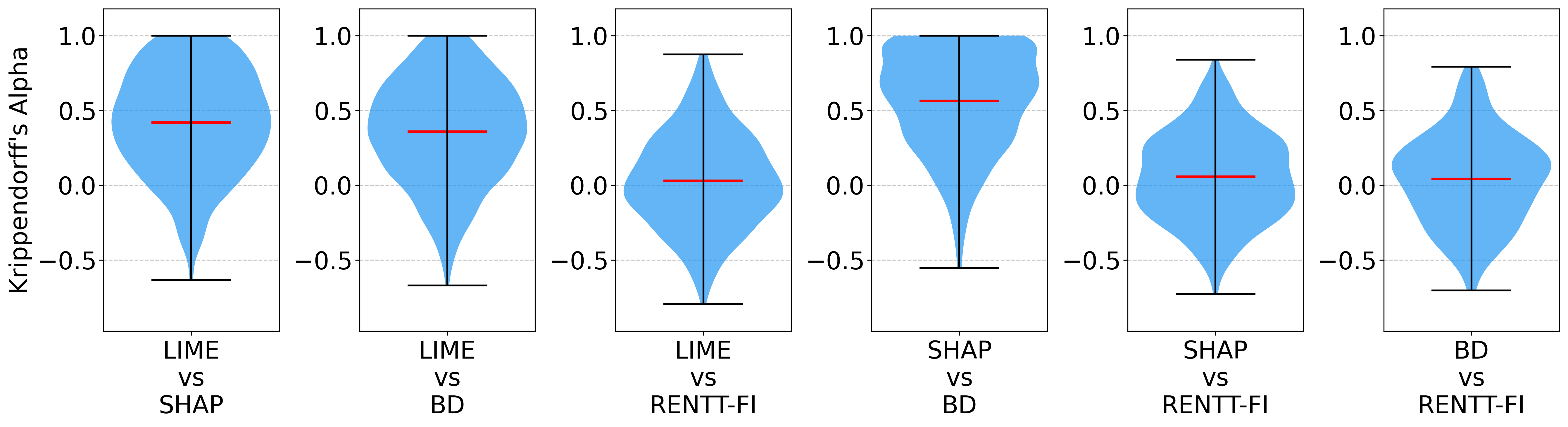}%
    }\\[0.5cm]
    \subfigure[Interval Krippendorff's $\alpha$ (values)]{%
        \includegraphics[width=0.99\textwidth]{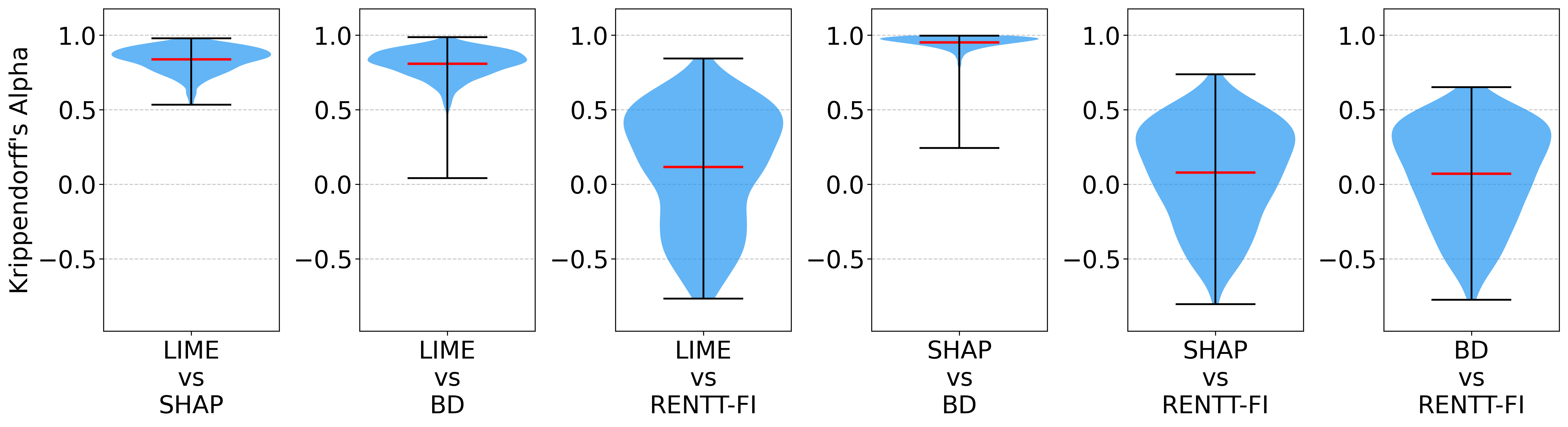}%
    }
    \caption{Violin plots of Krippendorff's $\alpha$ for the Diabetes Reg data set, calculated samplewise between different \ac{fi} methods. Only the 95th percentile of samples is pictured.}
    \label{fig:FI_diabetes_reg_krippendorff}
\end{figure}

\begin{figure}[H]
    \centering
    \subfigure[\ac{se} feature contribution]{%
        \includegraphics[height=0.23\textheight]{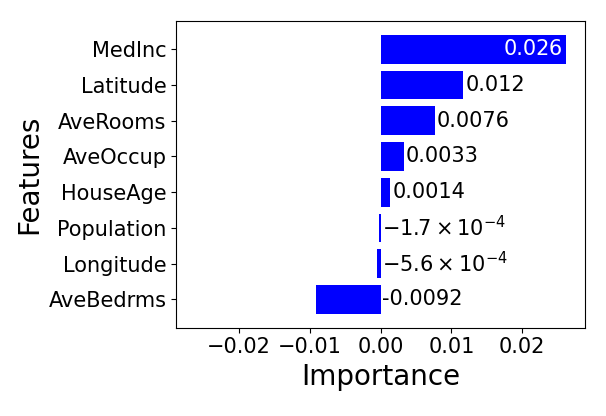}%
    }\hfill
    \subfigure[\ac{sage} feature contribution]{%
        \includegraphics[height=0.23\textheight]{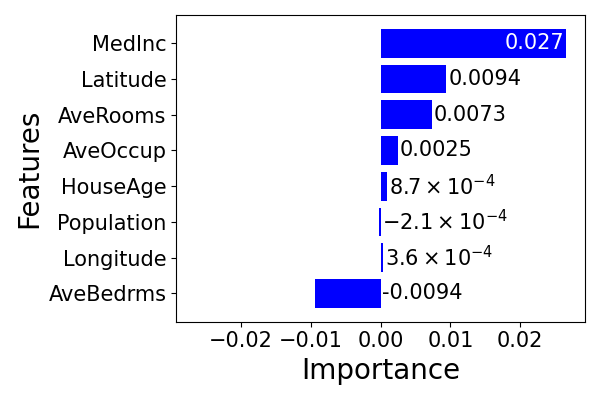}%
    }\\
    \subfigure[\ac{rentt}-\ac{fi} feature contribution]{%
        \includegraphics[height=0.23\textheight]{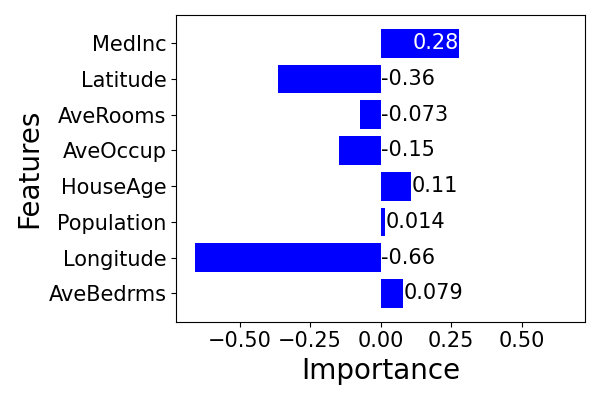}%
    }\hfill
    \subfigure[\ac{rentt}-\ac{fi} feature effect]{%
        \includegraphics[height=0.23\textheight]{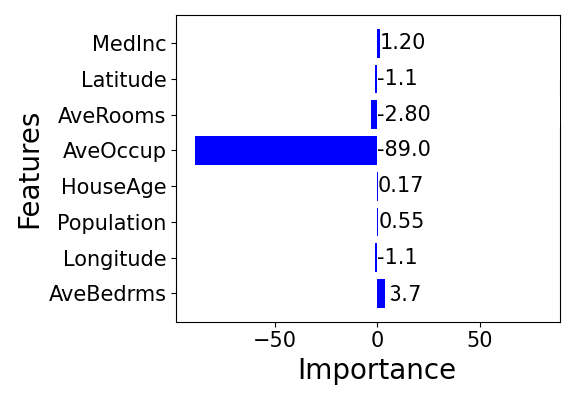}%
    }
    \caption{Global \ac{fi} by \ac{se}, \ac{sage} and \ac{rentt} on an \ac{nn} trained on the California Housing data set.}
    \label{fig:FI_California_global}
\end{figure}

\begin{figure}[H]
    \centering
    \includegraphics[width=1\linewidth]{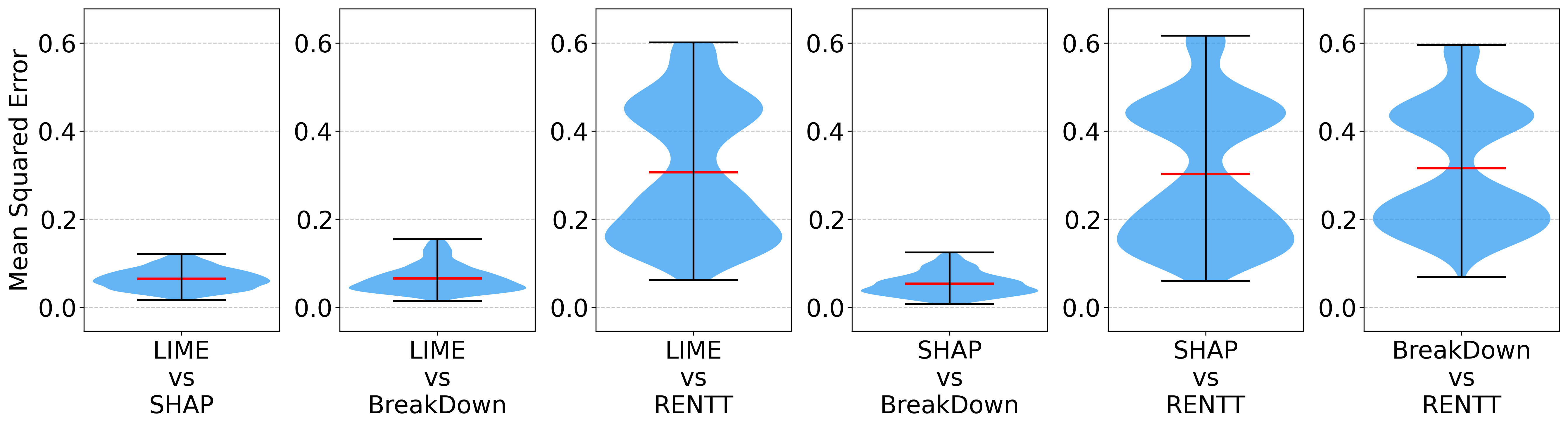}
    \caption{Violin plot of the \ac{rmse} between the different \ac{fi}-methods for the California Housing data set. Only the 95th percentile of samples is pictured.}
    \label{fig:FI_california_rmse_violin}
\end{figure}

\begin{figure}[H]
    \centering
    \subfigure[Ordinal Krippendorff's $\alpha$ (ranking)]{%
        \includegraphics[width=0.99\textwidth]{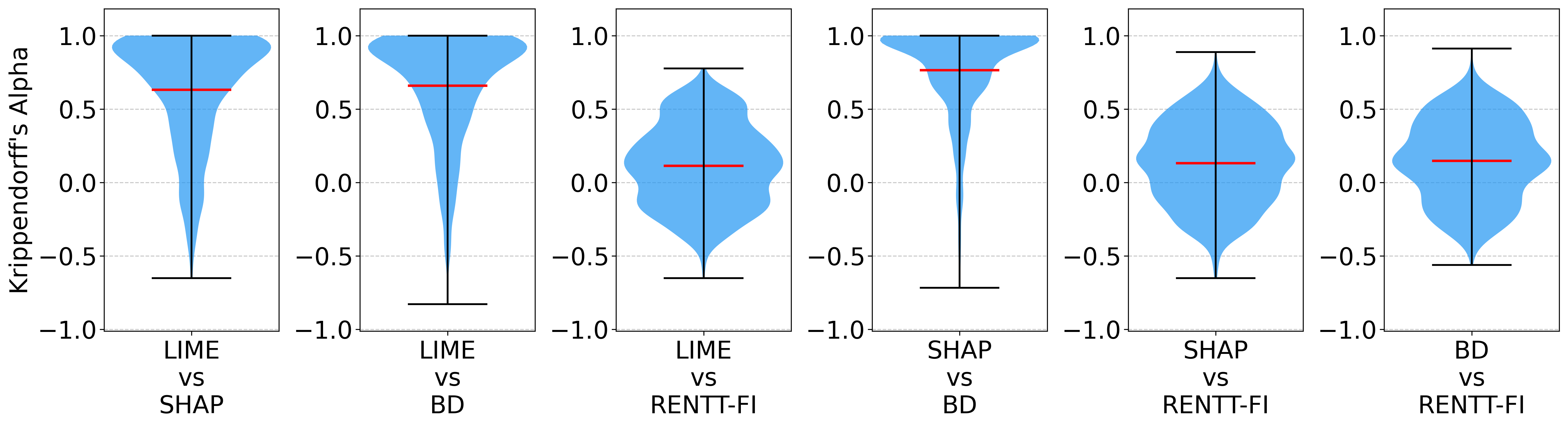}%
    }\\[0.5cm]
    \subfigure[Interval Krippendorff's $\alpha$ (values)]{%
        \includegraphics[width=0.99\textwidth]{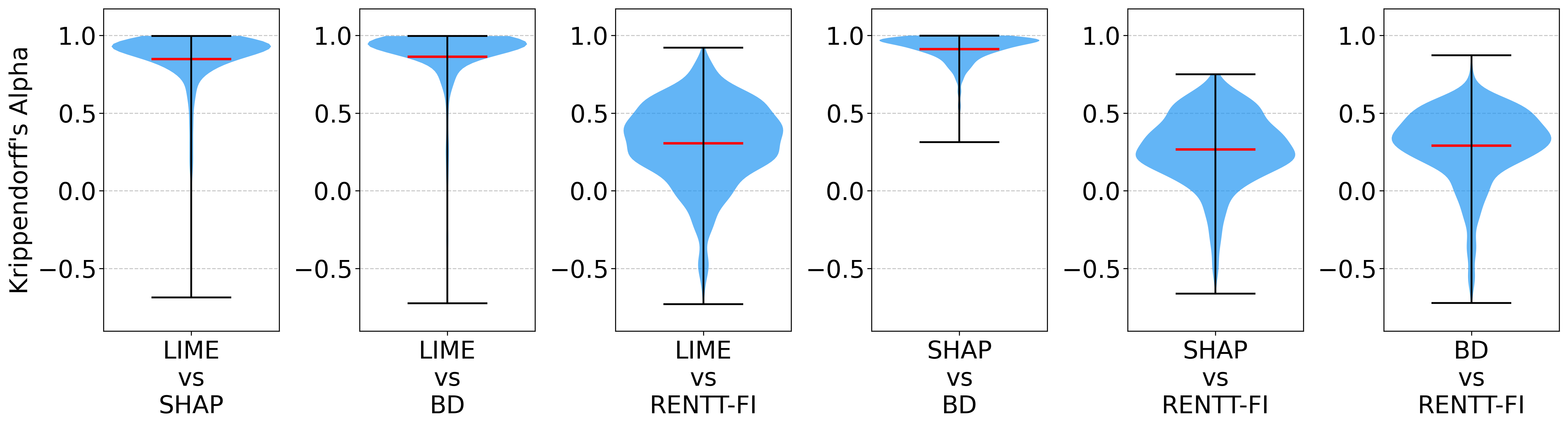}%
    }
    \caption{Violin plots of Krippendorff's $\alpha$ for the California Housing data set, calculated samplewise between different \ac{fi} methods. Only the 95th percentile of samples is pictured.}
    \label{fig:FI_california_krippendorff}
\end{figure}

\begin{figure}[H]
    \centering
    \subfigure[\ac{se} feature contribution]{%
        \includegraphics[width=0.48\textwidth]{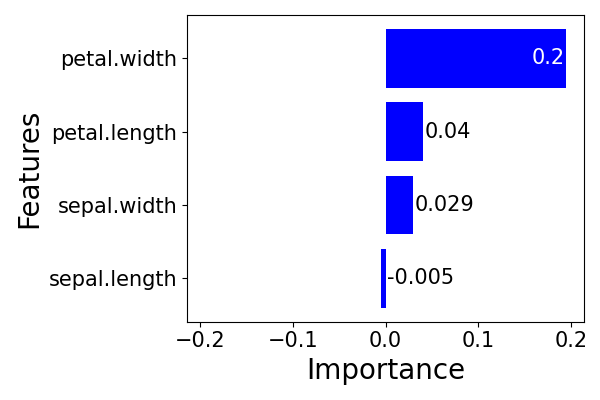}%
    }\hfill
    \subfigure[\ac{sage} feature contribution]{%
        \includegraphics[width=0.48\textwidth]{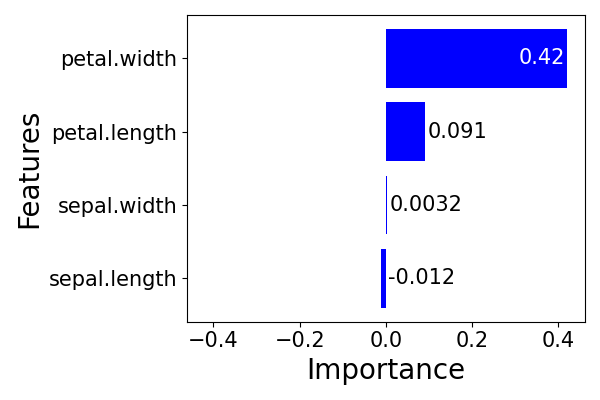}%
    }\\
    \subfigure[\ac{rentt}-\ac{fi} feature contribution]{%
        \includegraphics[width=0.48\textwidth]{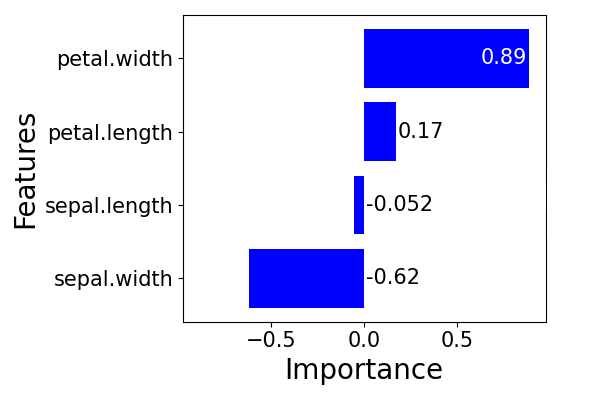}%
    }\hfill
    \subfigure[\ac{rentt}-\ac{fi} feature effect]{%
        \includegraphics[width=0.48\textwidth]{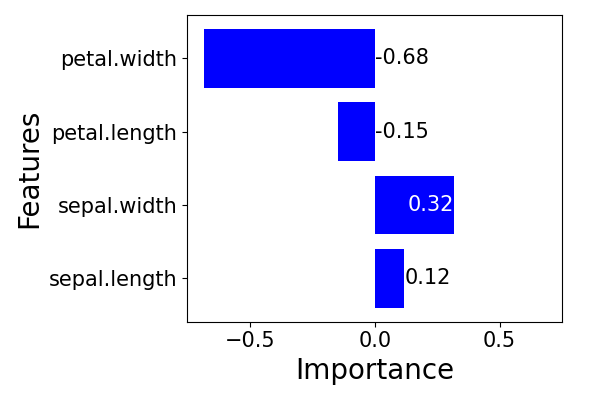}%
    }\\
    \subfigure[\ac{rentt}-\ac{fi} feature effect classwise (Classes: Setosa, Versicolour, and Virginica)]{%
        \includegraphics[width=1\textwidth]{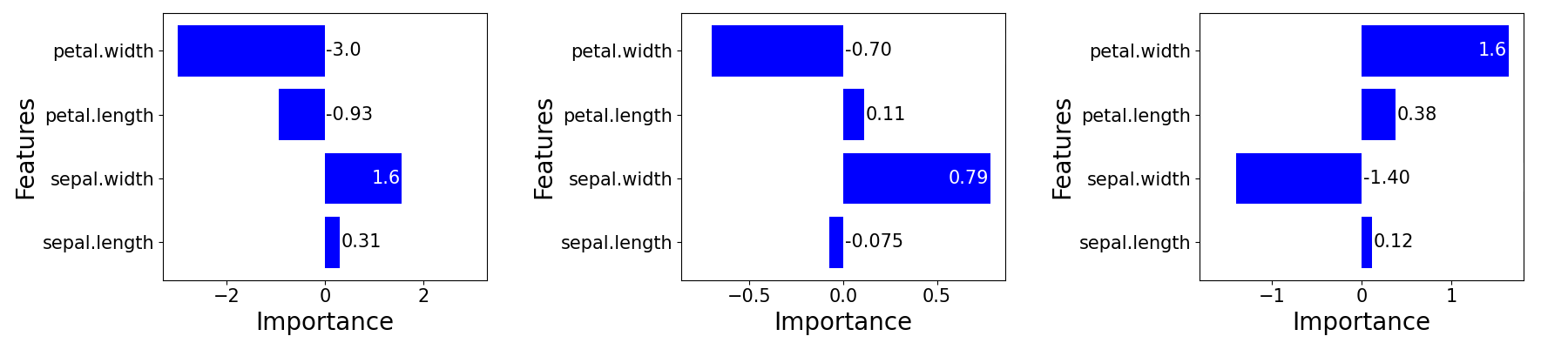}%
    }
    \caption{Global \ac{fi} by \ac{se}, \ac{sage} and \ac{rentt} on an \ac{nn} trained on the Iris data set.}
    \label{fig:FI_iris_global}
\end{figure}

\begin{figure}[H]
    \centering
    \includegraphics[width=1\linewidth]{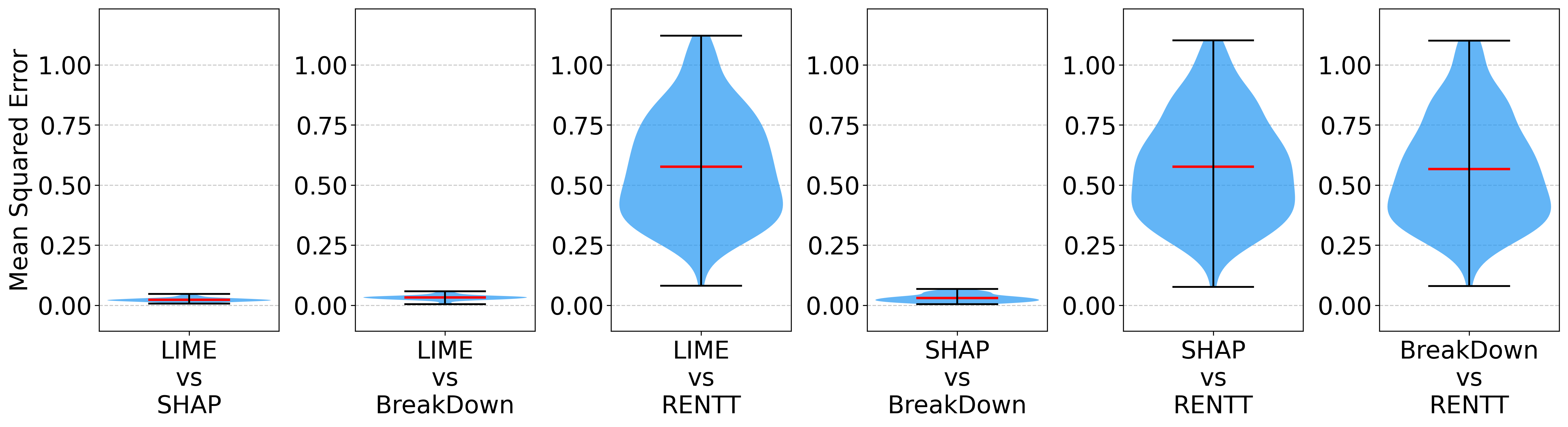}
    \caption{Violin plot of the \ac{rmse} between the different \ac{fi}-methods for the Iris data set. Only the 95th percentile of samples is pictured.}
    \label{fig:FI_iris_rmse_violin}
\end{figure}

\begin{figure}[H]
    \centering
    \subfigure[Ordinal Krippendorff's $\alpha$ (ranking)]{%
        \includegraphics[width=0.99\textwidth]{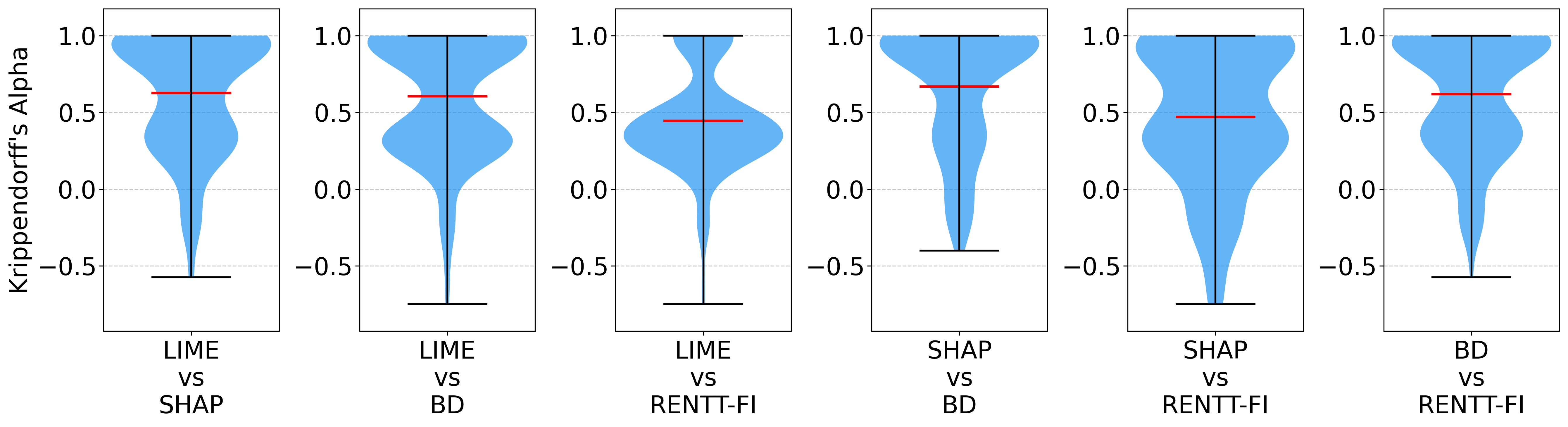}%
    }\\[0.5cm]
    \subfigure[Interval Krippendorff's $\alpha$ (values)]{%
        \includegraphics[width=0.99\textwidth]{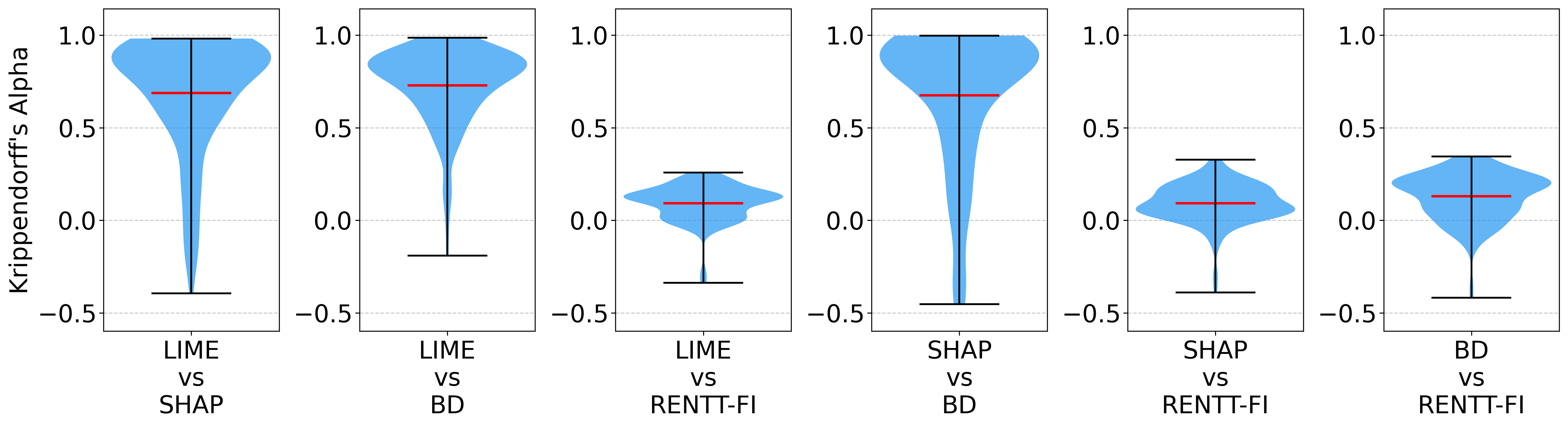}%
    }
    \caption{Violin plots of Krippendorff's $\alpha$ for the Iris data set, calculated samplewise between different \ac{fi} methods. Only the 95th percentile of samples is pictured.}
    \label{fig:FI_iris_krippendorff}
\end{figure}

\begin{figure}[H]
    \centering
    \subfigure[\ac{se} feature contribution]{%
        \includegraphics[height=0.24\textheight]{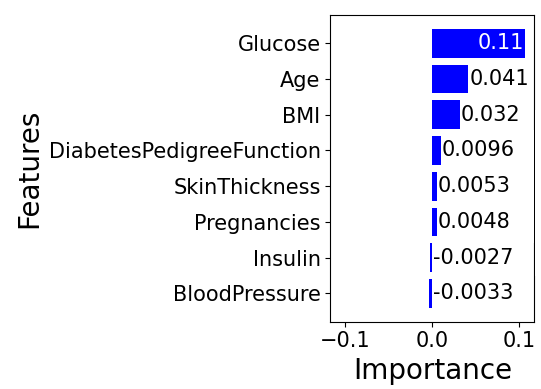}%
    }\hfill
    \subfigure[\ac{sage} feature contribution]{%
        \includegraphics[height=0.24\textheight]{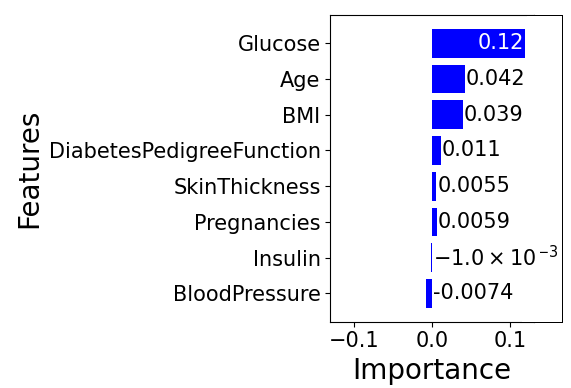}%
    }\\
    \subfigure[\ac{rentt}-\ac{fi} feature contribution]{%
        \includegraphics[height=0.24\textheight]{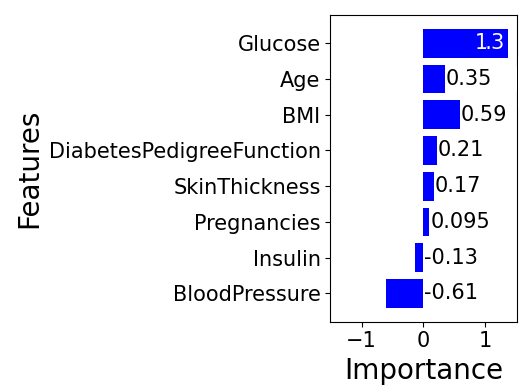}%
    }\hfill
    \subfigure[\ac{rentt}-\ac{fi} feature effect]{%
        \includegraphics[height=0.24\textheight]{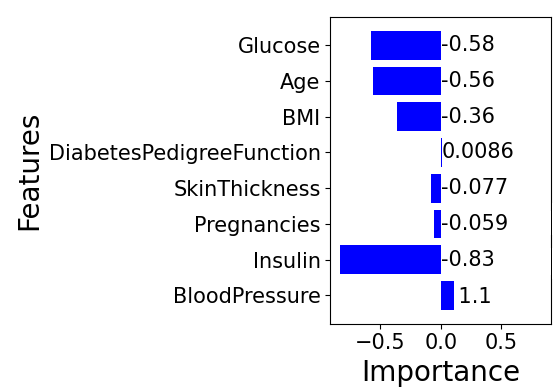}%
    }\\
    \subfigure[\ac{rentt}-\ac{fi} feature effect classwise (Classes: negative, positive)]{%
        \includegraphics[height=0.24\textheight]{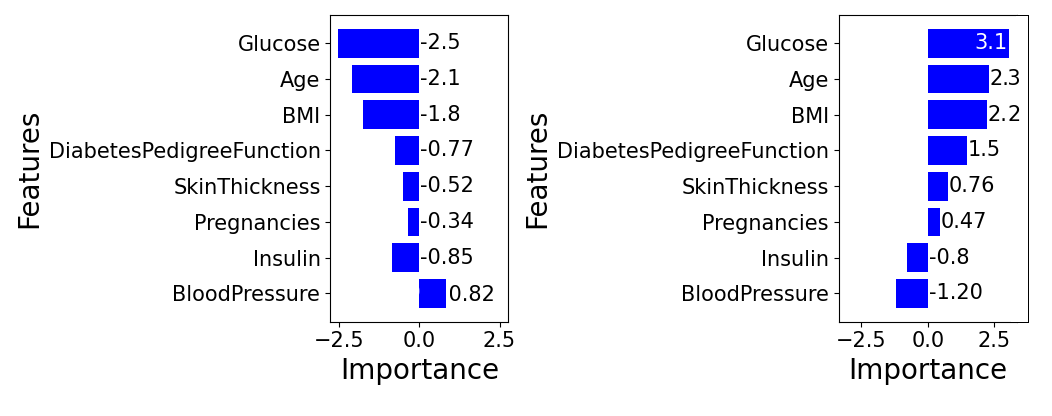}%
    }
    \caption{Global \ac{fi} by \ac{se}, \ac{sage} and \ac{rentt} on an \ac{nn} trained on the Diabetes Class data set.}
    \label{fig:FI_diabetes_global}
\end{figure}

\begin{figure}[H]
    \centering
    \includegraphics[width=1\linewidth]{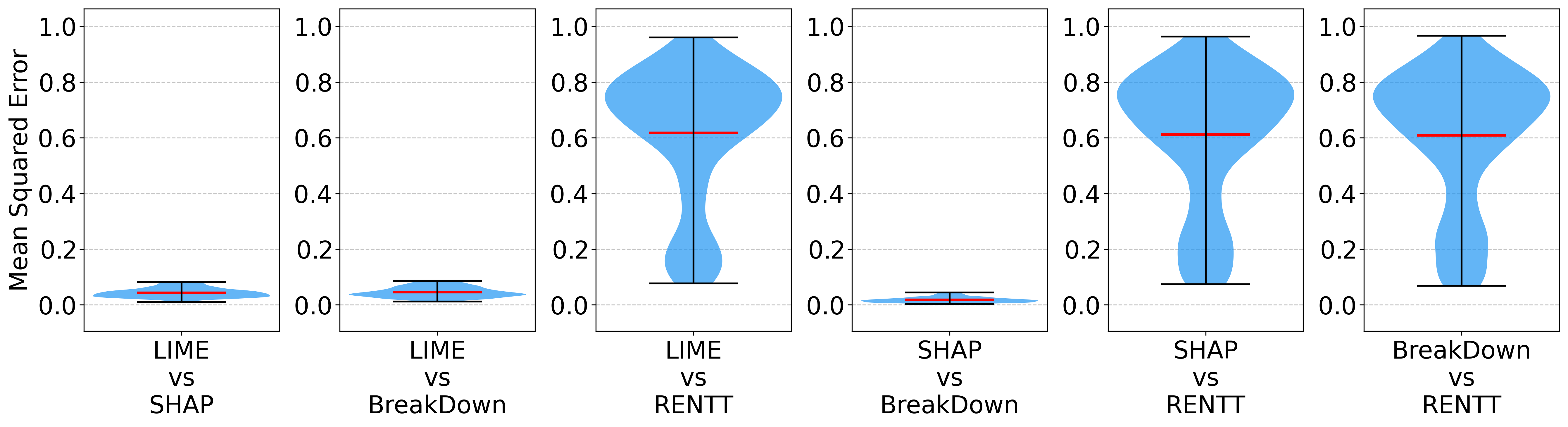}
    \caption{Violin plot of the \ac{rmse} between the different \ac{fi}-methods for the Diabetes Class data set. Only the 95th percentile of samples is pictured.}
    \label{fig:FI_diabetes_rmse_violin}
\end{figure}

\begin{figure}[H]
    \centering
    \subfigure[Ordinal Krippendorff's $\alpha$ (ranking)]{%
        \includegraphics[width=0.99\textwidth]{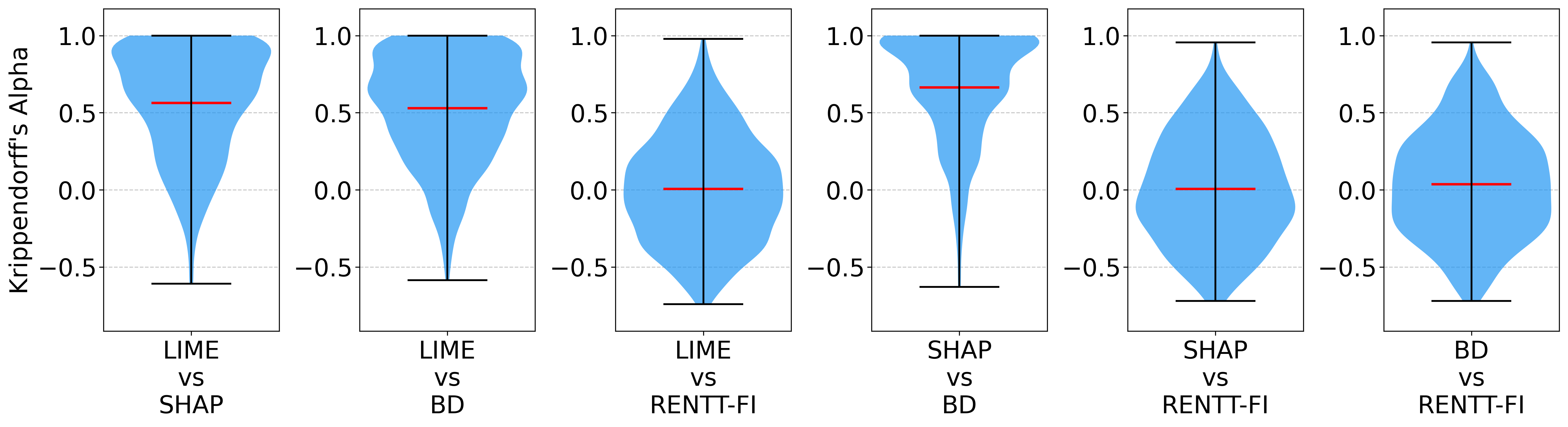}%
    }\\[0.5cm]
    \subfigure[Interval Krippendorff's $\alpha$ (values)]{%
        \includegraphics[width=0.99\textwidth]{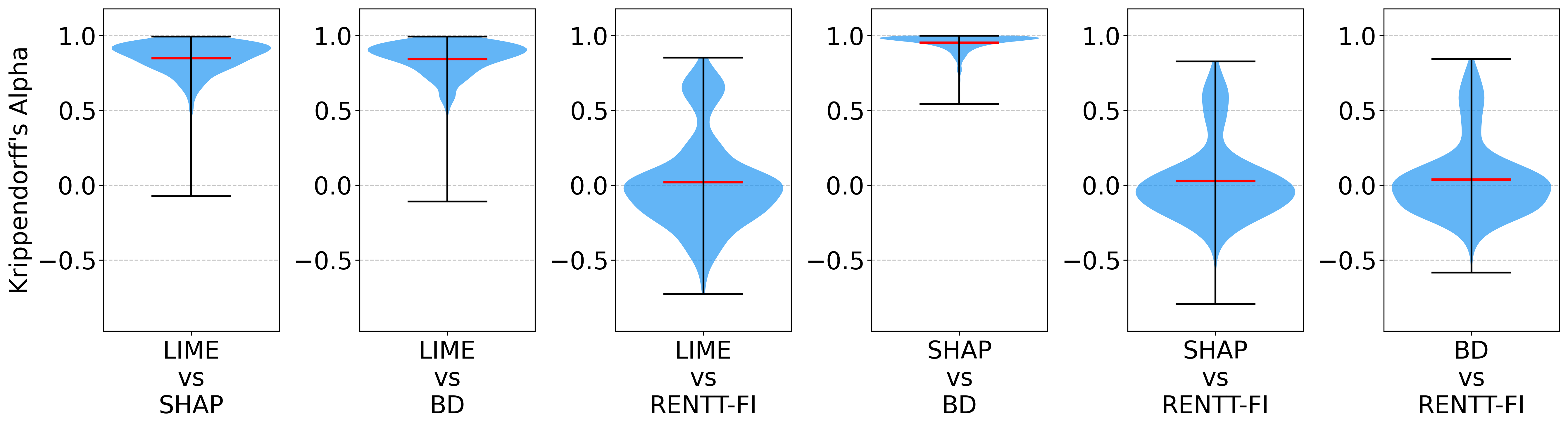}%
    }
    \caption{Violin plots of Krippendorff's $\alpha$ for the Diabetes Class data set, calculated samplewise between different \ac{fi} methods. Only the 95th percentile of samples is pictured.}
    \label{fig:FI_diabetes_class_krippendorff}
\end{figure}

\begin{figure}[H]
    \centering
    \subfigure[\ac{se} feature contribution]{%
        \includegraphics[width=0.48\textwidth]{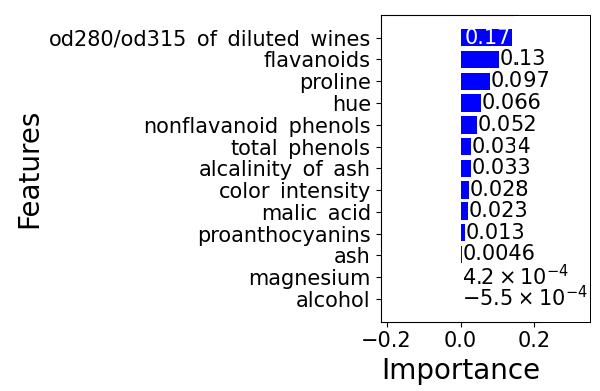}%
    }\hfill
    \subfigure[\ac{sage} feature contribution]{%
        \includegraphics[width=0.48\textwidth]{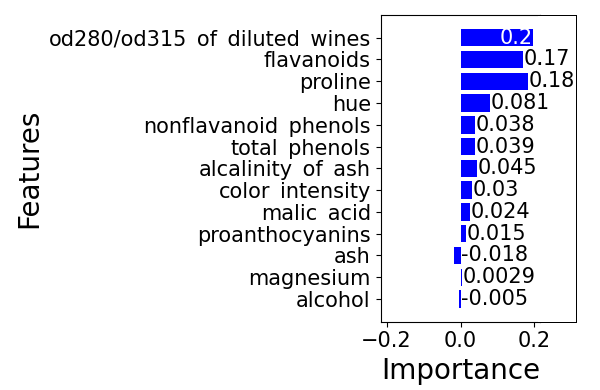}%
    }\\
    \subfigure[\ac{rentt}-\ac{fi} feature contribution]{%
        \includegraphics[width=0.48\textwidth]{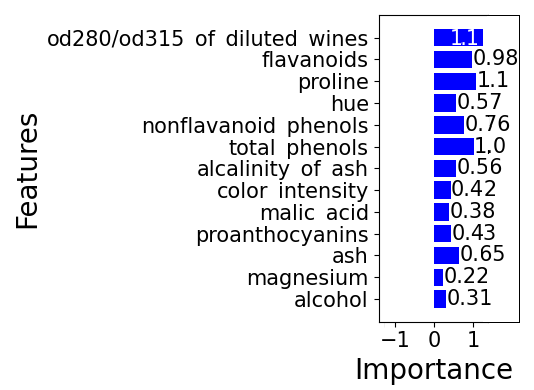}%
    }\hfill
    \subfigure[\ac{rentt}-\ac{fi} feature effect]{%
        \includegraphics[width=0.48\textwidth]{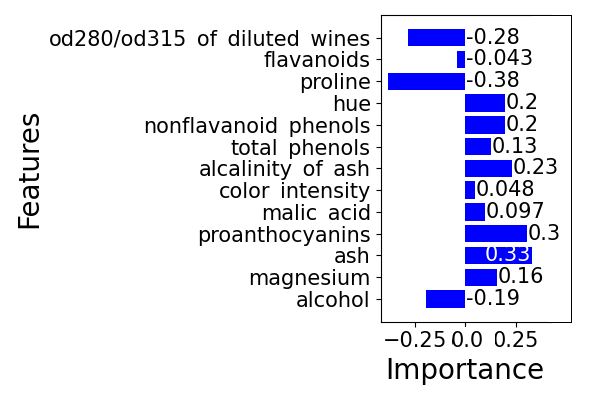}%
    }\\
    \subfigure[\ac{rentt}-\ac{fi} feature effect classwise (Classes:low, medium, high quality)]{%
        \includegraphics[width=1\textwidth]{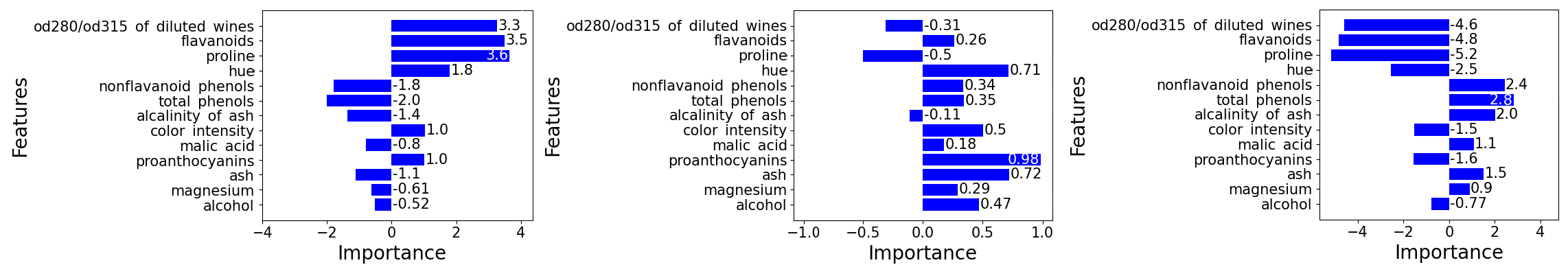}%
    }
    \caption{Global \ac{fi} by \ac{se}, \ac{sage} and \ac{rentt} on an \ac{nn} trained on the Wine Quality data set.}
    \label{fig:FI_Wine_global}
\end{figure}

\begin{figure}[H]
    \centering
    \includegraphics[width=1\linewidth]{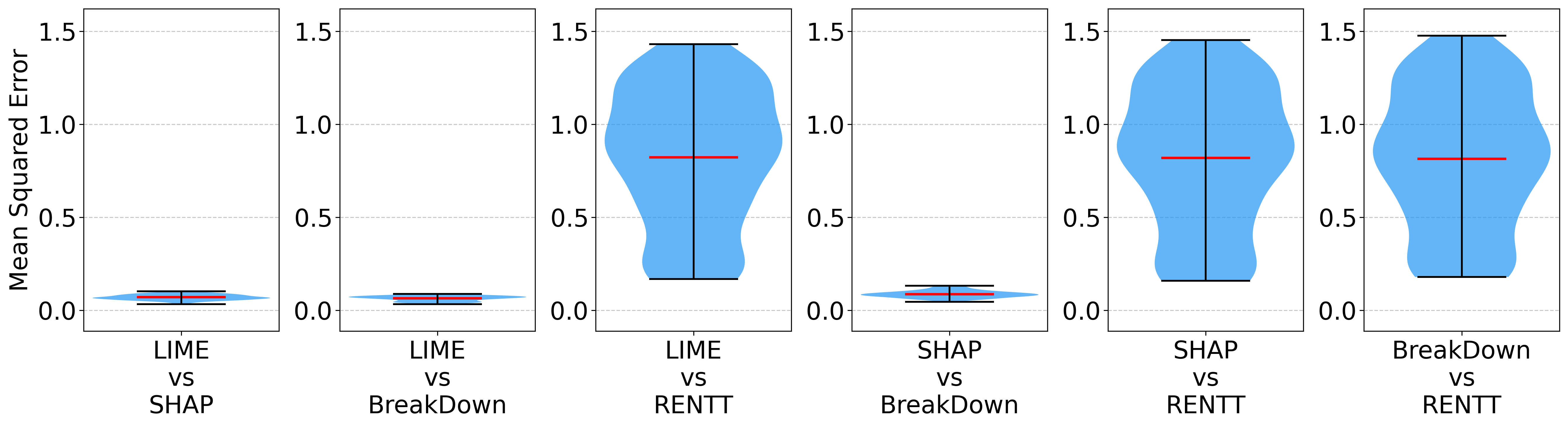}
    \caption{Violin plot of the \ac{rmse} between the different \ac{fi}-methods for the Wine Quality data set. Only the 95th percentile of samples is pictured.}
    \label{fig:FI_wine_rmse_violin}
\end{figure}

\begin{figure}[H]
    \centering
    \subfigure[Ordinal Krippendorff's $\alpha$ (ranking)]{%
        \includegraphics[width=0.99\textwidth]{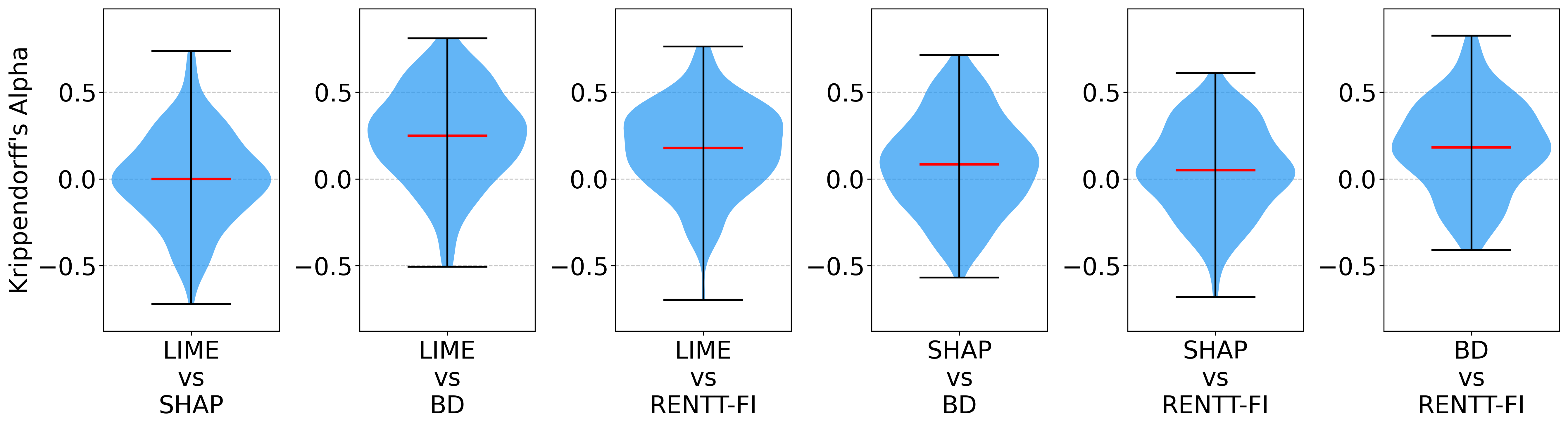}%
    }\\[0.5cm]
    \subfigure[Interval Krippendorff's $\alpha$ (values)]{%
        \includegraphics[width=0.99\textwidth]{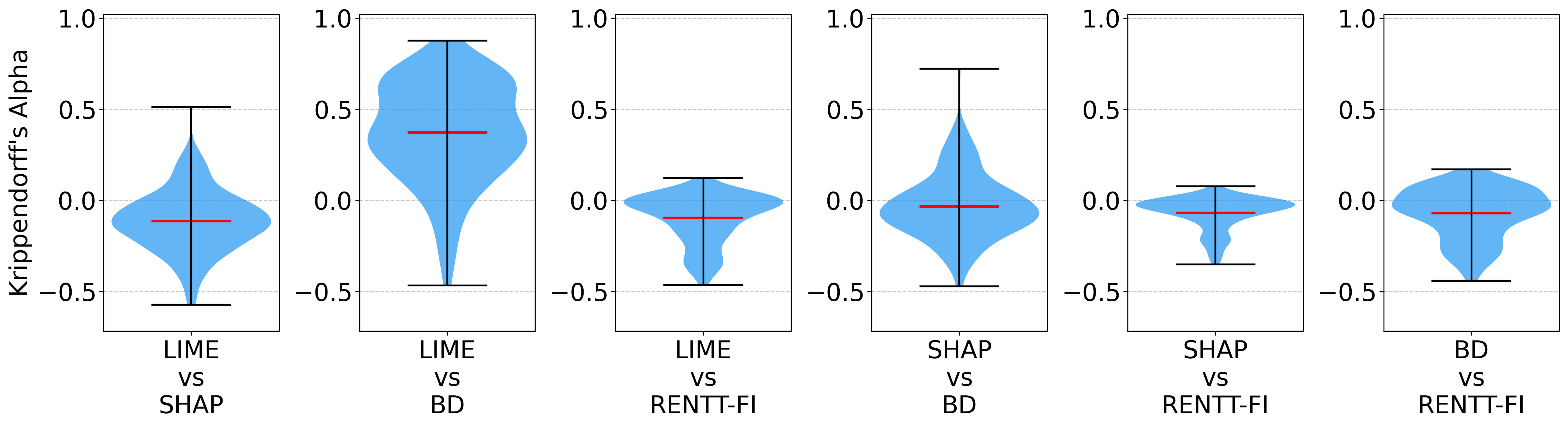}%
    }
    \caption{Violin plots of Krippendorff's $\alpha$ for the Wine Quality data set, calculated samplewise between different \ac{fi} methods. Only the 95th percentile of samples is pictured.}
    \label{fig:FI_wine_krippendorff}
\end{figure}

\begin{figure}[H]
    \centering
    \subfigure[\ac{se} feature contribution]{%
        \includegraphics[width=0.43\textwidth]{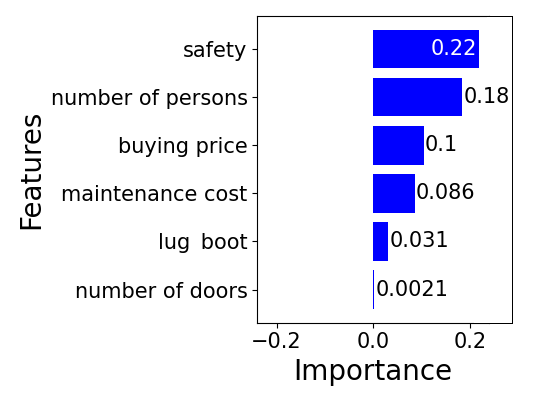}%
    }\hfill
    \subfigure[\ac{sage} feature contribution]{%
        \includegraphics[width=0.48\textwidth]{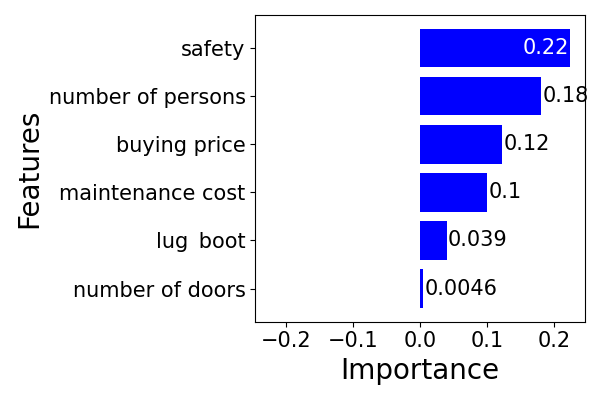}%
    }\\
    \subfigure[\ac{rentt}-\ac{fi} feature contribution]{%
        \includegraphics[width=0.34\textwidth]{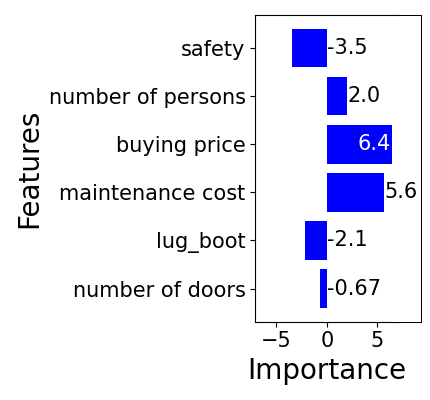}%
    }\hfill
    \subfigure[\ac{rentt}-\ac{fi} feature effect]{%
        \includegraphics[width=0.48\textwidth]{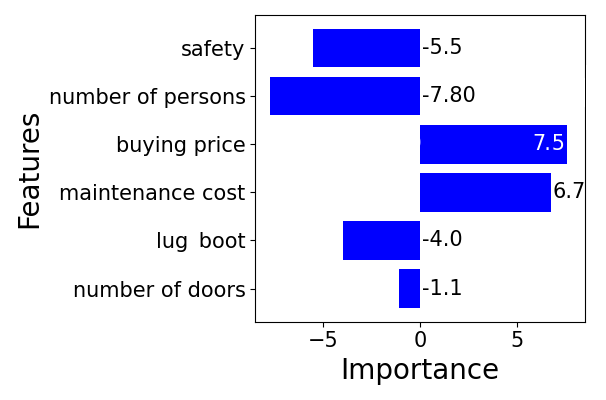}%
    }\\
    \subfigure[\ac{rentt}-\ac{fi} feature effect classwise (Classes: unacceptable, acceptable, good, very good)]{%
        \includegraphics[width=1\textwidth]{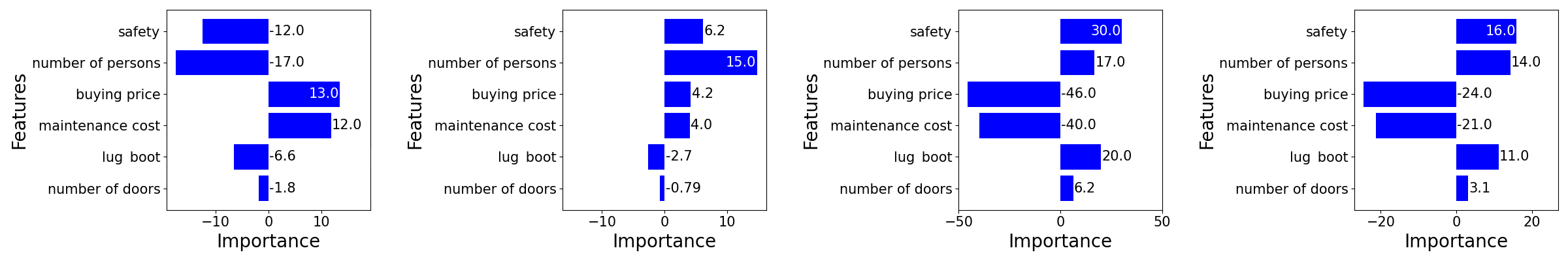}%
    }
    \caption{Global \ac{fi} by \ac{se}, \ac{sage} and \ac{rentt} on an \ac{nn} trained on the Car Evaluation data set.}
    \label{fig:FI_CarEvaluation_global}
\end{figure}

\begin{figure}[H]
    \centering
    \includegraphics[width=1\linewidth]{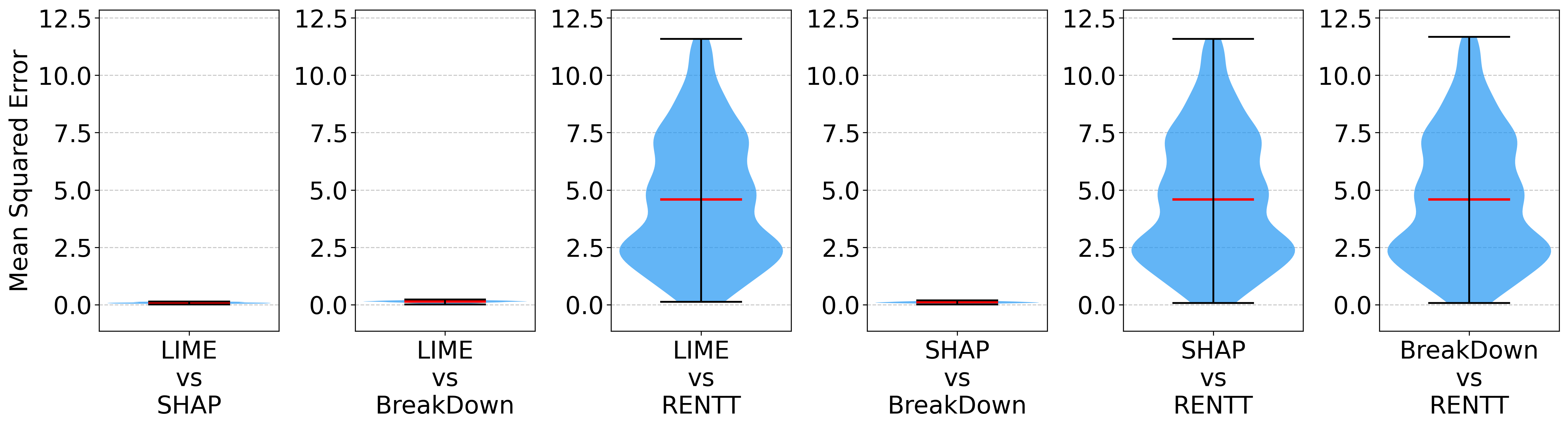}
    \caption{Violin plot of the \ac{rmse} between the different \ac{fi}-methods for the Car Evaluation data set. Only the 95th percentile of samples is pictured.}
    \label{fig:FI_car_rmse_violin}
\end{figure}

\begin{figure}[H]
    \centering
    \subfigure[Ordinal Krippendorff's $\alpha$ (ranking)]{%
        \includegraphics[width=0.99\textwidth]{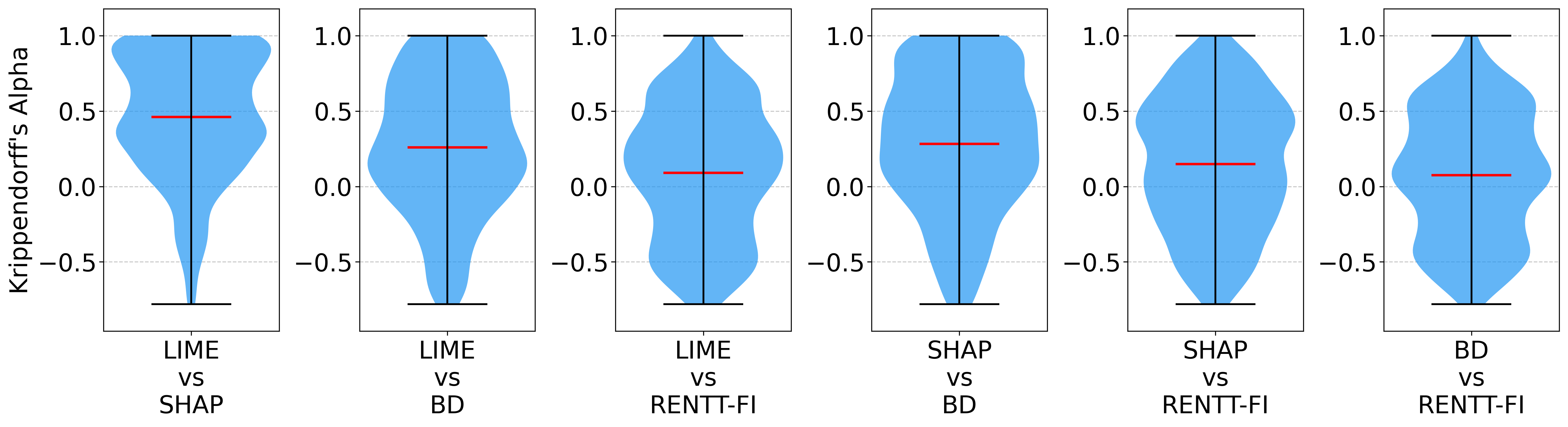}%
    }\\[0.5cm]
    \subfigure[Interval Krippendorff's $\alpha$ (values)]{%
        \includegraphics[width=0.99\textwidth]{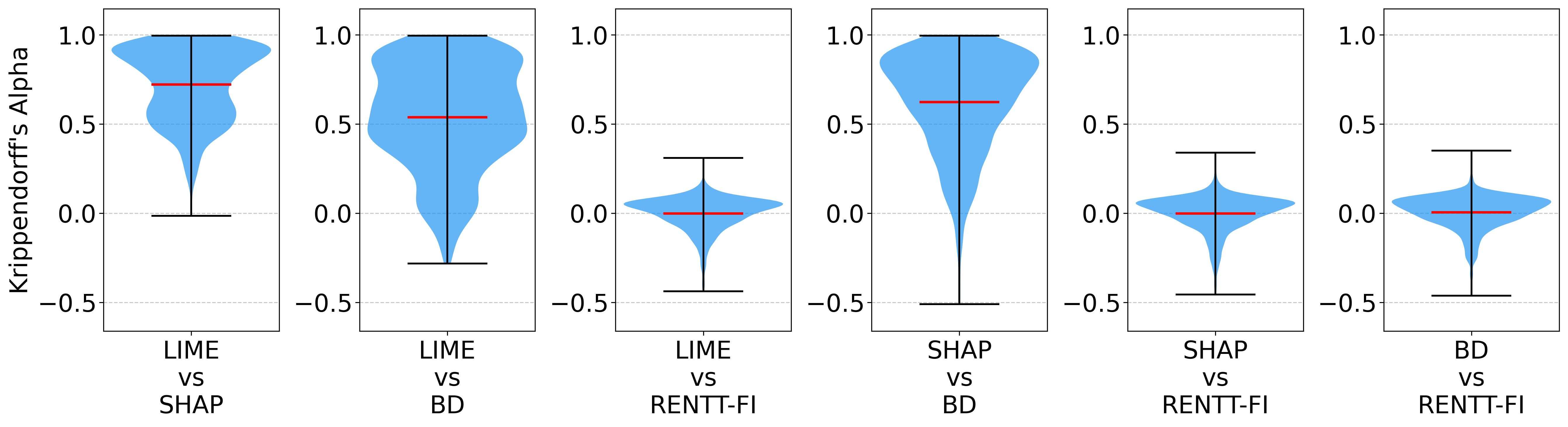}%
    }
    \caption{Violin plots of Krippendorff's $\alpha$ for the Car Evaluation data set, calculated samplewise between different \ac{fi} methods. Only the 95th percentile of samples is pictured.}
    \label{fig:FI_car_krippendorff}
\end{figure}

\begin{figure}[H]
    \centering
    \subfigure[\ac{se} feature contribution]{%
        \includegraphics[height=0.5\textheight]{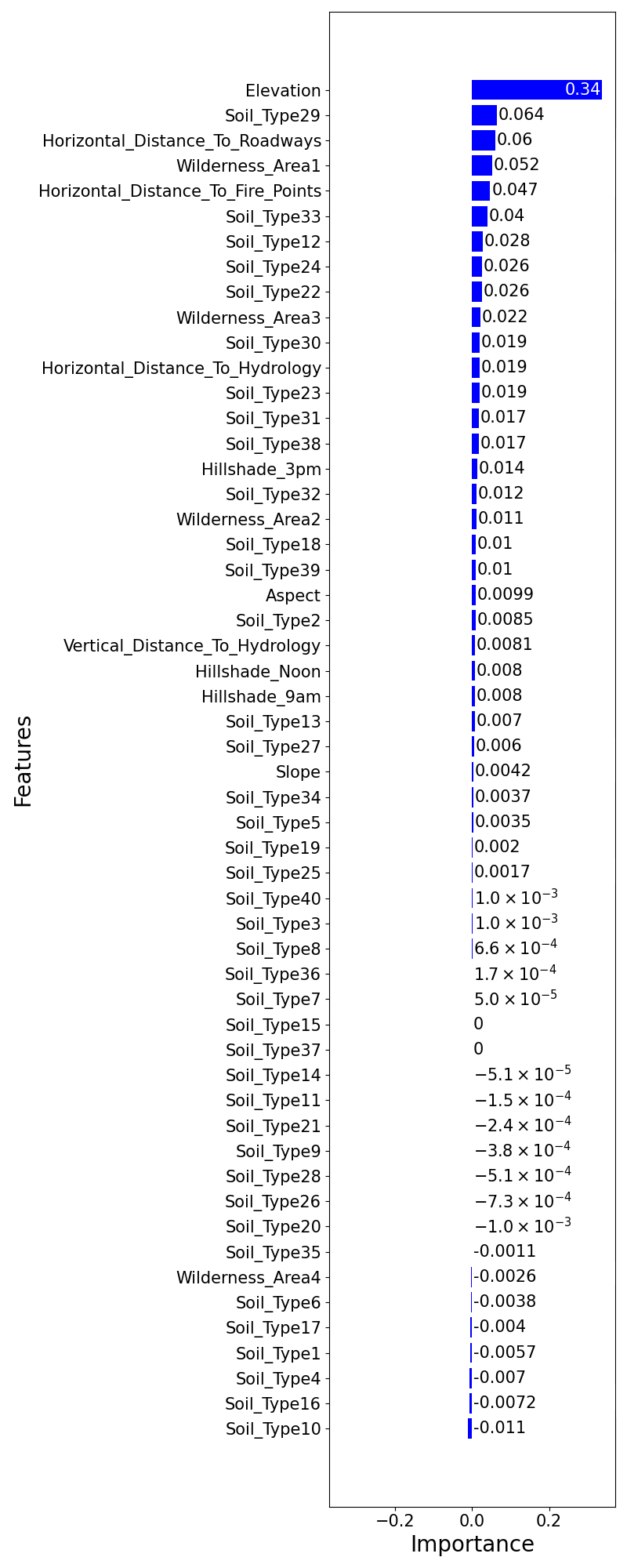}%
    }\hfill
    \subfigure[\ac{sage} feature contribution]{%
        \includegraphics[height=0.5\textheight]{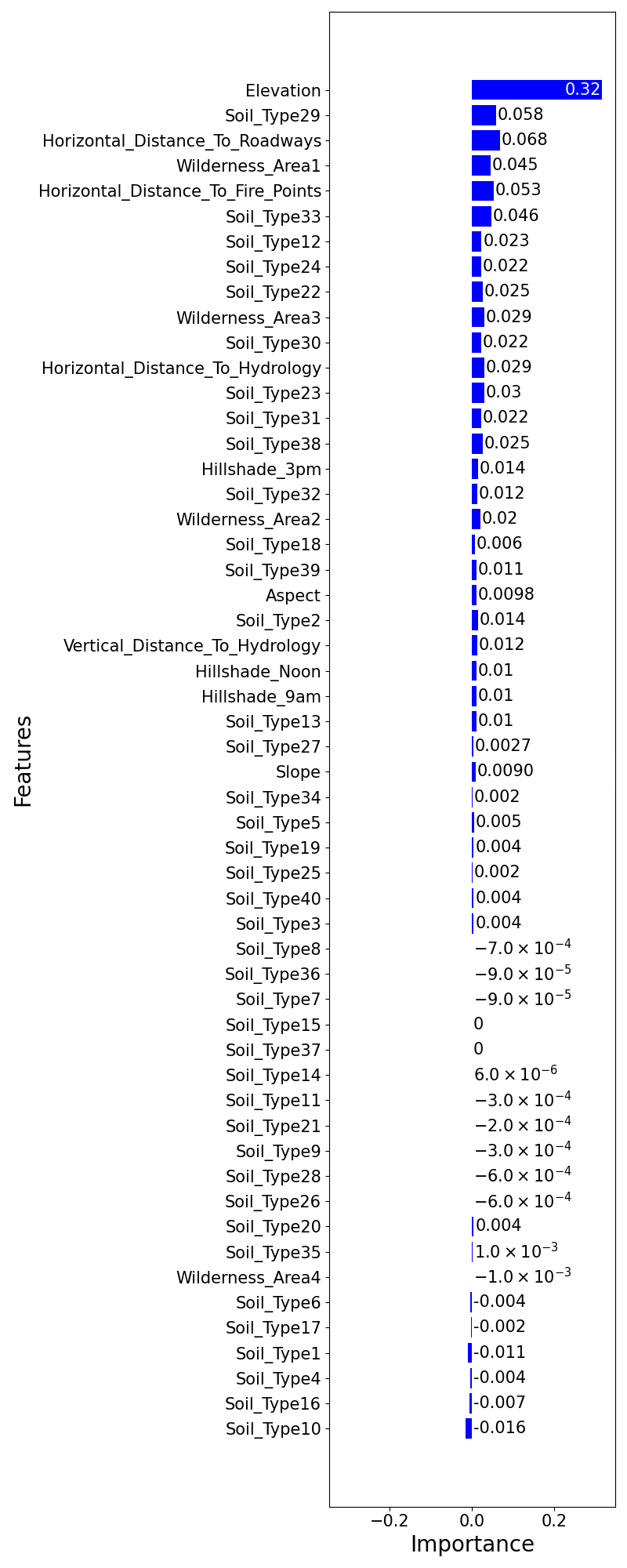}%
    }\hfill
    \subfigure[\ac{rentt}-\ac{fi} feature contribution]{%
        \includegraphics[height=0.5\textheight]{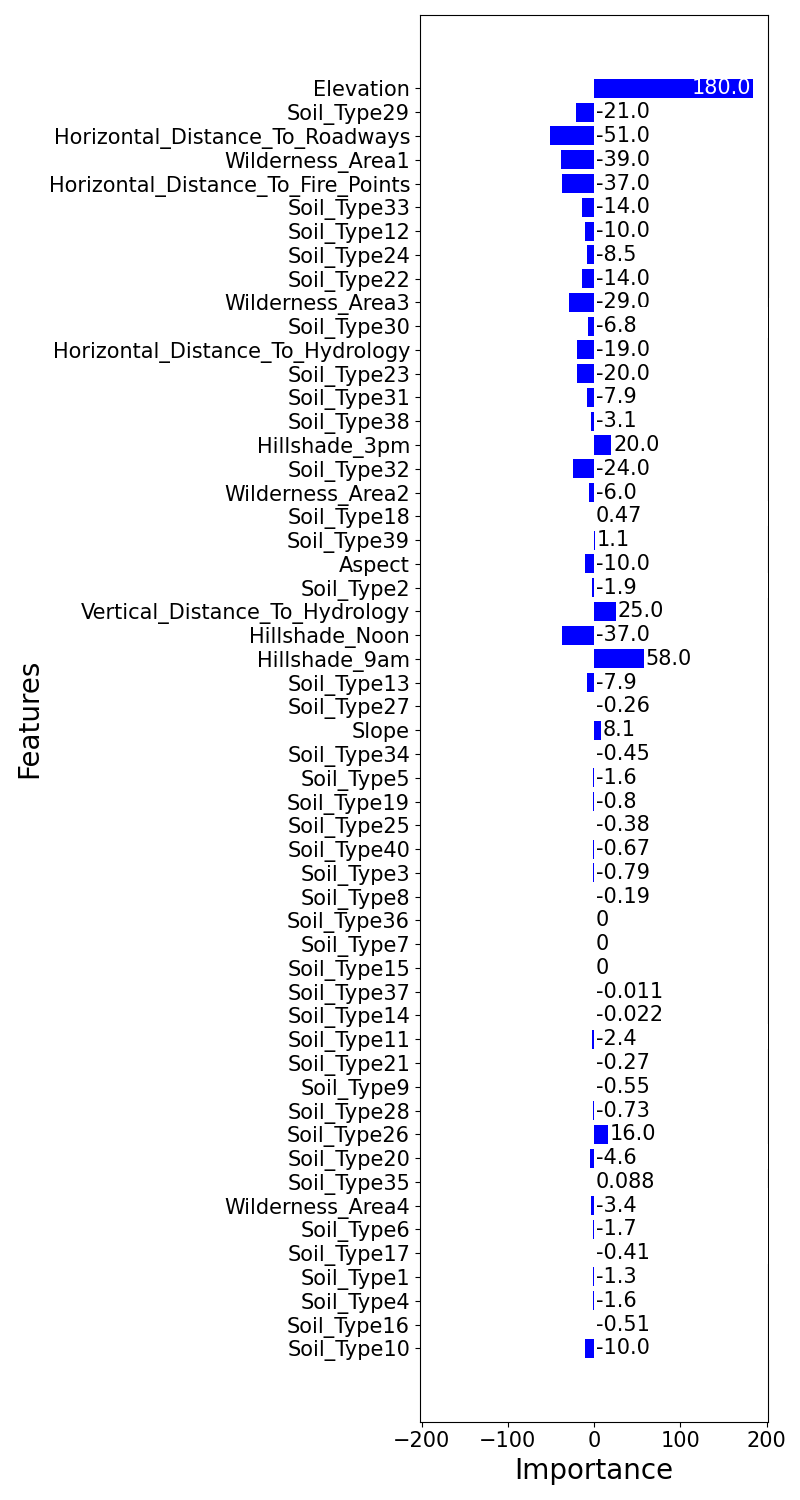}%
    }
    \caption{Global feature contribution by \ac{se}, \ac{sage} and \ac{rentt} on an \ac{nn} trained on the Forest Cover Type data set.}
    \label{fig:FI_Covertype_global_contribution}
\end{figure}

\begin{figure}[H]
    \centering
    \subfigure[\ac{rentt} global feature effect]{%
        \includegraphics[height=0.29\textheight]{img/Results/FI/Covertype/FI_FeatureContribution_global.png}%
    }\hfill
    \subfigure[\ac{rentt} feature effect for the classes Spruce/Fir, Lodgepole Pine, Ponderosa Pine]{%
        \includegraphics[height=0.29\textheight]{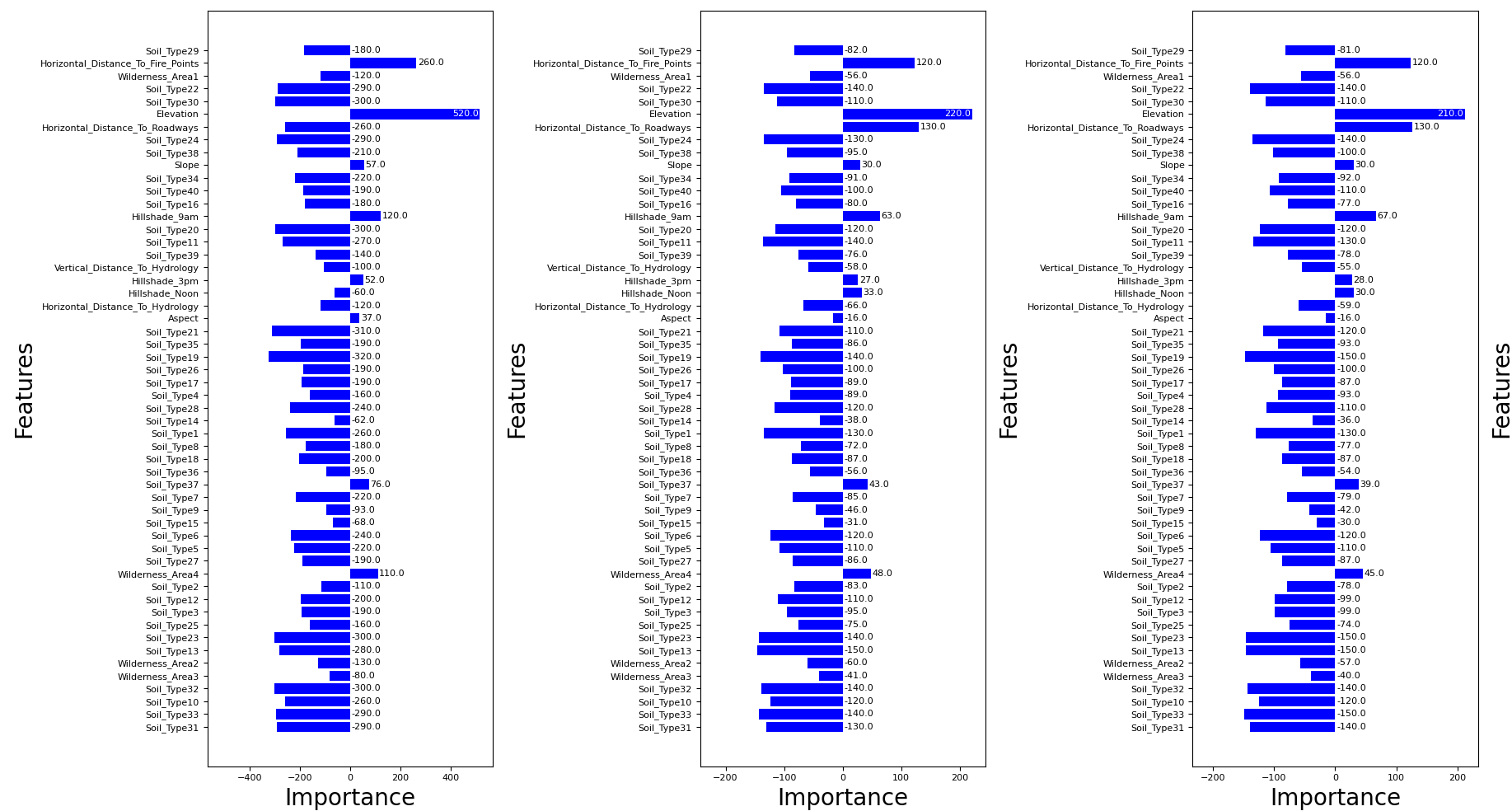}%
    }\\
    \subfigure[\ac{rentt} feature effect for the classes Cottonwood/Willow, Aspen, Douglas-fir, Krummholz]{%
        \includegraphics[width=\textwidth]{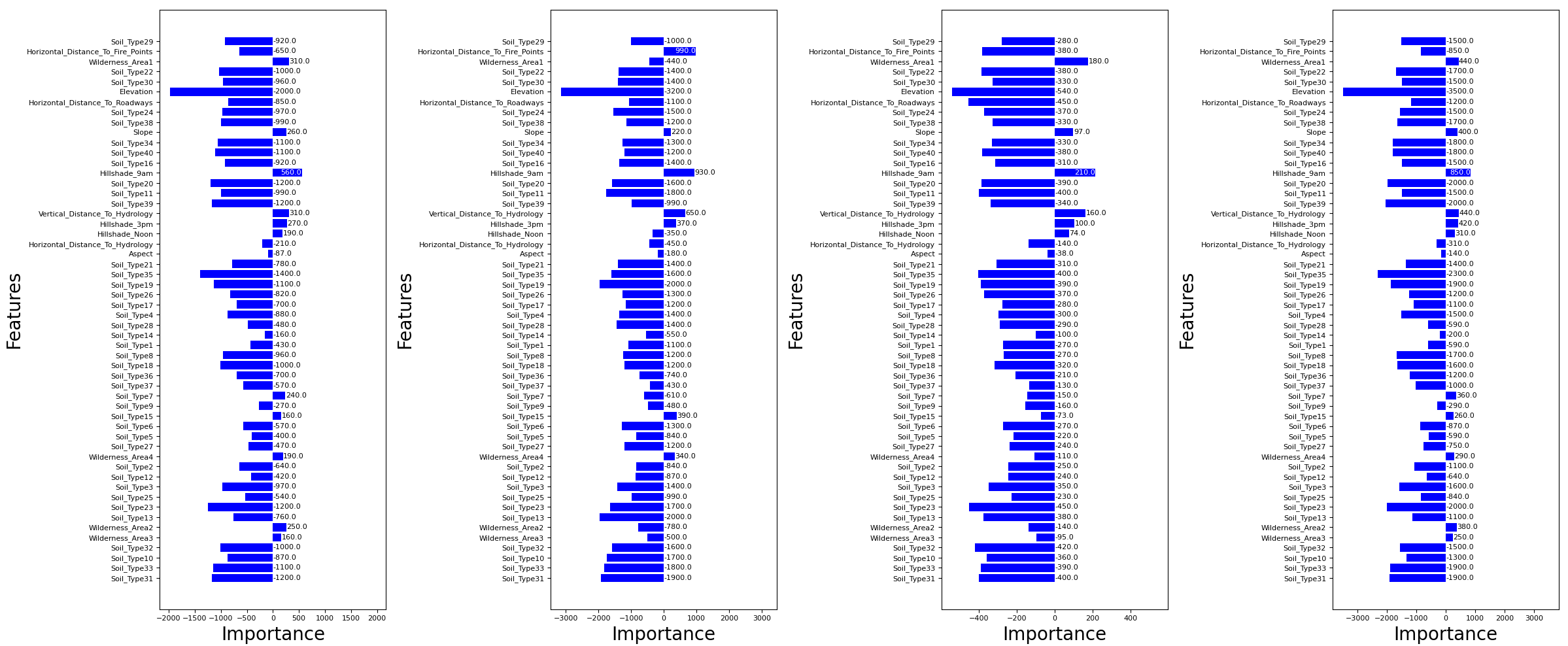}%
    }
    \caption{Feature effect global and classwise by \ac{rentt} on an \ac{nn} trained on the Forest Cover Type data set.}
    \label{fig:FI_Covertype_global_effect}
\end{figure}

    

\pagebreak
\section{Results for the Runtime Analysis of Feature Importance Methods} \label{a-ssec:runtime}
This section presents the computational efficiency analysis of different feature importance methods. The runtime measurements demonstrate the significant computational advantages of RENTT-FI over traditional perturbation-based methods.
The results show that RENTT-FI is orders of magnitude faster than LIME, SHAP, BD, SE, and SAGE methods across all data sets.

\begin{table}[H]
    \caption{Mean runtime results for local \ac{fi} methods with standard deviation.}
    \centering
    \begin{tabular}{|c|c|c|}
        \hline
        \textbf{Data Set} & \textbf{Method} & \textbf{Runtime / s} \\
            \hline
         & LIME & $13.0 \pm 0.2$ \\
        Linear &  \ac{shap} & $74.4 \pm 0.2$ \\
         &  \ac{bd} & $25.8 \pm 0.1$ \\
         &  \ac{rentt}-\ac{fi} & $0.024 \pm 0.000$ \\
        \hline

        & \ac{lime} & $12.8 \pm 0.1$ \\
         Diabetes & \ac{shap} & $163.5 \pm 0.4$ \\
         Reg & \ac{bd} & $62.1 \pm 0.1$ \\
         & \ac{rentt}-\ac{fi} & $0.027 \pm 0.000$ \\
        \hline
          & \ac{lime} & $12.8 \pm 0.1$ \\
        California & \ac{shap} & $154.8 \pm 1.2$ \\
         Housing & \ac{bd} & $58.0 \pm 0.4$ \\
         &  \ac{rentt}-\ac{fi}  & $0.042 \pm 0.007$ \\
        \hline
        \hline
         & \ac{lime} & $24.6 \pm 0.1$ \\
         Iris & \ac{shap} & $74.8 \pm 0.1$ \\
         & \ac{bd} & $26.4 \pm 0.1$ \\
         &  \ac{rentt}-\ac{fi}  & $0.025 \pm 0.001$ \\
        \hline
         & \ac{lime} & $24.9 \pm 0.1$ \\
        Diabetes & \ac{shap} & $151.9 \pm 0.6$ \\
        Class & \ac{bd} & $56.4 \pm 0.3$ \\
         &  \ac{rentt}-\ac{fi}  & $0.027 \pm 0.001$ \\
        \hline
         & \ac{lime} & $27.8 \pm 0.2$ \\
        Car & \ac{shap} & $134.8 \pm 0.3$ \\
        Evaluation & \ac{bd} & $49.2 \pm 0.3$ \\
         &  \ac{rentt}-\ac{fi}  & $0.043 \pm 0.007$ \\
        \hline
          & \ac{lime} & $24.4 \pm 0.1$ \\
        Wine & \ac{shap} & $198.1 \pm 0.1$ \\
         Quality & \ac{bd} & $75.7 \pm 0.1$ \\
         &  \ac{rentt}-\ac{fi}  & $0.025 \pm 0.000$ \\
            \hline
    \end{tabular}
    \label{tab:runtime_local}
\end{table}

\begin{table}[H]
    \caption{Mean runtime results for global \ac{fi} methods with standard deviation.}
    \centering
    \begin{tabular}{|c|c|c|}
        \hline
        \textbf{Data Set} & \textbf{Method} & \textbf{Runtime / s} \\
        \hline
        & \ac{se}& $0.250 \pm 0.040$ \\
         Linear &  \ac{sage} & $0.233 \pm 0.063$ \\
         &  \ac{rentt}-\ac{fi} & $(16.0 \pm 0.4)\cdot 10^{-4}$ \\
        \hline
         Diabetes & \ac{se} & $1.8 \pm 0.1$ \\
         Reg & \ac{sage} & $2.0 \pm 0.2$ \\
         & \ac{rentt}-\ac{fi} & $(4.1e-03 \pm 4.3e-05)$ \\
         \hline
        California  & \ac{se}& $12.6 \pm 0.8$ \\
         Housing &  \ac{sage} & $22.4 \pm 1.8$ \\
         &  \ac{rentt}-\ac{fi}  & $0.019 \pm 0.007$ \\
        \hline
        \hline
         & SE& $0.060 \pm 0.005$ \\
        Iris &  \ac{sage} & $0.025 \pm 0.005$ \\
         &  \ac{rentt}-\ac{fi}  & $0.0025 \pm 0.0002$ \\
        \hline
        Diabetes & \ac{se}& $1.4 \pm 0.0$ \\
        Class &  \ac{sage} & $12.9 \pm 0.7$ \\
        &  \ac{rentt}-\ac{fi}  & $0.0040 \pm 0.0003$ \\
        \hline
        Car & \ac{se}& $175.2 \pm 5.4$ \\
        Evaluation &  \ac{sage} & $222.1 \pm 18.6$ \\
        &  \ac{rentt}-\ac{fi}  & $0.020 \pm 0.007$ \\
        \hline
        Wine  & SE& $0.460 \pm 0.057$ \\
        Quality &  \ac{sage} & $0.37 \pm 0.04$ \\
         &  \ac{rentt}-\ac{fi}  & $0.0017 \pm 0.0002$ \\
        \hline
    \end{tabular}
    \label{tab:runtime_global}
\end{table}